\documentclass[11pt]{article}
\usepackage{amsmath}

\usepackage[preprint]{acl}

\usepackage{times}
\usepackage{latexsym}

\usepackage[T1]{fontenc}
\usepackage[utf8]{inputenc}

\usepackage{microtype}

\usepackage{inconsolata}

\usepackage{graphicx}

\usepackage{comment}
\usepackage{multirow}
\usepackage{mathtools,amssymb}
\usepackage{booktabs}
\usepackage{array}
\usepackage{tikz}
\usepackage{pgf-pie}
\usepackage{xcolor}
\usepackage{diagbox}
\definecolor{darkgreen}{RGB}{0,110,0}
\usepackage{mdframed}

\title{On the Fallacy of Global Token Perplexity \\ in Spoken Language Model Evaluation}

\author{%
  Chan-Jan Hsu$^{\ddagger}$, Liang-Hsuan Tseng$^{\flat}$, Yi-Cheng Lin$^\flat$, Yen-Chun Kuo$^\flat$ \\
  \textbf{Ju-Chieh Chou}$^\natural$\textbf{,} \textbf{Kai-Wei Chang}$^\clubsuit$\textbf{,} \textbf{Hung-yi Lee}$^\flat$\textbf{,} \textbf{Carlos Busso}$^\ddagger$ \\
  $^\ddagger$Carnegie Mellon University, $^\flat$National Taiwan University, \\
  $^\natural$Toyota Technological Institute at Chicago, $^\clubsuit$Massachusetts Institute of Technology \\
  \texttt{chanjanh@andrew.cmu.edu, busso@cmu.edu} \\
}

\begin{document}
\newcommand{\TabCompositionComprehensive}{%
\begin{table*}[t]
\caption{Token-type ablations and Shapley attributions for Spirit-LM (HuBERT $H$, pitch $P$, style $S$) and Llama-Mimi (layers $0$--$3$), under four evaluation settings that combine estimator locality (Global vs. Localized) and scoring normalization (Original vs. Normalized). With the accuracy in each panel (top), the corresponding Shapley values $\phi$ for each component can be calculated (bottom), with a null baseline of $50\%$ on every task. Under the default evaluation setting (Global, Original), Spirit-LM’s largest contribution comes from HuBERT tokens, whereas Llama-Mimi’s largest contribution comes from layer~1. Across settings, the most pronounced differences concentrate on speaker-related attributes (sentiment, speaker, and gender).}
\label{tab:composition-comprehensive}
\small

\centering
\renewcommand{\arraystretch}{1.05}
\setlength{\tabcolsep}{4pt}

\begin{tabular}{@{}c cc@{}}
\hline
\multicolumn{3}{c}{\textbf{Spirit-LM}}\\
\hline
& \textbf{Original} & \textbf{Normalized} \\
\rotatebox{90}{\textbf{Global}}
&
\begin{minipage}[t]{0.47\textwidth}
\centering
\resizebox{\linewidth}{!}{%
\begin{tabular}{ccc|cccccc|c}
\toprule
H & P & S & Sent. & Spk. & Gen. & BgD. & BgR. & Room & Avg \\
\midrule
\checkmark & \checkmark & \checkmark & 71.0 & 81.5 & 85.5 & 55.5 & 64.0 & 55.5 & 68.8 \\
\checkmark & \checkmark & -- & 73.0 & 81.5 & 85.0 & 57.0 & 63.0 & 54.5 & 69.0 \\
\checkmark & -- & \checkmark & 57.5 & 67.0 & 74.0 & 49.5 & 58.0 & 57.0 & 60.5 \\
-- & \checkmark & \checkmark & 57.0 & 70.0 & 74.0 & 54.0 & 56.5 & 51.0 & 60.4 \\
\checkmark & -- & -- & 56.5 & 67.5 & 73.0 & 50.5 & 60.0 & 54.5 & 60.3 \\
-- & \checkmark & -- & 56.0 & 69.5 & 72.5 & 54.0 & 57.0 & 51.5 & 60.1 \\
-- & -- & \checkmark & 54.0 & 48.5 & 46.0 & 45.5 & 49.0 & 52.0 & 49.2 \\
\midrule
\multicolumn{3}{c|}{$\phi_H$} & \textcolor{darkgreen}{+10.2} & \textcolor{darkgreen}{+14.8} & \textcolor{darkgreen}{+18.2} & \textcolor{darkgreen}{+1.8} & \textcolor{darkgreen}{+8.3} & \textcolor{darkgreen}{+4.3} & \textcolor{darkgreen}{+9.6} \\
\multicolumn{3}{c|}{$\phi_P$} & \textcolor{darkgreen}{+9.8} & \textcolor{darkgreen}{+17.2} & \textcolor{darkgreen}{+18.0} & \textcolor{darkgreen}{+5.8} & \textcolor{darkgreen}{+6.1} & \textcolor{red}{-0.2} & \textcolor{darkgreen}{+9.5} \\
\multicolumn{3}{c|}{$\phi_S$} & \textcolor{darkgreen}{+1.0} & \textcolor{red}{-0.5} & \textcolor{red}{-0.7} & \textcolor{red}{-2.2} & \textcolor{red}{-0.4} & \textcolor{darkgreen}{+1.3} & \textcolor{red}{-0.3} \\
\bottomrule
\end{tabular}
}
\end{minipage}
&
\begin{minipage}[t]{0.47\textwidth}
\centering
\resizebox{\linewidth}{!}{%
\begin{tabular}{ccc|cccccc|c}
\toprule
H & P & S & Sent. & Spk. & Gen. & BgD. & BgR. & Room & Avg \\
\midrule
\checkmark & \checkmark & \checkmark & 73.5 & 80.0 & 87.0 & 53.5 & 56.5 & 67.5 & 69.7 \\
\checkmark & \checkmark & -- & 78.5 & 78.0 & 89.5 & 53.0 & 56.0 & 72.0 & 71.2 \\
\checkmark & -- & \checkmark & 66.0 & 70.5 & 71.0 & 55.5 & 46.5 & 52.5 & 60.3 \\
-- & \checkmark & \checkmark & 66.0 & 65.0 & 78.0 & 49.5 & 58.5 & 69.5 & 64.4 \\
\checkmark & -- & -- & 64.0 & 73.5 & 72.5 & 50.5 & 51.0 & 52.0 & 60.6 \\
-- & \checkmark & -- & 67.0 & 67.5 & 82.0 & 53.0 & 61.5 & 73.0 & 67.3 \\
-- & -- & \checkmark & 49.0 & 51.5 & 52.5 & 44.5 & 49.5 & 53.0 & 50.0 \\
\midrule
\multicolumn{3}{c|}{$\phi_H$} & \textcolor{darkgreen}{+11.9} & \textcolor{darkgreen}{+17.8} & \textcolor{darkgreen}{+14.8} & \textcolor{darkgreen}{+3.3} & \textcolor{red}{-1.8} & \textcolor{red}{-0.2} & \textcolor{darkgreen}{+7.6} \\
\multicolumn{3}{c|}{$\phi_P$} & \textcolor{darkgreen}{+13.4} & \textcolor{darkgreen}{+12.0} & \textcolor{darkgreen}{+23.1} & \textcolor{darkgreen}{+1.6} & \textcolor{darkgreen}{+9.5} & \textcolor{darkgreen}{+18.8} & \textcolor{darkgreen}{+13.1} \\
\multicolumn{3}{c|}{$\phi_S$} & \textcolor{red}{-1.8} & \textcolor{darkgreen}{+0.2} & \textcolor{red}{-0.9} & \textcolor{red}{-1.4} & \textcolor{red}{-1.2} & \textcolor{red}{-1.0} & \textcolor{red}{-1.0} \\
\bottomrule
\end{tabular}
}
\end{minipage}
\\[1.6ex]
\raisebox{-3em}{\rotatebox{90}{\textbf{Localized ($t=0.5$s)}}}
&
\begin{minipage}[t]{0.47\textwidth}
\centering
\resizebox{\linewidth}{!}{%
\begin{tabular}{ccc|cccccc|c}
\toprule
H & P & S & Sent. & Spk. & Gen. & BgD. & BgR. & Room & Avg \\
\midrule
\checkmark & \checkmark & \checkmark & 73.5 & 80.5 & 84.5 & 54.5 & 55.0 & 58.5 & 67.8 \\
\checkmark & \checkmark & -- & 75.0 & 81.5 & 85.5 & 57.0 & 55.0 & 58.5 & 68.8 \\
\checkmark & -- & \checkmark & 59.5 & 67.5 & 72.0 & 53.5 & 50.0 & 59.5 & 60.3 \\
-- & \checkmark & \checkmark & 64.5 & 68.5 & 80.0 & 52.5 & 46.5 & 53.0 & 60.8 \\
\checkmark & -- & -- & 61.0 & 66.0 & 71.5 & 56.5 & 54.0 & 60.0 & 61.5 \\
-- & \checkmark & -- & 62.0 & 71.0 & 80.0 & 52.0 & 47.0 & 52.0 & 60.7 \\
-- & -- & \checkmark & 54.0 & 43.0 & 46.2 & 48.2 & 47.8 & 47.2 & 47.8 \\
\midrule
\multicolumn{3}{c|}{$\phi_H$} & \textcolor{darkgreen}{+9.8} & \textcolor{darkgreen}{+15.2} & \textcolor{darkgreen}{+13.9} & \textcolor{darkgreen}{+4.5} & \textcolor{darkgreen}{+5.9} & \textcolor{darkgreen}{+8.3} & \textcolor{darkgreen}{+9.6} \\
\multicolumn{3}{c|}{$\phi_P$} & \textcolor{darkgreen}{+12.7} & \textcolor{darkgreen}{+18.2} & \textcolor{darkgreen}{+22.1} & \textcolor{darkgreen}{+1.8} & \textcolor{darkgreen}{+0.6} & \textcolor{darkgreen}{+1.0} & \textcolor{darkgreen}{+9.4} \\
\multicolumn{3}{c|}{$\phi_S$} & \textcolor{darkgreen}{+1.0} & \textcolor{red}{-2.8} & \textcolor{red}{-1.5} & \textcolor{red}{-1.8} & \textcolor{red}{-1.5} & \textcolor{red}{-0.8} & \textcolor{red}{-1.2} \\
\bottomrule
\end{tabular}
}
\end{minipage}
&
\begin{minipage}[t]{0.47\textwidth}
\centering
\resizebox{\linewidth}{!}{%
\begin{tabular}{ccc|cccccc|c}
\toprule
H & P & S & Sent. & Spk. & Gen. & BgD. & BgR. & Room & Avg \\
\midrule
\checkmark & \checkmark & \checkmark & 66.5 & 75.5 & 81.0 & 55.5 & 57.0 & 60.5 & 66.0 \\
\checkmark & \checkmark & -- & 73.0 & 80.0 & 84.0 & 54.0 & 62.0 & 63.5 & 69.4 \\
\checkmark & -- & \checkmark & 56.0 & 67.5 & 58.5 & 58.5 & 52.5 & 53.5 & 57.8 \\
-- & \checkmark & \checkmark & 62.0 & 66.5 & 83.0 & 57.0 & 53.5 & 62.0 & 64.0 \\
\checkmark & -- & -- & 54.0 & 69.5 & 59.5 & 55.5 & 52.5 & 56.0 & 57.8 \\
-- & \checkmark & -- & 68.5 & 70.5 & 86.0 & 58.0 & 56.5 & 62.0 & 66.9 \\
-- & -- & \checkmark & 46.0 & 42.5 & 42.5 & 52.8 & 47.0 & 57.0 & 48.0 \\
\midrule
\multicolumn{3}{c|}{$\phi_H$} & \textcolor{darkgreen}{+5.2} & \textcolor{darkgreen}{+15.2} & \textcolor{darkgreen}{+4.8} & \textcolor{darkgreen}{+1.6} & \textcolor{darkgreen}{+3.8} & \textcolor{darkgreen}{+1.2} & \textcolor{darkgreen}{+5.3} \\
\multicolumn{3}{c|}{$\phi_P$} & \textcolor{darkgreen}{+15.5} & \textcolor{darkgreen}{+15.2} & \textcolor{darkgreen}{+30.3} & \textcolor{darkgreen}{+2.1} & \textcolor{darkgreen}{+6.3} & \textcolor{darkgreen}{+8.4} & \textcolor{darkgreen}{+13.0} \\
\multicolumn{3}{c|}{$\phi_S$} & \textcolor{red}{-4.2} & \textcolor{red}{-5.0} & \textcolor{red}{-4.2} & \textcolor{darkgreen}{+1.7} & \textcolor{red}{-3.2} & \textcolor{darkgreen}{+0.9} & \textcolor{red}{-2.3} \\
\bottomrule
\end{tabular}
}
\end{minipage}
\\[2.0ex]
\multicolumn{3}{c}{\textbf{Llama-Mimi}}\\
\hline
& \textbf{Original} & \textbf{Normalized} \\
\raisebox{-1em}{\rotatebox{90}{\textbf{Global}}}
&
\begin{minipage}[t]{0.47\textwidth}
\centering
\resizebox{\linewidth}{!}{%
\begin{tabular}{cccc|cccccc|c}
\toprule
0 & 1 & 2 & 3 & Sent. & Spk. & Gen. & BgD. & BgR. & Room & Avg \\
\midrule
$\checkmark$ & $\checkmark$ & $\checkmark$ & $\checkmark$ & 79.5 & 85.5 & 82.0 & 75.0 & 72.0 & 92.0 & 81.0 \\
$\checkmark$ & -- & -- & -- & 57.0 & 71.0 & 65.0 & 67.5 & 67.5 & 71.5 & 66.6 \\
-- & $\checkmark$ & -- & -- & 60.5 & 70.0 & 73.5 & 64.5 & 62.5 & 71.5 & 67.1 \\
-- & -- & $\checkmark$ & -- & 75.0 & 82.5 & 84.5 & 68.0 & 67.0 & 91.0 & 78.0 \\
-- & -- & -- & $\checkmark$ & 62.5 & 78.5 & 77.0 & 67.0 & 67.0 & 81.5 & 72.2 \\
$\checkmark$ & $\checkmark$ & -- & -- & 72.5 & 76.5 & 75.5 & 67.0 & 68.5 & 75.0 & 72.5 \\
-- & $\checkmark$ & $\checkmark$ & -- & 76.0 & 80.0 & 84.0 & 70.5 & 69.5 & 90.5 & 78.4 \\
-- & -- & $\checkmark$ & $\checkmark$ & 71.0 & 84.5 & 85.5 & 72.5 & 73.0 & 96.0 & 80.4 \\
$\checkmark$ & -- & -- & $\checkmark$ & 61.0 & 79.0 & 73.5 & 73.5 & 74.0 & 79.5 & 73.4 \\
$\checkmark$ & -- & $\checkmark$ & -- & 70.0 & 78.5 & 80.5 & 71.5 & 70.0 & 92.5 & 77.2 \\
-- & $\checkmark$ & -- & $\checkmark$ & 76.0 & 80.5 & 81.0 & 73.0 & 70.5 & 80.5 & 76.9 \\
$\checkmark$ & $\checkmark$ & $\checkmark$ & -- & 78.0 & 82.0 & 80.0 & 73.5 & 72.0 & 91.5 & 79.5 \\
$\checkmark$ & $\checkmark$ & -- & $\checkmark$ & 74.0 & 79.5 & 77.5 & 71.0 & 71.0 & 80.5 & 75.6 \\
$\checkmark$ & -- & $\checkmark$ & $\checkmark$ & 67.0 & 83.0 & 80.5 & 73.5 & 72.5 & 95.5 & 78.7 \\
-- & $\checkmark$ & $\checkmark$ & $\checkmark$ & 80.5 & 85.5 & 85.0 & 71.0 & 72.5 & 93.0 & 81.2 \\
\midrule
\multicolumn{4}{c|}{$\phi_0$} & \textcolor{darkgreen}{+1.6} & \textcolor{darkgreen}{+5.5} & \textcolor{darkgreen}{+1.5} & \textcolor{darkgreen}{+6.6} & \textcolor{darkgreen}{+5.8} & \textcolor{darkgreen}{+5.4} & \textcolor{darkgreen}{+4.4} \\
\multicolumn{4}{c|}{$\phi_1$} & \textcolor{darkgreen}{+10.8} & \textcolor{darkgreen}{+6.5} & \textcolor{darkgreen}{+7.7} & \textcolor{darkgreen}{+4.5} & \textcolor{darkgreen}{+3.5} & \textcolor{darkgreen}{+4.4} & \textcolor{darkgreen}{+6.2} \\
\multicolumn{4}{c|}{$\phi_2$} & \textcolor{darkgreen}{+12.0} & \textcolor{darkgreen}{+12.8} & \textcolor{darkgreen}{+13.9} & \textcolor{darkgreen}{+7.2} & \textcolor{darkgreen}{+6.1} & \textcolor{darkgreen}{+21.4} & \textcolor{darkgreen}{+12.2} \\
\multicolumn{4}{c|}{$\phi_3$} & \textcolor{darkgreen}{+5.0} & \textcolor{darkgreen}{+10.8} & \textcolor{darkgreen}{+8.9} & \textcolor{darkgreen}{+6.8} & \textcolor{darkgreen}{+6.6} & \textcolor{darkgreen}{+10.8} & \textcolor{darkgreen}{+8.1} \\
\bottomrule
\end{tabular}
}
\end{minipage}
&
\begin{minipage}[t]{0.47\textwidth}
\centering
\resizebox{\linewidth}{!}{%
\begin{tabular}{cccc|cccccc|c}
\toprule
0 & 1 & 2 & 3 & Sent. & Spk. & Gen. & BgD. & BgR. & Room & Avg \\
\midrule
$\checkmark$ & $\checkmark$ & $\checkmark$ & $\checkmark$ & 89.0 & 95.0 & 99.5 & 73.5 & 87.0 & 98.5 & 90.4 \\
$\checkmark$ & -- & -- & -- & 55.5 & 69.0 & 73.5 & 60.0 & 62.0 & 75.5 & 65.9 \\
-- & $\checkmark$ & -- & -- & 75.5 & 72.5 & 77.5 & 54.5 & 67.0 & 80.0 & 71.2 \\
-- & -- & $\checkmark$ & -- & 84.5 & 89.5 & 93.0 & 67.5 & 70.0 & 86.5 & 81.8 \\
-- & -- & -- & $\checkmark$ & 72.0 & 78.0 & 82.5 & 64.5 & 67.5 & 83.5 & 74.7 \\
$\checkmark$ & $\checkmark$ & -- & -- & 72.5 & 78.5 & 84.5 & 60.5 & 72.0 & 81.5 & 74.9 \\
-- & $\checkmark$ & $\checkmark$ & -- & 90.5 & 93.0 & 98.0 & 65.0 & 79.5 & 94.0 & 86.7 \\
-- & -- & $\checkmark$ & $\checkmark$ & 84.5 & 92.5 & 96.5 & 70.5 & 77.0 & 92.5 & 85.6 \\
$\checkmark$ & -- & -- & $\checkmark$ & 72.5 & 83.0 & 88.0 & 66.0 & 74.5 & 87.0 & 78.5 \\
$\checkmark$ & -- & $\checkmark$ & -- & 77.5 & 90.5 & 93.5 & 67.5 & 73.5 & 89.0 & 81.9 \\
-- & $\checkmark$ & -- & $\checkmark$ & 82.5 & 86.0 & 89.5 & 63.0 & 75.5 & 87.5 & 80.7 \\
$\checkmark$ & $\checkmark$ & $\checkmark$ & -- & 83.5 & 92.5 & 95.5 & 69.5 & 80.5 & 95.0 & 86.1 \\
$\checkmark$ & $\checkmark$ & -- & $\checkmark$ & 79.5 & 86.0 & 92.5 & 65.0 & 79.0 & 91.0 & 82.2 \\
$\checkmark$ & -- & $\checkmark$ & $\checkmark$ & 82.0 & 92.0 & 95.0 & 71.5 & 80.0 & 94.5 & 85.8 \\
-- & $\checkmark$ & $\checkmark$ & $\checkmark$ & 89.5 & 93.5 & 98.0 & 71.5 & 83.0 & 96.0 & 88.6 \\
\midrule
\multicolumn{4}{c|}{$\phi_0$} & \textcolor{red}{-0.6} & \textcolor{darkgreen}{+6.0} & \textcolor{darkgreen}{+7.2} & \textcolor{darkgreen}{+4.2} & \textcolor{darkgreen}{+5.9} & \textcolor{darkgreen}{+8.2} & \textcolor{darkgreen}{+5.2} \\
\multicolumn{4}{c|}{$\phi_1$} & \textcolor{darkgreen}{+12.4} & \textcolor{darkgreen}{+8.6} & \textcolor{darkgreen}{+10.6} & \textcolor{darkgreen}{+1.5} & \textcolor{darkgreen}{+9.8} & \textcolor{darkgreen}{+11.1} & \textcolor{darkgreen}{+9.0} \\
\multicolumn{4}{c|}{$\phi_2$} & \textcolor{darkgreen}{+17.4} & \textcolor{darkgreen}{+19.4} & \textcolor{darkgreen}{+19.2} & \textcolor{darkgreen}{+10.4} & \textcolor{darkgreen}{+11.6} & \textcolor{darkgreen}{+16.5} & \textcolor{darkgreen}{+15.8} \\
\multicolumn{4}{c|}{$\phi_3$} & \textcolor{darkgreen}{+9.7} & \textcolor{darkgreen}{+11.0} & \textcolor{darkgreen}{+12.4} & \textcolor{darkgreen}{+7.3} & \textcolor{darkgreen}{+9.7} & \textcolor{darkgreen}{+12.8} & \textcolor{darkgreen}{+10.5} \\
\bottomrule
\end{tabular}
}
\end{minipage}
\\[1.4ex]
\smash{\raisebox{-0.5\height}{\rotatebox{90}{\textbf{Localized ($t=0.5$s)}}}}
&
\begin{minipage}[t]{0.47\textwidth}
\centering
\resizebox{\linewidth}{!}{%
\begin{tabular}{cccc|cccccc|c}
\toprule
0 & 1 & 2 & 3 & Sent. & Spk. & Gen. & BgD. & BgR. & Room & Avg \\
\midrule
$\checkmark$ & $\checkmark$ & $\checkmark$ & $\checkmark$ & 95.5 & 95.5 & 100.0 & 79.5 & 84.5 & 97.5 & 92.1 \\
$\checkmark$ & -- & -- & -- & 75.5 & 83.5 & 85.0 & 59.5 & 64.8 & 73.0 & 73.5 \\
-- & $\checkmark$ & -- & -- & 78.0 & 81.5 & 85.0 & 59.0 & 68.0 & 80.0 & 75.2 \\
-- & -- & $\checkmark$ & -- & 88.5 & 91.0 & 93.5 & 73.0 & 80.0 & 92.5 & 86.4 \\
-- & -- & -- & $\checkmark$ & 75.0 & 84.0 & 94.0 & 70.5 & 75.5 & 80.0 & 79.8 \\
$\checkmark$ & $\checkmark$ & -- & -- & 85.5 & 90.0 & 92.5 & 67.0 & 74.0 & 87.5 & 82.8 \\
-- & $\checkmark$ & $\checkmark$ & -- & 90.5 & 92.5 & 95.0 & 73.5 & 80.5 & 94.5 & 87.8 \\
-- & -- & $\checkmark$ & $\checkmark$ & 92.0 & 91.5 & 98.0 & 76.5 & 83.5 & 94.5 & 89.3 \\
$\checkmark$ & -- & -- & $\checkmark$ & 83.0 & 91.5 & 95.0 & 73.5 & 79.2 & 82.5 & 84.1 \\
$\checkmark$ & -- & $\checkmark$ & -- & 89.5 & 92.5 & 97.5 & 74.5 & 80.0 & 91.0 & 87.5 \\
-- & $\checkmark$ & -- & $\checkmark$ & 87.5 & 90.0 & 96.5 & 75.0 & 77.5 & 89.0 & 85.9 \\
$\checkmark$ & $\checkmark$ & $\checkmark$ & -- & 94.0 & 93.5 & 99.0 & 74.0 & 81.0 & 97.5 & 89.8 \\
$\checkmark$ & $\checkmark$ & -- & $\checkmark$ & 90.0 & 96.0 & 98.5 & 75.0 & 81.5 & 91.0 & 88.7 \\
$\checkmark$ & -- & $\checkmark$ & $\checkmark$ & 94.0 & 94.5 & 99.5 & 79.5 & 82.5 & 92.5 & 90.4 \\
-- & $\checkmark$ & $\checkmark$ & $\checkmark$ & 94.5 & 95.5 & 99.0 & 78.5 & 84.5 & 97.5 & 91.6 \\
\midrule
\multicolumn{4}{c|}{$\phi_0$} & \textcolor{darkgreen}{+8.7} & \textcolor{darkgreen}{+10.7} & \textcolor{darkgreen}{+10.7} & \textcolor{darkgreen}{+4.0} & \textcolor{darkgreen}{+4.8} & \textcolor{darkgreen}{+6.7} & \textcolor{darkgreen}{+7.6} \\
\multicolumn{4}{c|}{$\phi_1$} & \textcolor{darkgreen}{+10.6} & \textcolor{darkgreen}{+10.1} & \textcolor{darkgreen}{+10.3} & \textcolor{darkgreen}{+3.5} & \textcolor{darkgreen}{+6.3} & \textcolor{darkgreen}{+12.4} & \textcolor{darkgreen}{+8.9} \\
\multicolumn{4}{c|}{$\phi_2$} & \textcolor{darkgreen}{+16.8} & \textcolor{darkgreen}{+13.4} & \textcolor{darkgreen}{+14.6} & \textcolor{darkgreen}{+11.2} & \textcolor{darkgreen}{+12.7} & \textcolor{darkgreen}{+18.5} & \textcolor{darkgreen}{+14.5} \\
\multicolumn{4}{c|}{$\phi_3$} & \textcolor{darkgreen}{+9.4} & \textcolor{darkgreen}{+11.3} & \textcolor{darkgreen}{+14.4} & \textcolor{darkgreen}{+10.8} & \textcolor{darkgreen}{+10.7} & \textcolor{darkgreen}{+9.9} & \textcolor{darkgreen}{+11.1} \\
\bottomrule
\end{tabular}
}
\end{minipage}
&
\begin{minipage}[t]{0.47\textwidth}
\centering
\resizebox{\linewidth}{!}{%
\begin{tabular}{cccc|cccccc|c}
\toprule
0 & 1 & 2 & 3 & Sent. & Spk. & Gen. & BgD. & BgR. & Room & Avg \\
\midrule
$\checkmark$ & $\checkmark$ & $\checkmark$ & $\checkmark$ & 92.0 & 98.0 & 99.0 & 81.0 & 85.0 & 97.0 & 92.0 \\
$\checkmark$ & -- & -- & -- & 64.0 & 76.0 & 86.0 & 64.0 & 66.0 & 71.5 & 71.2 \\
-- & $\checkmark$ & -- & -- & 75.0 & 75.5 & 83.5 & 53.0 & 61.5 & 74.0 & 70.4 \\
-- & -- & $\checkmark$ & -- & 84.5 & 89.5 & 91.0 & 67.0 & 83.5 & 89.5 & 84.2 \\
-- & -- & -- & $\checkmark$ & 75.0 & 80.0 & 88.0 & 72.0 & 67.5 & 75.5 & 76.3 \\
$\checkmark$ & $\checkmark$ & -- & -- & 76.5 & 85.5 & 90.5 & 66.5 & 66.0 & 83.5 & 78.1 \\
-- & $\checkmark$ & $\checkmark$ & -- & 91.5 & 94.0 & 95.5 & 72.5 & 81.0 & 94.5 & 88.2 \\
-- & -- & $\checkmark$ & $\checkmark$ & 86.5 & 94.0 & 96.5 & 77.0 & 79.5 & 93.0 & 87.8 \\
$\checkmark$ & -- & -- & $\checkmark$ & 76.0 & 87.5 & 94.0 & 75.0 & 68.5 & 84.5 & 80.9 \\
$\checkmark$ & -- & $\checkmark$ & -- & 83.5 & 92.0 & 94.0 & 74.5 & 83.0 & 90.5 & 86.2 \\
-- & $\checkmark$ & -- & $\checkmark$ & 85.0 & 86.0 & 94.0 & 67.0 & 68.5 & 85.0 & 80.9 \\
$\checkmark$ & $\checkmark$ & $\checkmark$ & -- & 90.0 & 94.0 & 98.5 & 76.0 & 84.5 & 95.0 & 89.7 \\
$\checkmark$ & $\checkmark$ & -- & $\checkmark$ & 86.5 & 91.0 & 98.0 & 72.0 & 72.0 & 90.5 & 85.0 \\
$\checkmark$ & -- & $\checkmark$ & $\checkmark$ & 84.5 & 95.5 & 97.5 & 83.0 & 85.5 & 94.0 & 90.0 \\
-- & $\checkmark$ & $\checkmark$ & $\checkmark$ & 92.0 & 95.5 & 98.5 & 74.0 & 79.5 & 96.0 & 89.2 \\
\midrule
\multicolumn{4}{c|}{$\phi_0$} & \textcolor{darkgreen}{+3.5} & \textcolor{darkgreen}{+9.3} & \textcolor{darkgreen}{+11.1} & \textcolor{darkgreen}{+8.5} & \textcolor{darkgreen}{+6.9} & \textcolor{darkgreen}{+7.8} & \textcolor{darkgreen}{+7.8} \\
\multicolumn{4}{c|}{$\phi_1$} & \textcolor{darkgreen}{+12.5} & \textcolor{darkgreen}{+9.2} & \textcolor{darkgreen}{+10.9} & \textcolor{darkgreen}{+0.1} & \textcolor{darkgreen}{+3.0} & \textcolor{darkgreen}{+10.1} & \textcolor{darkgreen}{+7.6} \\
\multicolumn{4}{c|}{$\phi_2$} & \textcolor{darkgreen}{+16.4} & \textcolor{darkgreen}{+17.8} & \textcolor{darkgreen}{+14.2} & \textcolor{darkgreen}{+11.5} & \textcolor{darkgreen}{+19.5} & \textcolor{darkgreen}{+18.9} & \textcolor{darkgreen}{+16.4} \\
\multicolumn{4}{c|}{$\phi_3$} & \textcolor{darkgreen}{+9.7} & \textcolor{darkgreen}{+11.6} & \textcolor{darkgreen}{+12.8} & \textcolor{darkgreen}{+11.0} & \textcolor{darkgreen}{+5.5} & \textcolor{darkgreen}{+10.2} & \textcolor{darkgreen}{+10.1} \\
\bottomrule
\end{tabular}
}
\end{minipage}

\end{tabular}
\\[1.4ex]
\end{table*}
}

\newcommand{\TabCompositionComprehensiveLlamaMimiOnly}{%
\begin{table*}[t]
\centering
\renewcommand{\arraystretch}{1.05}
\setlength{\tabcolsep}{4pt}

\begin{tabular}{@{}c cc@{}}
\hline
& \textbf{Original} & \textbf{Normalized} \\
\rotatebox{90}{\textbf{Global}}
&
\begin{minipage}[t]{0.47\textwidth}
\centering
\resizebox{\linewidth}{!}{%
\begin{tabular}{cccc|cccccc|c}
\toprule
0 & 1 & 2 & 3 & Sent. & Spk. & Gen. & BgD. & BgR. & Room & Avg \\
\midrule
$\checkmark$ & $\checkmark$ & $\checkmark$ & $\checkmark$ & 79.5 & 85.5 & 82.0 & 75.0 & 72.0 & 92.0 & 81.0 \\
$\checkmark$ & -- & -- & -- & 57.0 & 71.0 & 65.0 & 67.5 & 67.5 & 71.5 & 66.6 \\
-- & $\checkmark$ & -- & -- & 60.5 & 70.0 & 73.5 & 64.5 & 62.5 & 71.5 & 67.1 \\
-- & -- & $\checkmark$ & -- & 75.0 & 82.5 & 84.5 & 68.0 & 67.0 & 91.0 & 78.0 \\
-- & -- & -- & $\checkmark$ & 62.5 & 78.5 & 77.0 & 67.0 & 67.0 & 81.5 & 72.2 \\
$\checkmark$ & $\checkmark$ & -- & -- & 72.5 & 76.5 & 75.5 & 67.0 & 68.5 & 75.0 & 72.5 \\
-- & $\checkmark$ & $\checkmark$ & -- & 76.0 & 80.0 & 84.0 & 70.5 & 69.5 & 90.5 & 78.4 \\
-- & -- & $\checkmark$ & $\checkmark$ & 71.0 & 84.5 & 85.5 & 72.5 & 73.0 & 96.0 & 80.4 \\
$\checkmark$ & -- & -- & $\checkmark$ & 61.0 & 79.0 & 73.5 & 73.5 & 74.0 & 79.5 & 73.4 \\
$\checkmark$ & -- & $\checkmark$ & -- & 70.0 & 78.5 & 80.5 & 71.5 & 70.0 & 92.5 & 77.2 \\
-- & $\checkmark$ & -- & $\checkmark$ & 76.0 & 80.5 & 81.0 & 73.0 & 70.5 & 80.5 & 76.9 \\
$\checkmark$ & $\checkmark$ & $\checkmark$ & -- & 78.0 & 82.0 & 80.0 & 73.5 & 72.0 & 91.5 & 79.5 \\
$\checkmark$ & $\checkmark$ & -- & $\checkmark$ & 74.0 & 79.5 & 77.5 & 71.0 & 71.0 & 80.5 & 75.6 \\
$\checkmark$ & -- & $\checkmark$ & $\checkmark$ & 67.0 & 83.0 & 80.5 & 73.5 & 72.5 & 95.5 & 78.7 \\
-- & $\checkmark$ & $\checkmark$ & $\checkmark$ & 80.5 & 85.5 & 85.0 & 71.0 & 72.5 & 93.0 & 81.2 \\
\midrule
\multicolumn{4}{c|}{$\phi_0$} & \textcolor{darkgreen}{+1.6} & \textcolor{darkgreen}{+5.5} & \textcolor{darkgreen}{+1.5} & \textcolor{darkgreen}{+6.6} & \textcolor{darkgreen}{+5.8} & \textcolor{darkgreen}{+5.4} & \textcolor{darkgreen}{+4.4} \\
\multicolumn{4}{c|}{$\phi_1$} & \textcolor{darkgreen}{+10.8} & \textcolor{darkgreen}{+6.5} & \textcolor{darkgreen}{+7.7} & \textcolor{darkgreen}{+4.5} & \textcolor{darkgreen}{+3.5} & \textcolor{darkgreen}{+4.4} & \textcolor{darkgreen}{+6.2} \\
\multicolumn{4}{c|}{$\phi_2$} & \textcolor{darkgreen}{+12.0} & \textcolor{darkgreen}{+12.8} & \textcolor{darkgreen}{+13.9} & \textcolor{darkgreen}{+7.2} & \textcolor{darkgreen}{+6.1} & \textcolor{darkgreen}{+21.4} & \textcolor{darkgreen}{+12.2} \\
\multicolumn{4}{c|}{$\phi_3$} & \textcolor{darkgreen}{+5.0} & \textcolor{darkgreen}{+10.8} & \textcolor{darkgreen}{+8.9} & \textcolor{darkgreen}{+6.8} & \textcolor{darkgreen}{+6.6} & \textcolor{darkgreen}{+10.8} & \textcolor{darkgreen}{+8.1} \\
\bottomrule
\end{tabular}
}
\end{minipage}
&
\begin{minipage}[t]{0.47\textwidth}
\centering
\resizebox{\linewidth}{!}{%
\begin{tabular}{cccc|cccccc|c}
\toprule
0 & 1 & 2 & 3 & Sent. & Spk. & Gen. & BgD. & BgR. & Room & Avg \\
\midrule
$\checkmark$ & $\checkmark$ & $\checkmark$ & $\checkmark$ & 89.0 & 95.0 & 99.5 & 73.5 & 87.0 & 98.5 & 90.4 \\
$\checkmark$ & -- & -- & -- & 55.5 & 69.0 & 73.5 & 60.0 & 62.0 & 75.5 & 65.9 \\
-- & $\checkmark$ & -- & -- & 75.5 & 72.5 & 77.5 & 54.5 & 67.0 & 80.0 & 71.2 \\
-- & -- & $\checkmark$ & -- & 84.5 & 89.5 & 93.0 & 67.5 & 70.0 & 86.5 & 81.8 \\
-- & -- & -- & $\checkmark$ & 72.0 & 78.0 & 82.5 & 64.5 & 67.5 & 83.5 & 74.7 \\
$\checkmark$ & $\checkmark$ & -- & -- & 72.5 & 78.5 & 84.5 & 60.5 & 72.0 & 81.5 & 74.9 \\
-- & $\checkmark$ & $\checkmark$ & -- & 90.5 & 93.0 & 98.0 & 65.0 & 79.5 & 94.0 & 86.7 \\
-- & -- & $\checkmark$ & $\checkmark$ & 84.5 & 92.5 & 96.5 & 70.5 & 77.0 & 92.5 & 85.6 \\
$\checkmark$ & -- & -- & $\checkmark$ & 72.5 & 83.0 & 88.0 & 66.0 & 74.5 & 87.0 & 78.5 \\
$\checkmark$ & -- & $\checkmark$ & -- & 77.5 & 90.5 & 93.5 & 67.5 & 73.5 & 89.0 & 81.9 \\
-- & $\checkmark$ & -- & $\checkmark$ & 82.5 & 86.0 & 89.5 & 63.0 & 75.5 & 87.5 & 80.7 \\
$\checkmark$ & $\checkmark$ & $\checkmark$ & -- & 83.5 & 92.5 & 95.5 & 69.5 & 80.5 & 95.0 & 86.1 \\
$\checkmark$ & $\checkmark$ & -- & $\checkmark$ & 79.5 & 86.0 & 92.5 & 65.0 & 79.0 & 91.0 & 82.2 \\
$\checkmark$ & -- & $\checkmark$ & $\checkmark$ & 82.0 & 92.0 & 95.0 & 71.5 & 80.0 & 94.5 & 85.8 \\
-- & $\checkmark$ & $\checkmark$ & $\checkmark$ & 89.5 & 93.5 & 98.0 & 71.5 & 83.0 & 96.0 & 88.6 \\
\midrule
\multicolumn{4}{c|}{$\phi_0$} & \textcolor{red}{-0.6} & \textcolor{darkgreen}{+6.0} & \textcolor{darkgreen}{+7.2} & \textcolor{darkgreen}{+4.2} & \textcolor{darkgreen}{+5.9} & \textcolor{darkgreen}{+8.2} & \textcolor{darkgreen}{+5.2} \\
\multicolumn{4}{c|}{$\phi_1$} & \textcolor{darkgreen}{+12.4} & \textcolor{darkgreen}{+8.6} & \textcolor{darkgreen}{+10.6} & \textcolor{darkgreen}{+1.5} & \textcolor{darkgreen}{+9.8} & \textcolor{darkgreen}{+11.1} & \textcolor{darkgreen}{+9.0} \\
\multicolumn{4}{c|}{$\phi_2$} & \textcolor{darkgreen}{+17.4} & \textcolor{darkgreen}{+19.4} & \textcolor{darkgreen}{+19.2} & \textcolor{darkgreen}{+10.4} & \textcolor{darkgreen}{+11.6} & \textcolor{darkgreen}{+16.5} & \textcolor{darkgreen}{+15.8} \\
\multicolumn{4}{c|}{$\phi_3$} & \textcolor{darkgreen}{+9.7} & \textcolor{darkgreen}{+11.0} & \textcolor{darkgreen}{+12.4} & \textcolor{darkgreen}{+7.3} & \textcolor{darkgreen}{+9.7} & \textcolor{darkgreen}{+12.8} & \textcolor{darkgreen}{+10.5} \\
\bottomrule
\end{tabular}
}
\end{minipage}
\\[1.6ex]
\rotatebox{90}{\textbf{Local ($t=0.5$s)}}
&
\begin{minipage}[t]{0.47\textwidth}
\centering
\resizebox{\linewidth}{!}{%
\begin{tabular}{cccc|cccccc|c}
\toprule
0 & 1 & 2 & 3 & Sent. & Spk. & Gen. & BgD. & BgR. & Room & Avg \\
\midrule
$\checkmark$ & $\checkmark$ & $\checkmark$ & $\checkmark$ & 95.5 & 95.5 & 100.0 & 79.5 & 84.5 & 97.5 & 92.1 \\
$\checkmark$ & -- & -- & -- & 75.5 & 83.5 & 85.0 & 59.5 & 64.8 & 73.0 & 73.5 \\
-- & $\checkmark$ & -- & -- & 78.0 & 81.5 & 85.0 & 59.0 & 68.0 & 80.0 & 75.2 \\
-- & -- & $\checkmark$ & -- & 88.5 & 91.0 & 93.5 & 73.0 & 80.0 & 92.5 & 86.4 \\
-- & -- & -- & $\checkmark$ & 75.0 & 84.0 & 94.0 & 70.5 & 75.5 & 80.0 & 79.8 \\
$\checkmark$ & $\checkmark$ & -- & -- & 85.5 & 90.0 & 92.5 & 67.0 & 74.0 & 87.5 & 82.8 \\
-- & $\checkmark$ & $\checkmark$ & -- & 90.5 & 92.5 & 95.0 & 73.5 & 80.5 & 94.5 & 87.8 \\
-- & -- & $\checkmark$ & $\checkmark$ & 92.0 & 91.5 & 98.0 & 76.5 & 83.5 & 94.5 & 89.3 \\
$\checkmark$ & -- & -- & $\checkmark$ & 83.0 & 91.5 & 95.0 & 73.5 & 79.2 & 82.5 & 84.1 \\
$\checkmark$ & -- & $\checkmark$ & -- & 89.5 & 92.5 & 97.5 & 74.5 & 80.0 & 91.0 & 87.5 \\
-- & $\checkmark$ & -- & $\checkmark$ & 87.5 & 90.0 & 96.5 & 75.0 & 77.5 & 89.0 & 85.9 \\
$\checkmark$ & $\checkmark$ & $\checkmark$ & -- & 94.0 & 93.5 & 99.0 & 74.0 & 81.0 & 97.5 & 89.8 \\
$\checkmark$ & $\checkmark$ & -- & $\checkmark$ & 90.0 & 96.0 & 98.5 & 75.0 & 81.5 & 91.0 & 88.7 \\
$\checkmark$ & -- & $\checkmark$ & $\checkmark$ & 94.0 & 94.5 & 99.5 & 79.5 & 82.5 & 92.5 & 90.4 \\
-- & $\checkmark$ & $\checkmark$ & $\checkmark$ & 94.5 & 95.5 & 99.0 & 78.5 & 84.5 & 97.5 & 91.6 \\
\midrule
\multicolumn{4}{c|}{$\phi_0$} & \textcolor{darkgreen}{+8.7} & \textcolor{darkgreen}{+10.7} & \textcolor{darkgreen}{+10.7} & \textcolor{darkgreen}{+4.0} & \textcolor{darkgreen}{+4.8} & \textcolor{darkgreen}{+6.7} & \textcolor{darkgreen}{+7.6} \\
\multicolumn{4}{c|}{$\phi_1$} & \textcolor{darkgreen}{+10.6} & \textcolor{darkgreen}{+10.1} & \textcolor{darkgreen}{+10.3} & \textcolor{darkgreen}{+3.5} & \textcolor{darkgreen}{+6.3} & \textcolor{darkgreen}{+12.4} & \textcolor{darkgreen}{+8.9} \\
\multicolumn{4}{c|}{$\phi_2$} & \textcolor{darkgreen}{+16.8} & \textcolor{darkgreen}{+13.4} & \textcolor{darkgreen}{+14.6} & \textcolor{darkgreen}{+11.2} & \textcolor{darkgreen}{+12.7} & \textcolor{darkgreen}{+18.5} & \textcolor{darkgreen}{+14.5} \\
\multicolumn{4}{c|}{$\phi_3$} & \textcolor{darkgreen}{+9.4} & \textcolor{darkgreen}{+11.3} & \textcolor{darkgreen}{+14.4} & \textcolor{darkgreen}{+10.8} & \textcolor{darkgreen}{+10.7} & \textcolor{darkgreen}{+9.9} & \textcolor{darkgreen}{+11.1} \\
\bottomrule
\end{tabular}
}
\end{minipage}
&
\begin{minipage}[t]{0.47\textwidth}
\centering
\resizebox{\linewidth}{!}{%
\begin{tabular}{cccc|cccccc|c}
\toprule
0 & 1 & 2 & 3 & Sent. & Spk. & Gen. & BgD. & BgR. & Room & Avg \\
\midrule
$\checkmark$ & $\checkmark$ & $\checkmark$ & $\checkmark$ & 92.0 & 98.0 & 99.0 & 81.0 & 85.0 & 97.0 & 92.0 \\
$\checkmark$ & -- & -- & -- & 64.0 & 76.0 & 86.0 & 64.0 & 66.0 & 71.5 & 71.2 \\
-- & $\checkmark$ & -- & -- & 75.0 & 75.5 & 83.5 & 53.0 & 61.5 & 74.0 & 70.4 \\
-- & -- & $\checkmark$ & -- & 84.5 & 89.5 & 91.0 & 67.0 & 83.5 & 89.5 & 84.2 \\
-- & -- & -- & $\checkmark$ & 75.0 & 80.0 & 88.0 & 72.0 & 67.5 & 75.5 & 76.3 \\
$\checkmark$ & $\checkmark$ & -- & -- & 76.5 & 85.5 & 90.5 & 66.5 & 66.0 & 83.5 & 78.1 \\
-- & $\checkmark$ & $\checkmark$ & -- & 91.5 & 94.0 & 95.5 & 72.5 & 81.0 & 94.5 & 88.2 \\
-- & -- & $\checkmark$ & $\checkmark$ & 86.5 & 94.0 & 96.5 & 77.0 & 79.5 & 93.0 & 87.8 \\
$\checkmark$ & -- & -- & $\checkmark$ & 76.0 & 87.5 & 94.0 & 75.0 & 68.5 & 84.5 & 80.9 \\
$\checkmark$ & -- & $\checkmark$ & -- & 83.5 & 92.0 & 94.0 & 74.5 & 83.0 & 90.5 & 86.2 \\
-- & $\checkmark$ & -- & $\checkmark$ & 85.0 & 86.0 & 94.0 & 67.0 & 68.5 & 85.0 & 80.9 \\
$\checkmark$ & $\checkmark$ & $\checkmark$ & -- & 90.0 & 94.0 & 98.5 & 76.0 & 84.5 & 95.0 & 89.7 \\
$\checkmark$ & $\checkmark$ & -- & $\checkmark$ & 86.5 & 91.0 & 98.0 & 72.0 & 72.0 & 90.5 & 85.0 \\
$\checkmark$ & -- & $\checkmark$ & $\checkmark$ & 84.5 & 95.5 & 97.5 & 83.0 & 85.5 & 94.0 & 90.0 \\
-- & $\checkmark$ & $\checkmark$ & $\checkmark$ & 92.0 & 95.5 & 98.5 & 74.0 & 79.5 & 96.0 & 89.2 \\
\midrule
\multicolumn{4}{c|}{$\phi_0$} & \textcolor{darkgreen}{+3.5} & \textcolor{darkgreen}{+9.3} & \textcolor{darkgreen}{+11.1} & \textcolor{darkgreen}{+8.5} & \textcolor{darkgreen}{+6.9} & \textcolor{darkgreen}{+7.8} & \textcolor{darkgreen}{+7.8} \\
\multicolumn{4}{c|}{$\phi_1$} & \textcolor{darkgreen}{+12.5} & \textcolor{darkgreen}{+9.2} & \textcolor{darkgreen}{+10.9} & \textcolor{darkgreen}{+0.1} & \textcolor{darkgreen}{+3.0} & \textcolor{darkgreen}{+10.1} & \textcolor{darkgreen}{+7.6} \\
\multicolumn{4}{c|}{$\phi_2$} & \textcolor{darkgreen}{+16.4} & \textcolor{darkgreen}{+17.8} & \textcolor{darkgreen}{+14.2} & \textcolor{darkgreen}{+11.5} & \textcolor{darkgreen}{+19.5} & \textcolor{darkgreen}{+18.9} & \textcolor{darkgreen}{+16.4} \\
\multicolumn{4}{c|}{$\phi_3$} & \textcolor{darkgreen}{+9.7} & \textcolor{darkgreen}{+11.6} & \textcolor{darkgreen}{+12.8} & \textcolor{darkgreen}{+11.0} & \textcolor{darkgreen}{+5.5} & \textcolor{darkgreen}{+10.2} & \textcolor{darkgreen}{+10.1} \\
\bottomrule
\end{tabular}
}
\end{minipage}

\end{tabular}

\caption{Llama-Mimi ablation as a 2$\times$2 index: columns are \textbf{original} vs.\ \textbf{normalized}; rows are \textbf{global} (top, long window) vs.\ \textbf{local} ($t=0.5$s, bottom). Shapley baseline assumes $0000=50\%$. [20260416]}
\label{tab:composition-comprehensive-llama-mimi-only}
\end{table*}
}

\newcommand{\TabCompositionShapleyAvgOnly}{%
\begin{table}[t]
\caption{Shapley value decompositions for Spirit-LM-Expressive and Llama-Mimi over token types (HuBERT $\Phi_H$, pitch $\Phi_P$, style $\Phi_S$) and layer groups ($\Phi_0$--$\Phi_3$). The Shapley values of the primary tokens for the two models ($\Phi_H$ and $\Phi_0$) shift in opposite directions under localization and normalization.}
\label{tab:composition-shapley-avg-only}
\small
\centering
\setlength{\tabcolsep}{4pt}

\begin{tabular}{l l l c c}
\toprule
\textbf{Model} & \textbf{Window} & \textbf{Term} & \textbf{Original} & \textbf{Norm.} \\
\midrule
\multirow{6}{*}{\textbf{Spirit-LM Expr.}}
& \multirow{3}{*}{Global}
& $\phi_H$ & \textcolor{darkgreen}{+9.6} & \textcolor{darkgreen}{+7.6} \\
& & $\phi_P$ & \textcolor{darkgreen}{+9.5} & \textcolor{darkgreen}{+13.1} \\
& & $\phi_S$ & \textcolor{red}{-0.3} & \textcolor{red}{-1.0} \\
\cmidrule(lr){2-5}
& \multirow{3}{*}{\shortstack{Localized\\($t=0.5$s)}}
& $\phi_H$ & \textcolor{darkgreen}{+9.6} & \textcolor{darkgreen}{+5.3} \\
& & $\phi_P$ & \textcolor{darkgreen}{+9.4} & \textcolor{darkgreen}{+13.0} \\
& & $\phi_S$ & \textcolor{red}{-1.2} & \textcolor{red}{-2.3} \\
\midrule
\multirow{8}{*}{\textbf{Llama-Mimi}}
& \multirow{4}{*}{Global}
& $\phi_0$ & \textcolor{darkgreen}{+4.4} & \textcolor{darkgreen}{+5.2} \\
& & $\phi_1$ & \textcolor{darkgreen}{+6.2} & \textcolor{darkgreen}{+9.0} \\
& & $\phi_2$ & \textcolor{darkgreen}{+12.2} & \textcolor{darkgreen}{+15.8} \\
& & $\phi_3$ & \textcolor{darkgreen}{+8.1} & \textcolor{darkgreen}{+10.5} \\
\cmidrule(lr){2-5}
& \multirow{4}{*}{\shortstack{Localized\\($t=0.5$s)}}
& $\phi_0$ & \textcolor{darkgreen}{+7.6} & \textcolor{darkgreen}{+7.8} \\
& & $\phi_1$ & \textcolor{darkgreen}{+8.9} & \textcolor{darkgreen}{+7.6} \\
& & $\phi_2$ & \textcolor{darkgreen}{+14.5} & \textcolor{darkgreen}{+16.4} \\
& & $\phi_3$ & \textcolor{darkgreen}{+11.1} & \textcolor{darkgreen}{+10.1} \\
\bottomrule
\end{tabular}

\vspace{0.5ex}

\end{table}
}

\newcommand{\TabCompositionShapleyOnly}{%
\begin{table*}[t]

\tiny
\centering
\setlength{\tabcolsep}{4pt}

\begin{tabular}{@{}c cc@{}}
\hline
\multicolumn{3}{c}{\textbf{Spirit-LM Expressive}}\\
\hline
& \textbf{Original} & \textbf{Normalized} \\
\rotatebox{90}{\textbf{Global}}
&
\begin{minipage}[t]{0.47\textwidth}
\centering
\resizebox{\linewidth}{!}{%
\begin{tabular}{ccc|cccccc|c}
\toprule
 &  &  & Sent. & Spk. & Gen. & BgD. & BgR. & Room & Avg \\
\midrule
\multicolumn{3}{c|}{$\phi_H$} & \textcolor{darkgreen}{+10.2} & \textcolor{darkgreen}{+14.8} & \textcolor{darkgreen}{+18.2} & \textcolor{darkgreen}{+1.8} & \textcolor{darkgreen}{+8.3} & \textcolor{darkgreen}{+4.3} & \textcolor{darkgreen}{+9.6} \\
\multicolumn{3}{c|}{$\phi_P$} & \textcolor{darkgreen}{+9.8} & \textcolor{darkgreen}{+17.2} & \textcolor{darkgreen}{+18.0} & \textcolor{darkgreen}{+5.8} & \textcolor{darkgreen}{+6.1} & \textcolor{red}{-0.2} & \textcolor{darkgreen}{+9.5} \\
\multicolumn{3}{c|}{$\phi_S$} & \textcolor{darkgreen}{+1.0} & \textcolor{red}{-0.5} & \textcolor{red}{-0.7} & \textcolor{red}{-2.2} & \textcolor{red}{-0.4} & \textcolor{darkgreen}{+1.3} & \textcolor{red}{-0.3} \\
\bottomrule
\end{tabular}
}
\end{minipage}
&
\begin{minipage}[t]{0.47\textwidth}
\centering
\resizebox{\linewidth}{!}{%
\begin{tabular}{ccc|cccccc|c}
\toprule
 &  &  & Sent. & Spk. & Gen. & BgD. & BgR. & Room & Avg \\
\midrule
\multicolumn{3}{c|}{$\phi_H$} & \textcolor{darkgreen}{+11.9} & \textcolor{darkgreen}{+17.8} & \textcolor{darkgreen}{+14.8} & \textcolor{darkgreen}{+3.3} & \textcolor{red}{-1.8} & \textcolor{red}{-0.2} & \textcolor{darkgreen}{+7.6} \\
\multicolumn{3}{c|}{$\phi_P$} & \textcolor{darkgreen}{+13.4} & \textcolor{darkgreen}{+12.0} & \textcolor{darkgreen}{+23.1} & \textcolor{darkgreen}{+1.6} & \textcolor{darkgreen}{+9.5} & \textcolor{darkgreen}{+18.8} & \textcolor{darkgreen}{+13.1} \\
\multicolumn{3}{c|}{$\phi_S$} & \textcolor{red}{-1.8} & \textcolor{darkgreen}{+0.2} & \textcolor{red}{-0.9} & \textcolor{red}{-1.4} & \textcolor{red}{-1.2} & \textcolor{red}{-1.0} & \textcolor{red}{-1.0} \\
\bottomrule
\end{tabular}
}
\end{minipage}
\\[1.6ex]
\raisebox{-3em}{\rotatebox{90}{\textbf{Local ($t=0.5$s)}}}
&
\begin{minipage}[t]{0.47\textwidth}
\centering
\resizebox{\linewidth}{!}{%
\begin{tabular}{ccc|cccccc|c}
\toprule
 &  &  & Sent. & Spk. & Gen. & BgD. & BgR. & Room & Avg \\
\midrule
\multicolumn{3}{c|}{$\phi_H$} & \textcolor{darkgreen}{+9.8} & \textcolor{darkgreen}{+15.2} & \textcolor{darkgreen}{+13.9} & \textcolor{darkgreen}{+4.5} & \textcolor{darkgreen}{+5.9} & \textcolor{darkgreen}{+8.3} & \textcolor{darkgreen}{+9.6} \\
\multicolumn{3}{c|}{$\phi_P$} & \textcolor{darkgreen}{+12.7} & \textcolor{darkgreen}{+18.2} & \textcolor{darkgreen}{+22.1} & \textcolor{darkgreen}{+1.8} & \textcolor{darkgreen}{+0.6} & \textcolor{darkgreen}{+1.0} & \textcolor{darkgreen}{+9.4} \\
\multicolumn{3}{c|}{$\phi_S$} & \textcolor{darkgreen}{+1.0} & \textcolor{red}{-2.8} & \textcolor{red}{-1.5} & \textcolor{red}{-1.8} & \textcolor{red}{-1.5} & \textcolor{red}{-0.8} & \textcolor{red}{-1.2} \\
\bottomrule
\end{tabular}
}
\end{minipage}
&
\begin{minipage}[t]{0.47\textwidth}
\centering
\resizebox{\linewidth}{!}{%
\begin{tabular}{ccc|cccccc|c}
\toprule
 &  &  & Sent. & Spk. & Gen. & BgD. & BgR. & Room & Avg \\
\midrule
\multicolumn{3}{c|}{$\phi_H$} & \textcolor{darkgreen}{+5.2} & \textcolor{darkgreen}{+15.2} & \textcolor{darkgreen}{+4.8} & \textcolor{darkgreen}{+1.6} & \textcolor{darkgreen}{+3.8} & \textcolor{darkgreen}{+1.2} & \textcolor{darkgreen}{+5.3} \\
\multicolumn{3}{c|}{$\phi_P$} & \textcolor{darkgreen}{+15.5} & \textcolor{darkgreen}{+15.2} & \textcolor{darkgreen}{+30.3} & \textcolor{darkgreen}{+2.1} & \textcolor{darkgreen}{+6.3} & \textcolor{darkgreen}{+8.4} & \textcolor{darkgreen}{+13.0} \\
\multicolumn{3}{c|}{$\phi_S$} & \textcolor{red}{-4.2} & \textcolor{red}{-5.0} & \textcolor{red}{-4.2} & \textcolor{darkgreen}{+1.7} & \textcolor{red}{-3.2} & \textcolor{darkgreen}{+0.9} & \textcolor{red}{-2.3} \\
\bottomrule
\end{tabular}
}
\end{minipage}
\\[2.0ex]
\hline
\multicolumn{3}{c}{\textbf{Llama-Mimi}}\\
\hline
& \textbf{Original} & \textbf{Normalized} \\
\raisebox{-1em}{\rotatebox{90}{\textbf{Global}}}
&
\begin{minipage}[t]{0.47\textwidth}
\centering
\resizebox{\linewidth}{!}{%
\begin{tabular}{cccc|cccccc|c}
\toprule
 &  &  &  & Sent. & Spk. & Gen. & BgD. & BgR. & Room & Avg \\
\midrule
\multicolumn{4}{c|}{$\phi_0$} & \textcolor{darkgreen}{+1.6} & \textcolor{darkgreen}{+5.5} & \textcolor{darkgreen}{+1.5} & \textcolor{darkgreen}{+6.6} & \textcolor{darkgreen}{+5.8} & \textcolor{darkgreen}{+5.4} & \textcolor{darkgreen}{+4.4} \\
\multicolumn{4}{c|}{$\phi_1$} & \textcolor{darkgreen}{+10.8} & \textcolor{darkgreen}{+6.5} & \textcolor{darkgreen}{+7.7} & \textcolor{darkgreen}{+4.5} & \textcolor{darkgreen}{+3.5} & \textcolor{darkgreen}{+4.4} & \textcolor{darkgreen}{+6.2} \\
\multicolumn{4}{c|}{$\phi_2$} & \textcolor{darkgreen}{+12.0} & \textcolor{darkgreen}{+12.8} & \textcolor{darkgreen}{+13.9} & \textcolor{darkgreen}{+7.2} & \textcolor{darkgreen}{+6.1} & \textcolor{darkgreen}{+21.4} & \textcolor{darkgreen}{+12.2} \\
\multicolumn{4}{c|}{$\phi_3$} & \textcolor{darkgreen}{+5.0} & \textcolor{darkgreen}{+10.8} & \textcolor{darkgreen}{+8.9} & \textcolor{darkgreen}{+6.8} & \textcolor{darkgreen}{+6.6} & \textcolor{darkgreen}{+10.8} & \textcolor{darkgreen}{+8.1} \\
\bottomrule
\end{tabular}
}
\end{minipage}
&
\begin{minipage}[t]{0.47\textwidth}
\centering
\resizebox{\linewidth}{!}{%
\begin{tabular}{cccc|cccccc|c}
\toprule
 &  &  &  & Sent. & Spk. & Gen. & BgD. & BgR. & Room & Avg \\
\midrule
\multicolumn{4}{c|}{$\phi_0$} & \textcolor{red}{-0.6} & \textcolor{darkgreen}{+6.0} & \textcolor{darkgreen}{+7.2} & \textcolor{darkgreen}{+4.2} & \textcolor{darkgreen}{+5.9} & \textcolor{darkgreen}{+8.2} & \textcolor{darkgreen}{+5.2} \\
\multicolumn{4}{c|}{$\phi_1$} & \textcolor{darkgreen}{+12.4} & \textcolor{darkgreen}{+8.6} & \textcolor{darkgreen}{+10.6} & \textcolor{darkgreen}{+1.5} & \textcolor{darkgreen}{+9.8} & \textcolor{darkgreen}{+11.1} & \textcolor{darkgreen}{+9.0} \\
\multicolumn{4}{c|}{$\phi_2$} & \textcolor{darkgreen}{+17.4} & \textcolor{darkgreen}{+19.4} & \textcolor{darkgreen}{+19.2} & \textcolor{darkgreen}{+10.4} & \textcolor{darkgreen}{+11.6} & \textcolor{darkgreen}{+16.5} & \textcolor{darkgreen}{+15.8} \\
\multicolumn{4}{c|}{$\phi_3$} & \textcolor{darkgreen}{+9.7} & \textcolor{darkgreen}{+11.0} & \textcolor{darkgreen}{+12.4} & \textcolor{darkgreen}{+7.3} & \textcolor{darkgreen}{+9.7} & \textcolor{darkgreen}{+12.8} & \textcolor{darkgreen}{+10.5} \\
\bottomrule
\end{tabular}
}
\end{minipage}
\\[1.4ex]
\smash{\raisebox{-0.5\height}{\rotatebox{90}{\textbf{Local ($t=0.5$s)}}}}
&
\begin{minipage}[t]{0.47\textwidth}
\centering
\resizebox{\linewidth}{!}{%
\begin{tabular}{cccc|cccccc|c}
\toprule
 &  &  &  & Sent. & Spk. & Gen. & BgD. & BgR. & Room & Avg \\
\midrule
\multicolumn{4}{c|}{$\phi_0$} & \textcolor{darkgreen}{+8.7} & \textcolor{darkgreen}{+10.7} & \textcolor{darkgreen}{+10.7} & \textcolor{darkgreen}{+4.0} & \textcolor{darkgreen}{+4.8} & \textcolor{darkgreen}{+6.7} & \textcolor{darkgreen}{+7.6} \\
\multicolumn{4}{c|}{$\phi_1$} & \textcolor{darkgreen}{+10.6} & \textcolor{darkgreen}{+10.1} & \textcolor{darkgreen}{+10.3} & \textcolor{darkgreen}{+3.5} & \textcolor{darkgreen}{+6.3} & \textcolor{darkgreen}{+12.4} & \textcolor{darkgreen}{+8.9} \\
\multicolumn{4}{c|}{$\phi_2$} & \textcolor{darkgreen}{+16.8} & \textcolor{darkgreen}{+13.4} & \textcolor{darkgreen}{+14.6} & \textcolor{darkgreen}{+11.2} & \textcolor{darkgreen}{+12.7} & \textcolor{darkgreen}{+18.5} & \textcolor{darkgreen}{+14.5} \\
\multicolumn{4}{c|}{$\phi_3$} & \textcolor{darkgreen}{+9.4} & \textcolor{darkgreen}{+11.3} & \textcolor{darkgreen}{+14.4} & \textcolor{darkgreen}{+10.8} & \textcolor{darkgreen}{+10.7} & \textcolor{darkgreen}{+9.9} & \textcolor{darkgreen}{+11.1} \\
\bottomrule
\end{tabular}
}
\end{minipage}
&
\begin{minipage}[t]{0.47\textwidth}
\centering
\resizebox{\linewidth}{!}{%
\begin{tabular}{cccc|cccccc|c}
\toprule
 &  &  &  & Sent. & Spk. & Gen. & BgD. & BgR. & Room & Avg \\
\midrule
\multicolumn{4}{c|}{$\phi_0$} & \textcolor{darkgreen}{+3.5} & \textcolor{darkgreen}{+9.3} & \textcolor{darkgreen}{+11.1} & \textcolor{darkgreen}{+8.5} & \textcolor{darkgreen}{+6.9} & \textcolor{darkgreen}{+7.8} & \textcolor{darkgreen}{+7.8} \\
\multicolumn{4}{c|}{$\phi_1$} & \textcolor{darkgreen}{+12.5} & \textcolor{darkgreen}{+9.2} & \textcolor{darkgreen}{+10.9} & \textcolor{darkgreen}{+0.1} & \textcolor{darkgreen}{+3.0} & \textcolor{darkgreen}{+10.1} & \textcolor{darkgreen}{+7.6} \\
\multicolumn{4}{c|}{$\phi_2$} & \textcolor{darkgreen}{+16.4} & \textcolor{darkgreen}{+17.8} & \textcolor{darkgreen}{+14.2} & \textcolor{darkgreen}{+11.5} & \textcolor{darkgreen}{+19.5} & \textcolor{darkgreen}{+18.9} & \textcolor{darkgreen}{+16.4} \\
\multicolumn{4}{c|}{$\phi_3$} & \textcolor{darkgreen}{+9.7} & \textcolor{darkgreen}{+11.6} & \textcolor{darkgreen}{+12.8} & \textcolor{darkgreen}{+11.0} & \textcolor{darkgreen}{+5.5} & \textcolor{darkgreen}{+10.2} & \textcolor{darkgreen}{+10.1} \\
\bottomrule
\end{tabular}
}
\end{minipage}

\end{tabular}
\\[1.4ex]
\end{table*}
}

\newcommand{\TabCompositionShapleyOnlyNew}
{%
\begin{table*}[t]

\tiny
\centering

\begin{tabular}{@{}c cc@{}}
\hline
\multicolumn{3}{c}{\textbf{Spirit-LM}}\\
\hline
& \textbf{Original} & \textbf{Normalized} \\
\rotatebox[origin=c]{90}{\textbf{Global}}
&
\begin{tabular}{ccc|cccccc|c}
\toprule
 &  &  & Sent. & Spk. & Gen. & BgD. & BgR. & Room & Avg \\
\midrule
\multicolumn{3}{c|}{$\phi_H$} & \textcolor{darkgreen}{+10.2} & \textcolor{darkgreen}{+14.8} & \textcolor{darkgreen}{+18.2} & \textcolor{darkgreen}{+1.8} & \textcolor{darkgreen}{+8.3} & \textcolor{darkgreen}{+4.3} & \textcolor{darkgreen}{+9.6} \\
\multicolumn{3}{c|}{$\phi_P$} & \textcolor{darkgreen}{+9.8} & \textcolor{darkgreen}{+17.2} & \textcolor{darkgreen}{+18.0} & \textcolor{darkgreen}{+5.8} & \textcolor{darkgreen}{+6.1} & \textcolor{red}{-0.2} & \textcolor{darkgreen}{+9.5} \\
\multicolumn{3}{c|}{$\phi_S$} & \textcolor{darkgreen}{+1.0} & \textcolor{red}{-0.5} & \textcolor{red}{-0.7} & \textcolor{red}{-2.2} & \textcolor{red}{-0.4} & \textcolor{darkgreen}{+1.3} & \textcolor{red}{-0.3} \\
\bottomrule
\end{tabular}
&
\begin{tabular}{ccc|cccccc|c}
\toprule
 &  &  & Sent. & Spk. & Gen. & BgD. & BgR. & Room & Avg \\
\midrule
\multicolumn{3}{c|}{$\phi_H$} & \textcolor{darkgreen}{+11.9} & \textcolor{darkgreen}{+17.8} & \textcolor{darkgreen}{+14.8} & \textcolor{darkgreen}{+3.3} & \textcolor{red}{-1.8} & \textcolor{red}{-0.2} & \textcolor{darkgreen}{+7.6} \\
\multicolumn{3}{c|}{$\phi_P$} & \textcolor{darkgreen}{+13.4} & \textcolor{darkgreen}{+12.0} & \textcolor{darkgreen}{+23.1} & \textcolor{darkgreen}{+1.6} & \textcolor{darkgreen}{+9.5} & \textcolor{darkgreen}{+18.8} & \textcolor{darkgreen}{+13.1} \\
\multicolumn{3}{c|}{$\phi_S$} & \textcolor{red}{-1.8} & \textcolor{darkgreen}{+0.2} & \textcolor{red}{-0.9} & \textcolor{red}{-1.4} & \textcolor{red}{-1.2} & \textcolor{red}{-1.0} & \textcolor{red}{-1.0} \\
\bottomrule
\end{tabular}
\\[1.6ex]
\rotatebox[origin=c]{90}{\textbf{Local ($t=0.5$s)}}
&
\begin{tabular}{ccc|cccccc|c}
\toprule
 &  &  & Sent. & Spk. & Gen. & BgD. & BgR. & Room & Avg \\
\midrule
\multicolumn{3}{c|}{$\phi_H$} & \textcolor{darkgreen}{+9.8} & \textcolor{darkgreen}{+15.2} & \textcolor{darkgreen}{+13.9} & \textcolor{darkgreen}{+4.5} & \textcolor{darkgreen}{+5.9} & \textcolor{darkgreen}{+8.3} & \textcolor{darkgreen}{+9.6} \\
\multicolumn{3}{c|}{$\phi_P$} & \textcolor{darkgreen}{+12.7} & \textcolor{darkgreen}{+18.2} & \textcolor{darkgreen}{+22.1} & \textcolor{darkgreen}{+1.8} & \textcolor{darkgreen}{+0.6} & \textcolor{darkgreen}{+1.0} & \textcolor{darkgreen}{+9.4} \\
\multicolumn{3}{c|}{$\phi_S$} & \textcolor{darkgreen}{+1.0} & \textcolor{red}{-2.8} & \textcolor{red}{-1.5} & \textcolor{red}{-1.8} & \textcolor{red}{-1.5} & \textcolor{red}{-0.8} & \textcolor{red}{-1.2} \\
\bottomrule
\end{tabular}
&
\begin{tabular}{ccc|cccccc|c}
\toprule
 &  &  & Sent. & Spk. & Gen. & BgD. & BgR. & Room & Avg \\
\midrule
\multicolumn{3}{c|}{$\phi_H$} & \textcolor{darkgreen}{+5.2} & \textcolor{darkgreen}{+15.2} & \textcolor{darkgreen}{+4.8} & \textcolor{darkgreen}{+1.6} & \textcolor{darkgreen}{+3.8} & \textcolor{darkgreen}{+1.2} & \textcolor{darkgreen}{+5.3} \\
\multicolumn{3}{c|}{$\phi_P$} & \textcolor{darkgreen}{+15.5} & \textcolor{darkgreen}{+15.2} & \textcolor{darkgreen}{+30.3} & \textcolor{darkgreen}{+2.1} & \textcolor{darkgreen}{+6.3} & \textcolor{darkgreen}{+8.4} & \textcolor{darkgreen}{+13.0} \\
\multicolumn{3}{c|}{$\phi_S$} & \textcolor{red}{-4.2} & \textcolor{red}{-5.0} & \textcolor{red}{-4.2} & \textcolor{darkgreen}{+1.7} & \textcolor{red}{-3.2} & \textcolor{darkgreen}{+0.9} & \textcolor{red}{-2.3} \\
\bottomrule
\end{tabular}
\\[2.0ex]
\hline
\multicolumn{3}{c}{\textbf{Llama-Mimi}}\\
\hline
& \textbf{Original} & \textbf{Normalized} \\
\rotatebox[origin=c]{90}{\textbf{Global}}
&
\begin{tabular}{cccc|cccccc|c}
\toprule
 &  &  &  & Sent. & Spk. & Gen. & BgD. & BgR. & Room & Avg \\
\midrule
\multicolumn{4}{c|}{$\phi_0$} & \textcolor{darkgreen}{+1.6} & \textcolor{darkgreen}{+5.5} & \textcolor{darkgreen}{+1.5} & \textcolor{darkgreen}{+6.6} & \textcolor{darkgreen}{+5.8} & \textcolor{darkgreen}{+5.4} & \textcolor{darkgreen}{+4.4} \\
\multicolumn{4}{c|}{$\phi_1$} & \textcolor{darkgreen}{+10.8} & \textcolor{darkgreen}{+6.5} & \textcolor{darkgreen}{+7.7} & \textcolor{darkgreen}{+4.5} & \textcolor{darkgreen}{+3.5} & \textcolor{darkgreen}{+4.4} & \textcolor{darkgreen}{+6.2} \\
\multicolumn{4}{c|}{$\phi_2$} & \textcolor{darkgreen}{+12.0} & \textcolor{darkgreen}{+12.8} & \textcolor{darkgreen}{+13.9} & \textcolor{darkgreen}{+7.2} & \textcolor{darkgreen}{+6.1} & \textcolor{darkgreen}{+21.4} & \textcolor{darkgreen}{+12.2} \\
\multicolumn{4}{c|}{$\phi_3$} & \textcolor{darkgreen}{+5.0} & \textcolor{darkgreen}{+10.8} & \textcolor{darkgreen}{+8.9} & \textcolor{darkgreen}{+6.8} & \textcolor{darkgreen}{+6.6} & \textcolor{darkgreen}{+10.8} & \textcolor{darkgreen}{+8.1} \\
\bottomrule
\end{tabular}
&
\begin{tabular}{cccc|cccccc|c}
\toprule
 &  &  &  & Sent. & Spk. & Gen. & BgD. & BgR. & Room & Avg \\
\midrule
\multicolumn{4}{c|}{$\phi_0$} & \textcolor{red}{-0.6} & \textcolor{darkgreen}{+6.0} & \textcolor{darkgreen}{+7.2} & \textcolor{darkgreen}{+4.2} & \textcolor{darkgreen}{+5.9} & \textcolor{darkgreen}{+8.2} & \textcolor{darkgreen}{+5.2} \\
\multicolumn{4}{c|}{$\phi_1$} & \textcolor{darkgreen}{+12.4} & \textcolor{darkgreen}{+8.6} & \textcolor{darkgreen}{+10.6} & \textcolor{darkgreen}{+1.5} & \textcolor{darkgreen}{+9.8} & \textcolor{darkgreen}{+11.1} & \textcolor{darkgreen}{+9.0} \\
\multicolumn{4}{c|}{$\phi_2$} & \textcolor{darkgreen}{+17.4} & \textcolor{darkgreen}{+19.4} & \textcolor{darkgreen}{+19.2} & \textcolor{darkgreen}{+10.4} & \textcolor{darkgreen}{+11.6} & \textcolor{darkgreen}{+16.5} & \textcolor{darkgreen}{+15.8} \\
\multicolumn{4}{c|}{$\phi_3$} & \textcolor{darkgreen}{+9.7} & \textcolor{darkgreen}{+11.0} & \textcolor{darkgreen}{+12.4} & \textcolor{darkgreen}{+7.3} & \textcolor{darkgreen}{+9.7} & \textcolor{darkgreen}{+12.8} & \textcolor{darkgreen}{+10.5} \\
\bottomrule
\end{tabular}
\\[1.4ex]
\rotatebox[origin=c]{90}{\textbf{Local ($t=0.5$s)}}
&
\begin{tabular}{cccc|cccccc|c}
\toprule
 &  &  &  & Sent. & Spk. & Gen. & BgD. & BgR. & Room & Avg \\
\midrule
\multicolumn{4}{c|}{$\phi_0$} & \textcolor{darkgreen}{+8.7} & \textcolor{darkgreen}{+10.7} & \textcolor{darkgreen}{+10.7} & \textcolor{darkgreen}{+4.0} & \textcolor{darkgreen}{+4.8} & \textcolor{darkgreen}{+6.7} & \textcolor{darkgreen}{+7.6} \\
\multicolumn{4}{c|}{$\phi_1$} & \textcolor{darkgreen}{+10.6} & \textcolor{darkgreen}{+10.1} & \textcolor{darkgreen}{+10.3} & \textcolor{darkgreen}{+3.5} & \textcolor{darkgreen}{+6.3} & \textcolor{darkgreen}{+12.4} & \textcolor{darkgreen}{+8.9} \\
\multicolumn{4}{c|}{$\phi_2$} & \textcolor{darkgreen}{+16.8} & \textcolor{darkgreen}{+13.4} & \textcolor{darkgreen}{+14.6} & \textcolor{darkgreen}{+11.2} & \textcolor{darkgreen}{+12.7} & \textcolor{darkgreen}{+18.5} & \textcolor{darkgreen}{+14.5} \\
\multicolumn{4}{c|}{$\phi_3$} & \textcolor{darkgreen}{+9.4} & \textcolor{darkgreen}{+11.3} & \textcolor{darkgreen}{+14.4} & \textcolor{darkgreen}{+10.8} & \textcolor{darkgreen}{+10.7} & \textcolor{darkgreen}{+9.9} & \textcolor{darkgreen}{+11.1} \\
\bottomrule
\end{tabular}
&
\begin{tabular}{cccc|cccccc|c}
\toprule
 &  &  &  & Sent. & Spk. & Gen. & BgD. & BgR. & Room & Avg \\
\midrule
\multicolumn{4}{c|}{$\phi_0$} & \textcolor{darkgreen}{+3.5} & \textcolor{darkgreen}{+9.3} & \textcolor{darkgreen}{+11.1} & \textcolor{darkgreen}{+8.5} & \textcolor{darkgreen}{+6.9} & \textcolor{darkgreen}{+7.8} & \textcolor{darkgreen}{+7.8} \\
\multicolumn{4}{c|}{$\phi_1$} & \textcolor{darkgreen}{+12.5} & \textcolor{darkgreen}{+9.2} & \textcolor{darkgreen}{+10.9} & \textcolor{darkgreen}{+0.1} & \textcolor{darkgreen}{+3.0} & \textcolor{darkgreen}{+10.1} & \textcolor{darkgreen}{+7.6} \\
\multicolumn{4}{c|}{$\phi_2$} & \textcolor{darkgreen}{+16.4} & \textcolor{darkgreen}{+17.8} & \textcolor{darkgreen}{+14.2} & \textcolor{darkgreen}{+11.5} & \textcolor{darkgreen}{+19.5} & \textcolor{darkgreen}{+18.9} & \textcolor{darkgreen}{+16.4} \\
\multicolumn{4}{c|}{$\phi_3$} & \textcolor{darkgreen}{+9.7} & \textcolor{darkgreen}{+11.6} & \textcolor{darkgreen}{+12.8} & \textcolor{darkgreen}{+11.0} & \textcolor{darkgreen}{+5.5} & \textcolor{darkgreen}{+10.2} & \textcolor{darkgreen}{+10.1} \\
\bottomrule
\end{tabular}

\end{tabular}

\\[1.2ex]
\caption{Shapley value summaries for Spirit-LM and Llama-Mimi. Shapley baseline assumes $null=50\%$. [20260416]}
\label{tab:composition-shapley-only-new}
\end{table*}

}

\newcommand{\TabKendallTau}{%
\begin{table*}[t!]
\small
  \caption{Kendall $\tau$ correlation between evaluator scores and human MOS scores across the 6 acoustic-consistency splits on SALMon, based on the rankings of the 7 models present in the top row of Figure~\ref{fig:correlation}.}
  \label{tab:kendall-tau}
  \centering
  \begin{tabular}{lcccccc|c}
    \toprule
    \textbf{Method} & \textbf{Sentiment} & \textbf{Speaker} & \textbf{Gender} & \textbf{Bg (domain)} & \textbf{Bg (rand.)} & \textbf{Room} & \textbf{Avg} \\
    \midrule
    Global-PPL & 0.524 & 0.333 & 0.143 & 0.619 & 0.333 & \textbf{0.714} & 0.444 \\
    + Normalization & 0.524 & 0.451 & 0.048 & 0.524 & 0.619 & \textbf{0.714} & 0.480 \\
    Windowed-PPL & 0.524 & 0.429 & 0.238 & 0.619 & 0.586 & 0.619 & 0.502 \\
    Localized-PPL & 0.488 & 0.048 & 0.195 & 0.714 & 0.619 & 0.333 & 0.400 \\
    + Normalization & 0.429 & -0.048 & -0.098 & 0.238 & 0.619 & 0.524 & 0.277 \\
    Model-as-a-judge & \textbf{0.683} & \textbf{0.878} & \textbf{0.524} & \textbf{0.878} & \textbf{0.651} & 0.429 & \textbf{0.674} \\
    \bottomrule
  \end{tabular}%
\end{table*}
}
\newcommand{\TabJudgeTestGen}{%

\begin{table*}[t!]
  \caption{Generation performance of spoken language models judged by the appropriate model as a judge. Main numbers report performance of continuation samples from the SLM; parenthesized numbers report the performance of reconstructed audio that is generated from the speech tokens of SLMs. Boldface items indicate that the generation performance exceeds the human topline reported by \cite{maimon2025salmon}.
      \label{tab:judge-test-ent50}}
  \centering
  \resizebox{\textwidth}{!}{
  \renewcommand{\arraystretch}{2}
  \begin{tabular}{|>{\raggedright\arraybackslash}p{4cm}|cccccc|}
    \hline
      & \multicolumn{6}{c|}{\textbf{Acoustic Consistency}} \\
      & Sentiment $\uparrow$ & Speaker $\uparrow$ & Gender $\uparrow$ & Bg (domain) $\uparrow$ & Bg (rand.) $\uparrow$ & Room $\uparrow$ \\
    \midrule
    \midrule
    \diagbox[
        width=4cm,
        height=3.6\baselineskip,
        outerleftsep=0pt,
        outerrightsep=0pt,
        innerleftsep=0pt,
        innerrightsep=0pt,
      ]{Evaluated Model}{Judge Model} & TITANET & TITANET & TITANET & HuBERT-large-audioset & HuBERT-large-audioset & wav2vec2-large-audioset\\
    \midrule
    \midrule
    GSLM
      & \shortstack[c]{55.0\\{\small(57.5)}} & \shortstack[c]{55.0\\{\small(58.5)}} & \shortstack[c]{50.5\\{\small(50.0)}} & \shortstack[c]{59.5\\{\small(60.0)}} & \shortstack[c]{55.5\\{\small(52.0)}} & \shortstack[c]{53.0\\{\small(48.0)}} \\
    \midrule
    TWIST-350M
      & \shortstack[c]{48.0\\{\small(48.5)}} & \shortstack[c]{53.0\\{\small(53.5)}} & \shortstack[c]{52.5\\{\small(53.0)}} & \shortstack[c]{61.5\\{\small(60.0)}} & \shortstack[c]{53.0\\{\small(53.5)}} & \shortstack[c]{52.5\\{\small(49.0)}} \\
    \midrule
    TWIST-1.3B
      & \shortstack[c]{48.0\\{\small(50.5)}} & \shortstack[c]{55.0\\{\small(52.5)}} & \shortstack[c]{54.5\\{\small(52.0)}} & \shortstack[c]{62.5\\{\small(60.0)}} & \shortstack[c]{54.0\\{\small(54.0)}} & \shortstack[c]{53.5\\{\small(48.5)}} \\
    \midrule
    TWIST-7B
      & \shortstack[c]{48.0\\{\small(51.0)}} & \shortstack[c]{54.5\\{\small(53.0)}} & \shortstack[c]{52.0\\{\small(53.0)}} & \shortstack[c]{59.5\\{\small(60.5)}} & \shortstack[c]{54.0\\{\small(54.0)}} & \shortstack[c]{51.5\\{\small(49.0)}} \\
    \midrule
    pGSLM
      & \shortstack[c]{46.5\\{\small(52.0)}} & \shortstack[c]{49.5\\{\small(57.5)}} & \shortstack[c]{55.5\\{\small(58.0)}} & \shortstack[c]{51.0\\{\small(51.0)}} & \shortstack[c]{54.0\\{\small(53.5)}} & \shortstack[c]{52.5\\{\small(49.0)}} \\
    \midrule
    Spirit-LM
      & \shortstack[c]{47.5\\{\small(50.0)}} & \shortstack[c]{52.0\\{\small(50.5)}} & \shortstack[c]{50.5\\{\small(50.5)}} & \shortstack[c]{50.5\\{\small(51.5)}} & \shortstack[c]{52.5\\{\small(53.0)}} & \shortstack[c]{50.0\\{\small(47.5)}} \\
    \midrule
    Spirit-LM-expr.
      & \shortstack[c]{72.0\\{\small(81.0)}} & \shortstack[c]{54.0\\{\small(53.0)}} & \shortstack[c]{56.0\\{\small(58.0)}} & \shortstack[c]{58.5\\{\small(62.0)}} & \shortstack[c]{55.5\\{\small(56.0)}} & \shortstack[c]{54.5\\{\small(50.0)}} \\
    \midrule
    TASTE-emb.
      & \shortstack[c]{95.0\\{\small(95.0)}} & \textbf{\shortstack[c]{100.0\\{\small(99.0)}}} & \textbf{\shortstack[c]{100.0\\{\small(99.5)}}} & \shortstack[c]{61.5\\{\small(63.0)}} & \shortstack[c]{54.5\\{\small(53.5)}} & \shortstack[c]{57.5\\{\small(53.0)}} \\
    \midrule
    Flow-SLM-270M
      & \shortstack[c]{85.0\\{\small(98.0)}} & \shortstack[c]{87.0\\{\small(100.0)}} & \shortstack[c]{92.5\\{\small(100.0)}} & \shortstack[c]{77.0\\{\small(80.0)}} & \shortstack[c]{86.0\\{\small(94.0)}} & \shortstack[c]{69.5\\{\small(96.5)}} \\
    \midrule
    Flow-SLM-1B
      & \shortstack[c]{92.5\\{\small(98.0)}} & \textbf{\shortstack[c]{99.5\\{\small(100.0)}}} & \textbf{\shortstack[c]{99.0\\{\small(100.0)}}} & \shortstack[c]{77.0\\{\small(80.0)}} & \shortstack[c]{84.5\\{\small(94.0)}} & \shortstack[c]{77.0\\{\small(97.0)}} \\
    \midrule
    Flow-SLM-1B-Ext.
      & \shortstack[c]{93.0\\{\small(98.0)}} & \textbf{\shortstack[c]{98.0\\{\small(100.0)}}} & \textbf{\shortstack[c]{100.0\\{\small(100.0)}}} & \shortstack[c]{77.0\\{\small(80.0)}} & \shortstack[c]{86.5\\{\small(94.0)}} & \shortstack[c]{85.0\\{\small(97.0)}} \\
    \midrule
    Llama-mimi
      & \shortstack[c]{93.0\\{\small(92.0)}} & \shortstack[c]{90.0\\{\small(95.0)}} & \shortstack[c]{95.0\\{\small(95.0)}} & \shortstack[c]{81.0\\{\small(78.5)}} & \textbf{\shortstack[c]{90.0\\{\small(91.0)}}} & \shortstack[c]{78.5\\{\small(85.0)}} \\

    \midrule
    \midrule
    Human Topline
      & 97.2 
      & 91.5 
      & 98.6 
      & 83.1 
      & 88.7 
      & 94.4 \\
    \bottomrule
  \end{tabular}}
  
\end{table*}
}

\newcommand{\TabPPLMain}{%
\begin{table*}[t!]
  \caption{Comparison of spoken language model performance on the SALMon benchmark, including GSLM \cite{kharitonov-etal-2022-text}, TWIST~\cite{twist}, pGSLM \cite{pgslm}, Spirit-LM~\cite{spiritlm}, TASTE~\cite{tseng2025taste}, Flow-SLM \cite{chou2025flow}, and Llama-mimi \cite{sugiura2025llama}. For each model, we report task accuracies measured with global token perplexity and our proposed likelihood evaluators. Comparing accuracies within each model shows that the perceived performance difference between methods is quite huge. As a result, these evaluation protocols yield substantially different conclusions on the SLM performance landscape. Bold number highlight evaluation results that surpass the human topline performance by \cite{maimon2025salmon}. \label{tab:ppl-main}}
  \centering
  \resizebox{\textwidth}{!}{%
  \begin{tabular}{|l|l|l|cccccc|cc|}
    \toprule
    \multicolumn{3}{|c|}{} &
      \multicolumn{6}{c|}{\textbf{Acoustic Consistency}} &
      \multicolumn{2}{c|}{\textbf{Semantic-Acoustic Alignment}} \\
    \textbf{Model} & \textbf{Config} & \textbf{Method} &
      Sentiment $\uparrow$ & Speaker $\uparrow$ & Gender $\uparrow$ &
      Bg (domain) $\uparrow$ & Bg (rand.) $\uparrow$ & Room $\uparrow$ &
      Sentiment $\uparrow$ & Background $\uparrow$ \\
    \midrule

    \multirow{5}{*}{\rotatebox[origin=c]{90}{GSLM}} & \multirow{5}{*}{--} & Global-PPL & 52.5 & 50.5 & 53.0 & 47.5 & 50.5 & 48.0 & 55.0 & 52.5 \\
     &  & + Normalization (prop.) & 48.5 & 52.0 & 52.0 & 45.0 & 52.0 & 50.0 & - & - \\
     &  & Windowed-PPL (prop.) & 53.2 & 50.5 & 49.8 & 52.8 & 52.0 & 47.8 & 56.0 & 43.0 \\
     &  & Localized-PPL (prop.) & 52.0 & 49.0 & 48.5 & 47.5 & 46.5 & 42.5 & - & - \\
     &  & + Normalization (prop.) & 55.5 & 47.0 & 49.0 & 46.5 & 45.5 & 51.0 & - & - \\

    \midrule

    \multirow{15}{*}{\rotatebox[origin=c]{90}{TWIST}} & \multirow{5}{*}{350M} & Global-PPL & 59.0 & 69.5 & 68.0 & 54.0 & 61.5 & 59.0 & 51.0 & 56.5 \\
     &  & + Normalization (prop.) & 58.0 & 74.5 & 71.5 & 56.0 & 62.0 & 70.5 & - & - \\
     &  & Windowed-PPL (prop.) & 59.8 & 57.0 & 55.8 & 52.2 & 64.0 & 65.5 & 51.5 & 58.0 \\
     &  & Localized-PPL (prop.) & 53.0 & 60.5 & 63.5 & 53.2 & 53.5 & 57.2 & - & - \\
     &  & + Normalization (prop.) & 55.5 & 63.0 & 60.5 & 56.5 & 61.5 & 66.0 & - & - \\
    \cmidrule(lr){2-11}
     & \multirow{5}{*}{1.3B} & Global-PPL & 61.0 & 69.0 & 69.5 & 54.5 & 60.5 & 59.5 & 52.5 & 57.0 \\
     &  & + Normalization (prop.) & 54.0 & 72.5 & 73.0 & 57.5 & 63.5 & 69.5 & - & - \\
     &  & Windowed-PPL (prop.) & 58.2 & 55.2 & 55.0 & 55.2 & 61.8 & 66.0 & 54.0 & 55.0 \\
     &  & Localized-PPL (prop.) & 50.0 & 61.5 & 68.0 & 54.8 & 53.0 & 54.2 & - & - \\
     &  & + Normalization (prop.) & 52.5 & 65.5 & 62.0 & 60.5 & 62.5 & 62.0 & - & - \\
    \cmidrule(lr){2-11}
     & \multirow{5}{*}{7B} & Global-PPL & 61.5 & 70.5 & 70.0 & 55.5 & 59.5 & 61.5 & 51.0 & 53.5 \\
     &  & + Normalization (prop.) & 56.0 & 74.0 & 72.0 & 59.0 & 64.5 & 69.0 & - & - \\
     &  & Windowed-PPL (prop.) & 60.0 & 57.8 & 53.8 & 55.8 & 59.2 & 69.8 & 51.5 & 56.5 \\
     &  & Localized-PPL (prop.) & 53.0 & 55.5 & 65.0 & 58.8 & 56.0 & 57.8 & - & - \\
     &  & + Normalization (prop.) & 53.0 & 67.0 & 63.5 & 58.0 & 65.0 & 65.5 & - & - \\

    \midrule

    \multirow{5}{*}{\rotatebox[origin=c]{90}{pGSLM}} & \multirow{5}{*}{--} & Global-PPL & 56.0 & 74.0 & 69.0 & 65.5 & 71.5 & 52.0 & 53.0 & 54.0 \\
     &  & + Normalization (prop.) & 60.5 & 80.0 & 89.0 & 62.5 & 71.0 & 69.0 & - & - \\
     &  & Windowed-PPL (prop.) & 59.0 & 61.8 & 63.2 & 64.5 & 61.5 & 63.5 & 49.0 & 52.5 \\
     &  & Localized-PPL (prop.) & 70.5 & 78.5 & 68.0 & 50.5 & 57.5 & 56.0 & - & - \\
     &  & + Normalization (prop.) & 66.0 & 82.5 & 87.5 & 57.0 & 66.5 & 68.5 & - & - \\

    \midrule

    \multirow{10}{*}{\rotatebox[origin=c]{90}{Spirit LM}} & \multirow{5}{*}{base} & Global-PPL & 52.5 & 67.5 & 67.0 & 53.0 & 55.5 & 54.5 & 47.5 & 51.0 \\
     &  & + Normalization (prop.) & 51.5 & 60.0 & 67.0 & 49.0 & 52.0 & 64.5 & - & - \\
     &  & Windowed-PPL (prop.) & 48.8 & 50.8 & 58.5 & 53.8 & 56.0 & 68.5 & 46.0 & 60.8 \\
     &  & Localized-PPL (prop.) & 54.5 & 62.0 & 64.0 & 56.0 & 48.5 & 52.8 & - & - \\
     &  & + Normalization (prop.) & 54.0 & 57.0 & 59.5 & 55.5 & 57.0 & 52.0 & - & - \\
    \cmidrule(lr){2-11}
     & \multirow{5}{*}{Expr.} & Global-PPL & 71.0 & 81.5 & 85.5 & 55.5 & 64.0 & 55.5 & 53.0 & 59.5 \\
     &  & + Normalization (prop.) & 73.5 & 80.0 & 87.0 & 53.5 & 56.5 & 67.5 & - & - \\
     &  & Windowed-PPL (prop.) & 58.5 & 66.0 & 67.5 & 50.0 & 57.5 & 64.8 & 53.0 & 55.0 \\
     &  & Localized-PPL (prop.) & 73.5 & 80.5 & 84.5 & 54.5 & 55.0 & 58.5 & - & - \\
     &  & + Normalization (prop.) & 66.5 & 75.5 & 81.0 & 55.5 & 57.0 & 60.5 & - & - \\

    \midrule

    \multirow{5}{*}{\rotatebox[origin=c]{90}{TASTE}} & \multirow{5}{*}{emb} & Global-PPL & 56.5 & 69.5 & 75.5 & 39.5 & 48.5 & 55.5 & 54.0 & 42.0 \\
     &  & + Normalization (prop.) & 62.0 & 67.0 & 68.0 & 49.0 & 49.0 & 55.0 & - & - \\
     &  & Windowed-PPL (prop.) & 57.2 & 52.2 & 57.5 & 48.2 & 49.5 & 51.0 & 51.5 & 48.0 \\
     &  & Localized-PPL (prop.) & 54.5 & 57.0 & 60.5 & 44.5 & 43.8 & 54.0 & - & - \\
     &  & + Normalization (prop.) & 55.5 & 52.0 & 55.5 & 47.0 & 48.5 & 53.5 & - & - \\

    \midrule

    \multirow{15}{*}{\rotatebox[origin=c]{90}{Flow-SLM}} & \multirow{5}{*}{270M} & Global-PPL & 61.5 & 75.0 & 76.5 & 66.0 & 65.0 & 56.5 & 59.0 & 53.5 \\
     &  & + Normalization (prop.) & 77.5 & 86.0 & 82.5 & 56.0 & 63.5 & 80.5 & - & - \\
     &  & Windowed-PPL (prop.) & 65.5 & 68.2 & 71.8 & 55.5 & 60.8 & 66.5 & 54.5 & 55.0 \\
     &  & Localized-PPL (prop.) & 75.0 & 72.0 & 77.5 & 62.5 & 62.0 & 67.5 & - & - \\
     &  & + Normalization (prop.) & 75.5 & 68.5 & 77.0 & 54.5 & 63.0 & 78.0 & - & - \\
    \cmidrule(lr){2-11}
     & \multirow{5}{*}{1B} & Global-PPL & 65.5 & 77.0 & 79.0 & 69.0 & 66.0 & 76.0 & 58.5 & 54.5 \\
     &  & + Normalization (prop.) & 78.0 & 85.0 & 87.5 & 58.0 & 69.0 & 88.5 & - & - \\
     &  & Windowed-PPL (prop.) & 67.8 & 66.8 & 75.2 & 59.0 & 61.2 & 77.2 & 58.0 & 53.5 \\
     &  & Localized-PPL (prop.) & 79.5 & 73.5 & 79.0 & 64.5 & 68.0 & 81.5 & - & - \\
     &  & + Normalization (prop.) & 80.5 & 69.0 & 80.0 & 53.0 & 63.0 & 78.5 & - & - \\
    \cmidrule(lr){2-11}
     & \multirow{5}{*}{1B-ext.} & Global-PPL & 64.5 & 76.0 & 80.5 & 69.5 & 64.5 & 73.5 & 57.5 & 53.5 \\
     &  & + Normalization (prop.) & 82.5 & 85.0 & 88.5 & 58.5 & 73.5 & 85.5 & - & - \\
     &  & Windowed-PPL (prop.) & 66.5 & 71.2 & 73.0 & 61.5 & 61.8 & 77.0 & 52.0 & 57.0 \\
     &  & Localized-PPL (prop.) & 79.5 & 76.0 & 81.5 & 65.5 & 69.5 & 83.5 & - & - \\
     &  & + Normalization (prop.) & 82.0 & 74.5 & 81.0 & 54.0 & 67.5 & 81.0 & - & - \\

    \midrule

    \multirow{5}{*}{\rotatebox[origin=c]{90}{Llama-mimi}} & \multirow{5}{*}{1.3B} & Global-PPL & 79.5 & 85.5 & 82.0 & 75.0 & 72.0 & 92.0 & 53.0 & 49.0 \\
     &  & + Normalization (prop.) & 89.0 & \textbf{95.0} & \textbf{99.5} & 73.5 & 87.0 & \textbf{98.5} & - & - \\
     &  & Windowed-PPL (prop.) & 80.5 & 91.2 & 96.2 & 65.2 & 66.5 & 81.2 & 51.0 & 53.5 \\
     &  & Localized-PPL (prop.) & 95.5 & \textbf{95.5} & \textbf{100.0} & 79.5 & 84.5 & \textbf{97.5} & - & - \\
     &  & + Normalization (prop.) & 92.0 & \textbf{98.0} & \textbf{99.0} & 81.0 & 85.0 & \textbf{97.0} & - & - \\

    \midrule

    \multicolumn{11}{|c|}{\textbf{Human Topline (Measured by \cite{maimon2025salmon})}}\\
    \hline
    Human & -- & -- & 97.2 & 91.5 & 98.6 & 83.1 & 88.7 & 94.4 & 93.3 & 95.7 \\
    \bottomrule
  \end{tabular}
  }
  
\end{table*}
}

\newcommand{\TabExpHumanEval}{%
\begin{table*}[t!]
  \caption{Generation performance of SLMs judged by human ratings (MOS scores) with the model's associated rank.}
  \label{tab:exp-human-eval}
  \centering
  \resizebox{\textwidth}{!}{%
  \begin{tabular}{|l|cccccc|cc|}
    \toprule
      \multicolumn{9}{|c|}{\textbf{MOS Scores Evaluation}}\\
      \hline
      & Sentiment $\uparrow$ & Speaker $\uparrow$ & Gender $\uparrow$ &
        Bg (domain) $\uparrow$ & Bg (rand.) $\uparrow$ & Room $\uparrow$ &
        Avg $\uparrow$ & Rank \\
    \midrule

    GSLM
      & $1.88\pm1.06$ & $1.94\pm1.13$ & $2.76\pm1.61$ & $1.38\pm0.85$ & $1.36\pm0.86$ & $1.82\pm1.00$ & $1.86\pm0.45$ & 7 \\
    %TWIST 350M
      %& $1.95\pm1.15$ & $2.02\pm1.16$ & $2.74\pm1.59$ & $1.59\pm1.02$ & $1.50\pm0.89$ & $2.33\pm1.25$ & $2.02\pm0.49$ & 7 \\
    TWIST-1.3B
      & $1.91\pm1.08$ & $2.04\pm1.18$ & $2.73\pm1.55$ & $1.62\pm1.07$ & $1.59\pm0.99$ & $2.29\pm1.21$ & $2.03\pm0.49$ & 5 \\
    %TWIST 7B
      %& -- & -- & -- & -- & -- & -- & -- & -- \\
    pGSLM
      & $1.86\pm1.05$ & $1.76\pm1.05$ & $2.38\pm1.28$ & $1.34\pm0.78$ & $1.28\pm0.71$ & $1.65\pm0.97$ & $1.71\pm0.40$ & 8 \\
    %Spirit-LM
      %& -- & -- & -- & -- & -- & -- & -- & -- \\
    Spirit-LM-Expr.
      & $3.41\pm1.49$ & $1.98\pm1.14$ & $2.63\pm1.49$ & $1.27\pm0.73$ & $1.19\pm0.51$ & $1.58\pm0.98$ & $2.01\pm0.46$ & 6 \\
    TASTE-emb.
      & $3.68\pm1.40$ & $4.37\pm1.02$ & $4.63\pm0.93$ & $1.64\pm1.16$ & $1.60\pm1.04$ & $2.29\pm1.27$ & $3.03\pm0.47$ & 4 \\
    %Flow-SLM-270m
      %& -- & -- & -- & -- & -- & -- & -- & -- \\
    Flow-SLM-1B
      & $3.86\pm1.27$ & $4.21\pm1.13$ & $4.47\pm0.96$ & $1.89\pm1.08$ & $1.86\pm1.12$ & $3.25\pm1.34$ & $3.26\pm0.47$ & 3 \\
    Flow-SLM-1B-Ext.
      & $3.80\pm1.31$ & $4.20\pm1.10$ & $4.52\pm0.94$ & $1.98\pm1.13$ & $2.00\pm1.23$ & $3.08\pm1.41$ & $3.26\pm0.49$ & 2 \\
    Llama-Mimi-1.3B
      & $3.78\pm1.31$ & $4.14\pm1.16$ & $4.32\pm1.10$ & $2.20\pm1.30$ & $2.21\pm1.29$ & $3.11\pm1.41$ & $3.29\pm0.52$ & 1 \\

    \bottomrule
  \end{tabular}
  }
  
\end{table*}

% \begin{table*}[t!]
%   \caption{Generation Performance of SpokenLMs judged by Human.}
%   \label{tab:exp-human-eval}
%   \centering
%   \resizebox{\textwidth}{!}{%
%   \begin{tabular}{|l|cccccc|cc|}
%     \toprule
%       \multicolumn{9}{|c|}{\textbf{MOS Scores Evaluation}}\\
%       \hline
%       & Sentiment $\uparrow$ & Speaker $\uparrow$ & Gender $\uparrow$ &
%         Bg (domain) $\uparrow$ & Bg (rand.) $\uparrow$ & Room $\uparrow$ &
%         Avg $\uparrow$ & Rank \\
%     \midrule

%     GSLM
%       & 1.88 & 1.94 & 2.76 & 1.38 & 1.36 & 1.82 & 1.857 & 9 \\
%     %TWIST 350M
%       %& 1.95 & 2.02 & 2.74 & 1.59 & 1.50 & 2.33 & 2.022 & 7 \\
%     TWIST 1.3B
%       & 1.91 & 2.04 & 2.73 & 1.62 & 1.59 & 2.29 & 2.029 & 6 \\
%     %TWIST 7B
%       %& -- & -- & -- & -- & -- & -- & -- & -- \\
%     pGSLM
%       & 1.86 & 1.76 & 2.38 & 1.34 & 1.28 & 1.65 & 1.710 & 10 \\
%     %Spirit-LM
%       %& -- & -- & -- & -- & -- & -- & -- & -- \\
%     Spirit-LM-expr.
%       & 3.41 & 1.98 & 2.63 & 1.27 & 1.19 & 1.58 & 2.008 & 8 \\
%     TASTE-emb.
%       & 3.68  & 4.37 & 4.63	& 1.60 & 1.64 & 2.29 &  \\
%     %Flow-SLM-270m
%       %& -- & -- & -- & -- & -- & -- & -- & -- \\
%     Flow-SLM-1b
%       & 3.86 & 4.21 & 4.47 & 1.89 & 1.86 & 3.25 & 3.257 & 3 \\
%     Flow-SLM-1b-ext.
%       & 3.80 & 4.20 & 4.52 & 1.98 & 2.00 & 3.08 & 3.264 & 2 \\
%     Llama-mimi
%       & 3.78 & 4.14 & 4.32 & 2.20 & 2.21 & 3.11 & 3.293 & 1 \\

%     \bottomrule
%   \end{tabular}
%   }
% \end{table*}

}
\newcommand{\TabJudgeDev}{%
\begin{table*}[t!]
  \caption{Embedding model performance on SALMon, where the accuracy is aggregated over $d(S, P) < d (S,N)$. \label{tab:judge-dev}}
  \centering
  \resizebox{\textwidth}{!}{\begin{tabular}{|l|cccccc|}
    \toprule
      \multicolumn{7}{|c|}{\textbf{Embedding Model Performance}}\\
      \hline
      & \multicolumn{6}{c|}{\textbf{Acoustic Consistency}} \\
      & Sentiment $\uparrow$ & Speaker $\uparrow$ & Gender $\uparrow$ & Bg (domain) $\uparrow$ & Bg (rand.) $\uparrow$ & Room $\uparrow$ \\
    \midrule
    TITANET \cite{koluguri2022titanet}
      & \textbf{99.5} & \textbf{100.0} & \textbf{100.0} & 58.5 & 70.5 & 94.0 \\
    CAM++ \cite{wang2023cam++}
      & 95.5 & \textbf{95.5} & \textbf{99.0} & 69.5 & 84.5 & \textbf{94.5} \\
    CLAP \cite{elizalde2023clap}
      & 96.0 & 91.0 & 98.5 & 78.0 & \textbf{90.0} & \textbf{97.0} \\
    wav2vec2-large-audioset \cite{ARCH}
      & 91.5 & 70.0 & 75.5 & 82.5 & \textbf{95.5} & \textbf{99.0} \\
    HuBERT-large-audioset \cite{ARCH}
      & 90.0 & 74.5 & 74.5 & \textbf{86.5} & \textbf{97.5} & \textbf{95.5} \\
    data2vec-audio-large \cite{baevski2022data2vecgeneralframeworkselfsupervised}
      & 58.5 & 67.0 & 59.5 & 54.0 & 45.5 & 77.5 \\
    wavlm-large \cite{Chen_2022}
      & 75.0 & 66.5 & 69.5 & 52.5 & 56.0 & 93.0 \\
    %hubert-large-ll60k \cite{ARCH}
    %  & 83.5 & 68.0 & 70.0 & 63.5 & 77.5 & \textbf{100.0} \\
    wav2vec2-large-robust \cite{hsu2021robustwav2vec20analyzing}
      & 74.5 & 58.5 & 52.5 & 59.5 & 67.5 & 76.5 \\
    \midrule
    Selected Classifier
      & TITANET & TITANET & TITANET & HuBERT-large-audioset & hubert-large-audioset & wav2vec2-large-audioset \\
    \midrule
    Human
      & 97.2 & 91.5 & 98.6 & 83.1 & 88.7 & 94.4 \\
    \bottomrule
  \end{tabular}}
  
\end{table*}
}

\newcommand{\TabJudgeDevSmall}{%
\begin{table*}[t!]
  \centering
  \caption{The best-performing embedding model $E$ on each task provides a viable judge $J$ for evaluating continuation performance. In addition to surpassing the human baseline, four of the six models achieve near perfect performance.}
  \label{tab:judge-dev-small}
  \resizebox{\textwidth}{!}{%
    \begin{tabular}{|l|cccccc|}
      \toprule
      %\multicolumn{7}{|c|}{\textbf{Embedding Model Dev-set Performance}}\\
      %\hline
      & Sentiment $\uparrow$ & Speaker $\uparrow$ & Gender $\uparrow$ & Bg (domain) $\uparrow$ & Bg (rand.) $\uparrow$ & Room $\uparrow$ \\
      \midrule
      Selected embedding model    & TITANET & TITANET & TITANET & HuBERT-large-audioset & HuBERT-large-audioset & wav2vec2-large-audioset \\
      Classifier performance & \textbf{\underline{99.5}} & \textbf{\underline{100.0}} & \textbf{\underline{100.0}} & \textbf{\underline{86.5}} & \textbf{\underline{97.5}} & \textbf{\underline{100.0}} \\
Human performance      & 97.2 & 91.5 & 98.6 & 83.1 & 88.7 & 94.4 \\

      \bottomrule
    \end{tabular}%
  }

\end{table*}}

\maketitle
\begin{abstract}
%Generative spoken language models pretrained on large-scale raw audio are designed to preserve attribute continuity in speech during generations. 
Generative spoken language models pretrained on large-scale raw audio can continue a speech prompt with appropriate content while preserving attributes like speaker and emotion, serving as foundation models for spoken dialogue.
%
% In prior literature, these models are often evaluated by ``global token perplexity'', which directly adopts the text perplexity formula on speech tokens. However, there exists fundamental differences between speech and text modalities that ...
%
In prior literature, these models are often evaluated using ``global token perplexity'', which directly applies the text perplexity formulation to speech tokens. However, this practice overlooks fundamental differences between speech and text modalities, possibly leading to an underestimation of the speech characteristics.
%
% These models are often evaluated by overloading text perplexity into a speech setting to obtain “global token perplexity” of speech, but there exist fundamental differences between text and speech signals that complicate this derivation. 
%In this work, we propose alternative estimations to global token perplexity that better reflect perceived generation quality, as evidenced by stronger correlation with human-rated mean opinion score (MOS). 
In this work, we propose a variety of likelihood- and generative-based evaluation methods that serve in place of naive global token perplexity. We demonstrate that the proposed evaluations more faithfully reflect perceived generation quality, as evidenced by stronger correlations with human-rated mean opinion scores (MOS).
When assessed under the new metrics, the relative performance landscape of spoken language models is reshaped, revealing a significantly reduced gap between the best-performing model and the human topline. Together, these results suggest that appropriate evaluation is critical for accurately assessing progress in spoken language modeling.\footnote{Code and data open-sourced at \url{https://github.com/Lab-MSP/SpeechPerplexity}}
% \textcolor{red}{Adopting the new metric, ... .Our results with new evaluations reshape the performance landscape of spoken language models, where the gap between the best model and human topline is significantly closed. Additional analysis shed light on the internal dynamics that drive shifts in perceived model performance.}
\end{abstract}

\section{Introduction}
\begin{comment}
Recent years have witnessed the emergence of assistant-style spoken dialogue systems that interact with users through speech \cite{GPT-4o, chu2024qwen2audio, defossez2024moshi}.
Generative speech modeling is made available by framing it as a sequence to sequence task \cite{sugiura2025llama, chang2024speechprompt} -- speech waveforms are discretized into token sequences \cite{defossez2024moshi, an2024funaudiollm}, models predictions operate in the token space \cite{sugiura2025llama, tseng2025taste}, and the original audio is reconstructed from predicted tokens \cite{du2024cosyvoice, ju2024naturalspeech}. We refer this structure as spoken language model (SLM) throughout following \citet{arora2025landscape}.
Analogous to the development of text-only large language models (LLMs), SLMs are typically developed in two stages: auto-regressive large-scale pretraining on unlabeled speech data, followed by supervised finetuning on task- or instruction-specific data.
\end{comment}

Recent years have witnessed the emergence of assistant-style spoken dialogue systems that interact with users through speech \cite{GPT-4o, chu2024qwen2audio, defossez2024moshi}.
Analogous to text language models, generative speech modeling is commonly formulated as a sequence-to-sequence task \cite{sugiura2025llama, chang2024speechprompt}. In this paradigm, speech waveforms are first discretized into sequences of tokens \cite{defossez2024moshi, an2024funaudiollm}; model predictions are then performed in the token space \cite{sugiura2025llama, tseng2025taste}; finally, the resulting audio is reconstructed from the predicted tokens \cite{du2024cosyvoice, ju2024naturalspeech}.
We refer to this modeling framework as a \emph{spoken language model (SLM)}, following the terminology of \citet{arora2025landscape}.
Similar to the development of text-based large language models (LLMs), SLMs are typically trained in two stages: large-scale pretraining on unlabeled speech data, followed by supervised fine-tuning on task- or instruction-specific datasets.

\begin{comment}
Recent years have witnessed the emergence of assistant-style dialogue systems that engage users in both text \cite{openai2025gpt5, google2025gemini} and speech modalities \cite{GPT-4o, chu2024qwen2audio, defossez2024moshi} and are applied broadly across for downstream tasks \cite{wu2024bloomberggpt, cui2025effects, chen2021evaluating}. A central ingredient is large-scale unsupervised pretraining on raw, human-authored corpora \cite{floridi2020gpt, dubey2024llama}, where auto-regressive next-token prediction enables models to internalize the language regularities and world knowledge implicitly encoded in such data. This paradigm was first established in text models, and subsequently extended to spoken language models \cite{sugiura2025llama} -- speech waveforms are discretized into token sequences \cite{defossez2024moshi, an2024funaudiollm}, models predictions operate in the token space \cite{sugiura2025llama, tseng2025taste}, and the original audio is reconstructed from predicted tokens \cite{du2024cosyvoice, ju2024naturalspeech}.  
%Standardized evaluations are imperative, for these intermediate probabilistic models provide the foundation for universal utility.
\end{comment}
\begin{figure}[t!]
  \centering
  % tighten spacing between cells
  \setlength{\tabcolsep}{4pt}
  \renewcommand{\arraystretch}{0}
    \hspace*{0.015\textwidth}%
    \includegraphics[width=0.48\textwidth]{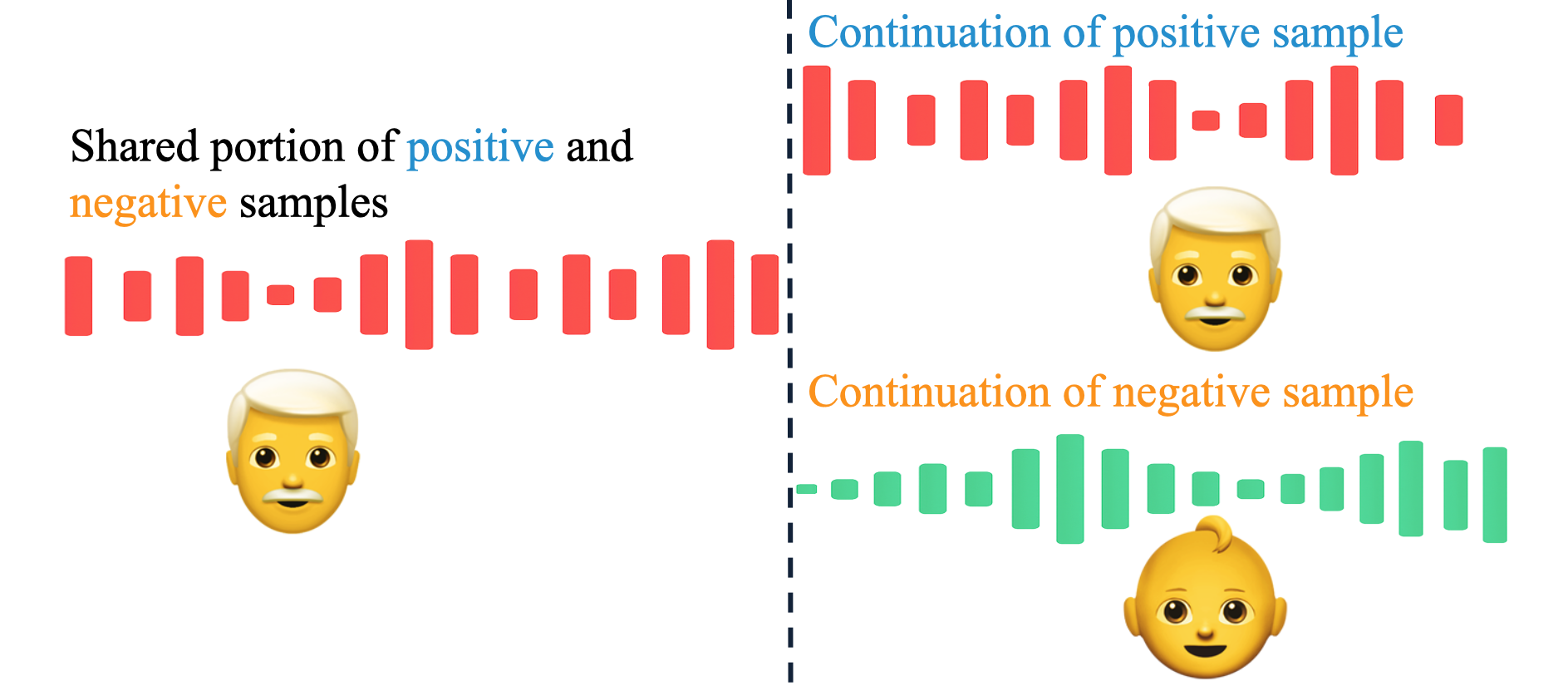}
    \includegraphics[width=0.48\textwidth]{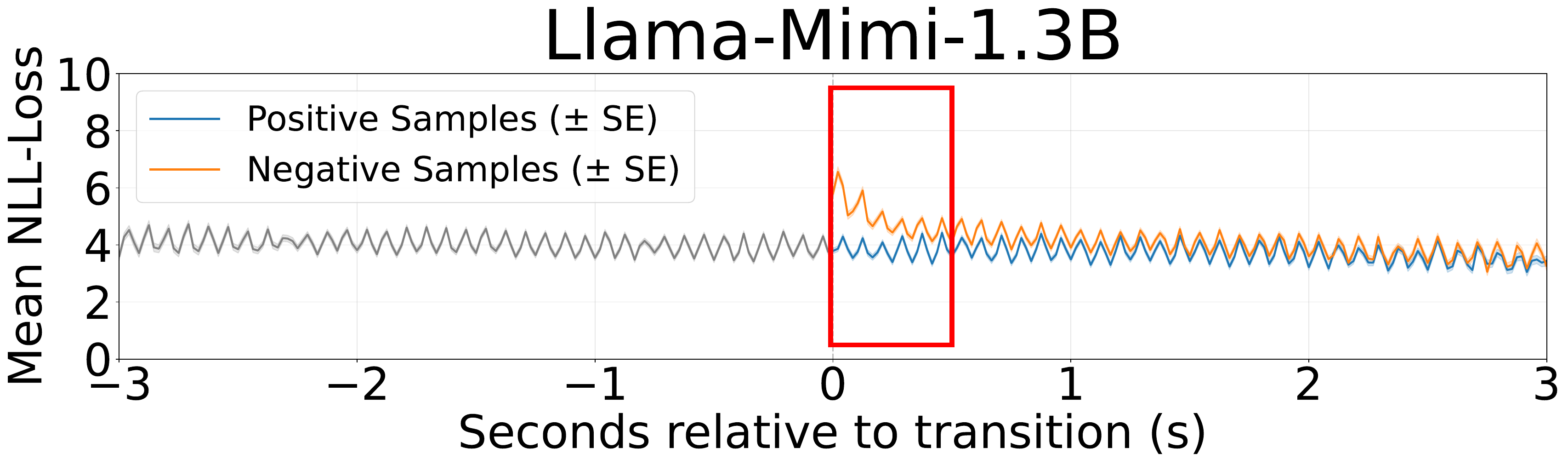}
    
  % \caption{Overall loss response breakdown of an exemplar spoken language model \texttt{Llama-mimi} on all \texttt{SALMon} samples, where negative samples are with acoustic discontinuity (illustrated by the waveform plot). 
  % %An abrupt transition in acoustic characteristics such as sentiment, speaker, or background occurs at $t=0$, which is reflected by a loss spike concentrated within a short temporal window. 
  % Each negative example has an abrupt transition in acoustic characteristics such as sentiment, speaker, or background that occurs at $t=0$, which causes a loss spike concentrated within a short temporal window.
  % Perplexity calculation exerted over the full time span (Sec.~\ref{sec:method-ppl}) aggregates losses well beyond this pulse-like response, motivating alternative evaluation methods (Sec.~\ref{sec:method-altppl},~\ref{sec:method-cont}).
  % \textcolor{red}{Standard error, x-axis, y-axis ppl-uncertainty, cut off ppl when volatile}
  % } %
  \caption{\textbf{Acoustic discontinuity disturbs negative log-likelihood loss (NLL) responses locally.} Top: SALMon samples consist of a shared prompt and a separate continuation, where positive samples maintain acoustic consistency, and negative samples contain abrupt acoustic transitions. Bottom: NLL response of Llama-Mimi-1.3B with standard error margins. Response on negative samples show localized spike within a short temporal window after the transition in contrast to the positive sample. Global token perplexity aggregates likelihood contributions outside this localized region (Sec.~\ref{sec:method-global-ppl}), making it susceptible to long-range loss volatility, thereby motivating our localized and normalized evaluation methods (Sec.~\ref{sec:method-altppl},~\ref{sec:method-cont})} 
  \label{fig:ppl-comp-vis}
\begin{comment}
Overall loss response breakdown of an exemplar spoken language model \texttt{spiritlm-expressive} on 1200 \texttt{SALMon} samples with acoustic discontinuity. An abrupt transition in acoustic characteristics such as sentiment, speaker, or background occurs at $t=0$, which is reflected by a loss spike concentrated within a short temporal window. Perplexity calculation exerted over the full time span aggregates losses well beyond this pulse-like response and is therefore heavily influenced by the volatility of semantic (HuBERT) contributions.
\end{comment}
\end{figure}

\begin{figure*}[t!]
  \centering
  % tighten spacing between cells
  \setlength{\tabcolsep}{4pt}
  \renewcommand{\arraystretch}{0}
  \begin{tabular}{c c c c}
  \multirow{2}{*}[6ex]{\rotatebox[origin=c]{90}{\small Mean NLL Loss}} &
  \includegraphics[width=0.3\textwidth]{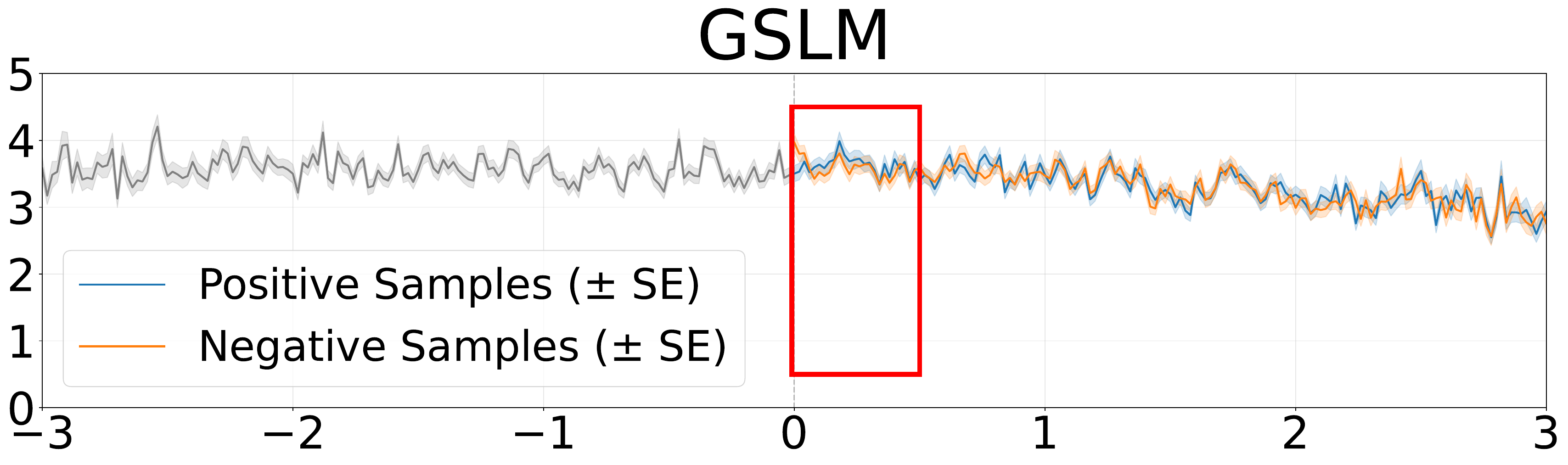} &
  \includegraphics[width=0.3\textwidth]{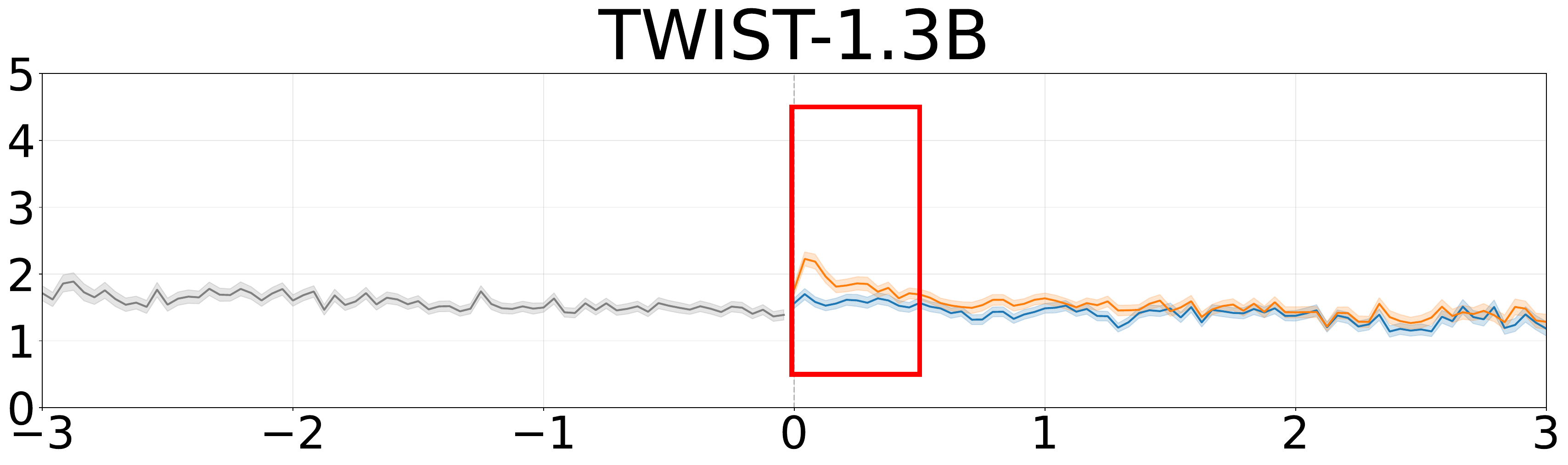} &
  \includegraphics[width=0.3\textwidth]{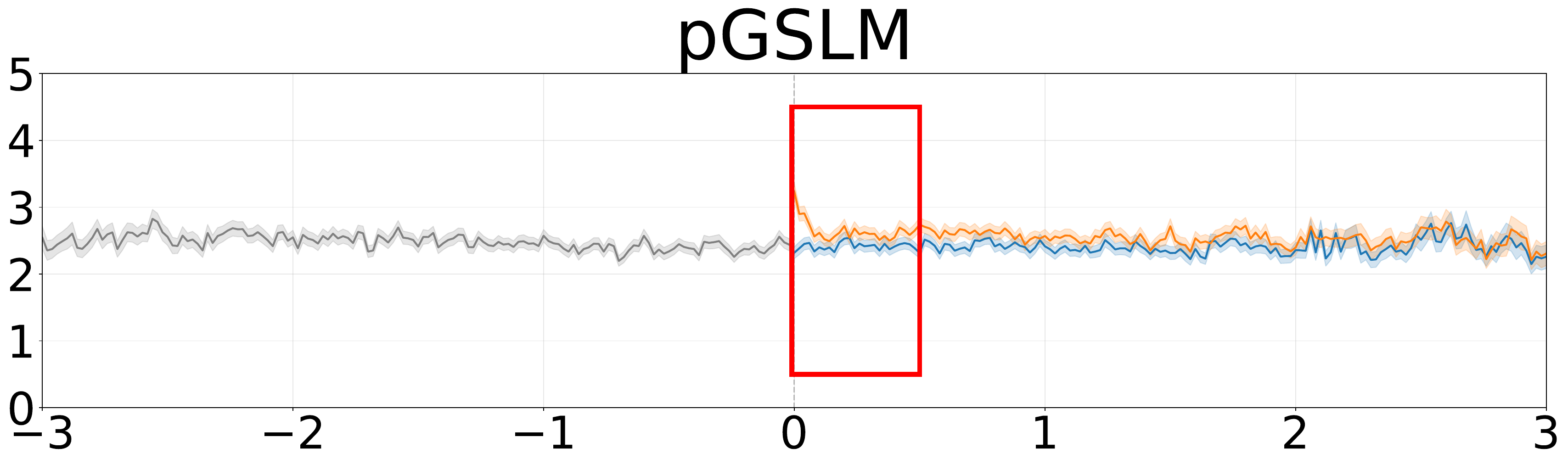} \\
  &
  \includegraphics[width=0.3\textwidth]{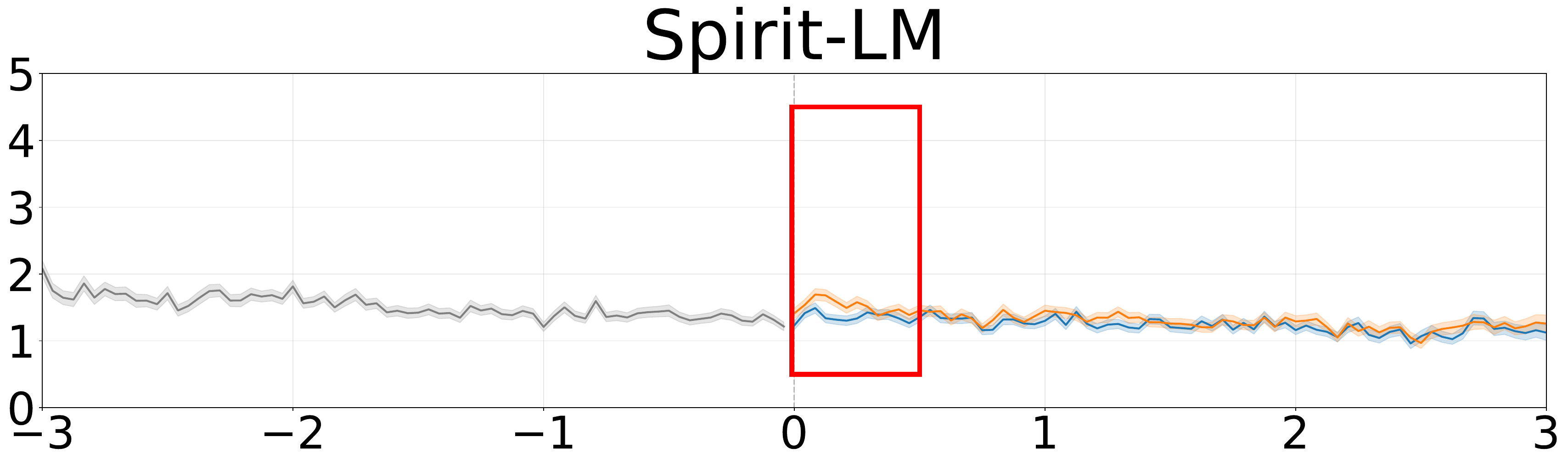} &
  \includegraphics[width=0.3\textwidth]{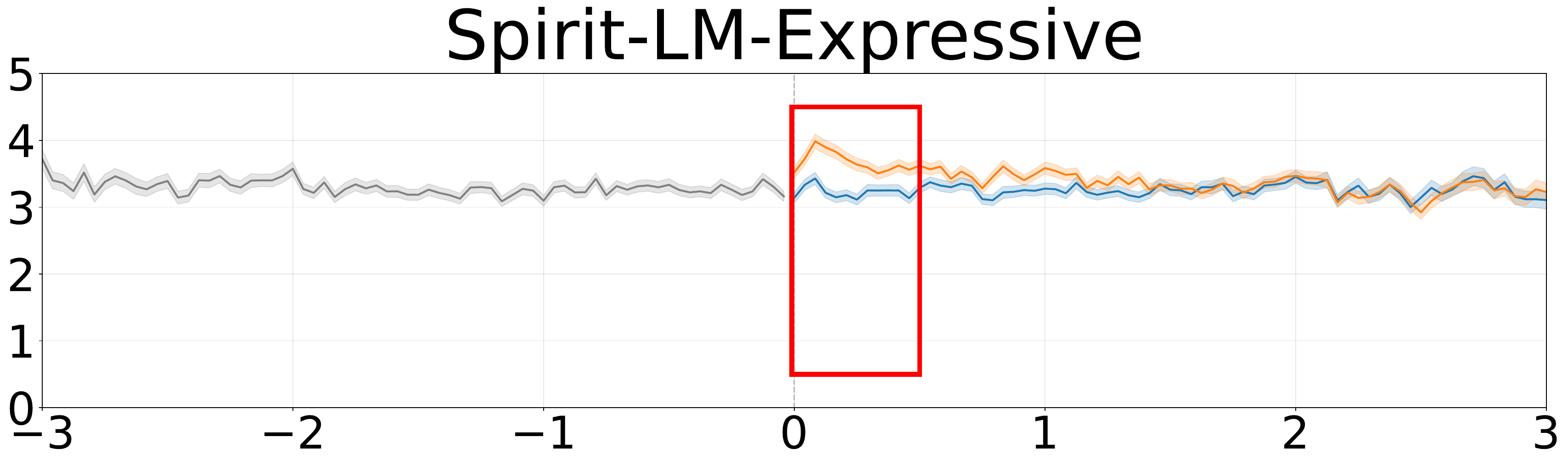} &
  \includegraphics[width=0.3\textwidth]{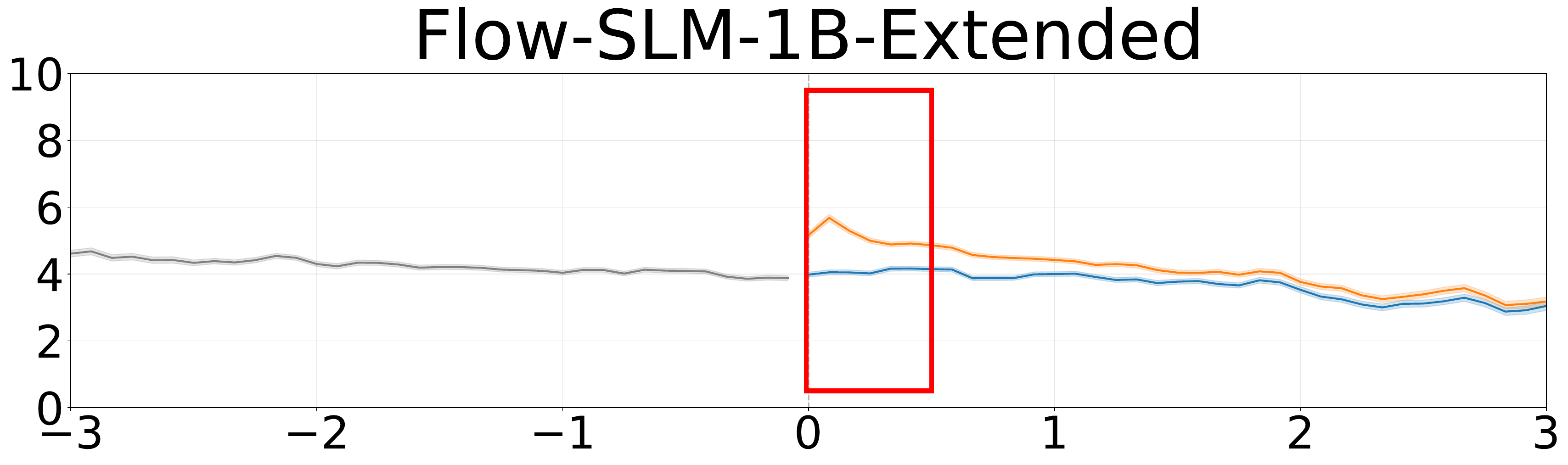} \\
  \multicolumn{1}{c}{} & \multicolumn{3}{c}{\small Seconds relative to transition (s)} \\
\end{tabular}
  \caption{NLL response of various models on SALMon  samples with standard error margins. High-scoring models on SALMon (e.g., Flow-SLM) exhibit localized NLL spikes for negative samples within a short temporal window after the transition. This behavior is less apparent in lower-performing models such as GSLM.}
  \label{fig:ppl-vis}
\end{figure*}

% Large-scale pretraining is central to extracting high-level knowledge from raw data; however, direct evaluation of the pretrained model remains challenging. 
% During the pre-training stage, the model learns to extract high-level knowledge from raw data, which ultimately benefits overall model performance. Consequently, the  quality of the pretrained model is of central importance. 
% However, evaluating pretrained models remains challenging, as assessing whether a generation is reasonable or coherent is inherently subjective. 

During the \emph{pre-training stage}, the model acquires foundational knowledge from raw data, which is directly reflected in its generation capabilities. Consequently, the quality of the pre-trained model lays the foundation for downstream performance \cite{raffel2020exploring, twist}. However, evaluating the generation quality of pre-trained models remains challenging, as assessing whether a generated output is plausible or coherent is inherently subjective.

% Generation can be approx. by cmp. pos. and neg.
%In text modeling, a common practice is to adopt \textit{perplexity}, a likelihood-based metric that scores model predictions on predefined token sequences. The interpretation of perplexity is more meaningful in comparative settings. For cross-model comparative setting lower perplexity on positive sentences is generally preferred \cite{kaplan2020scaling, hoffmann2022training}. For single model, comparitve setting consists of positive samples (good) and negative samples (include weirdness)

% Text Modeling (Perplexity)
In text modeling, a common practice is to adopt \emph{perplexity}, a likelihood-based metric that scores model predictions on textual token sequences. Likewise, pre-trained SLMs are often evaluated by computing perplexity over an entire sequence of discrete speech tokens. We refer to this metric as \textbf{``global token perplexity,''} which quantifies the sequence-level likelihood of a speech signal in the discrete token space.
In text modeling benchmarks, perplexity is typically interpreted in a comparative setting~\cite{brown2020language}, where each benchmark entry contrasts positive samples (i.e., fluent, coherent text) with negative samples (i.e., syntactically or semantically corrupted text). The model’s ability to assign higher likelihood to positive samples suggests a systematic preference for well-formed outputs during generation.
%Recently, these principles have been carried over to the benchmarking of pre-trained SLMs, where global token perplexity is adopted as a proxy for speech modeling quality.
The principles in text modeling have been carried over to the benchmarking of pretrained SLMs, where sWUGGY \cite{dunbar2021zerospeech}, sBLIMP \cite{dunbar2021zerospeech}, and SALMon \cite{maimon2025salmon} likewise adopt a comparative evaluation design. Each benchmark's contrastive pairs highlight various aspects of speech, from context to acoustics. 
While effective, directly computing global token-level perplexity may overlook characteristics unique to speech, risking misalignment with human perception.
From a cognitive perspective, textual and acoustic generation processes demand different attentional spans. 
% The \textit{Dependency Locality Theory} \cite{gibson1998locality, gibson2000dlt} and \textit{Surprisal Theory} \cite{levy2008expectation, smith2013effect} show that validation of proper textual generation requires \textbf{long-range attentiveness}. In contrast, non-semantic acoustic features are confined by signal continuity and regularized by \textit{Gradualness of Change} \cite{bregman1993auditory}, which entices \textbf{short-spanned conditioning}.
The \textit{Dependency Locality Theory} \cite{gibson1998locality, gibson2000dlt} and \textit{Surprisal Theory} \cite{levy2008expectation, smith2013effect} suggest that generating and evaluating coherent text relies on tracking dependencies over long contexts, thus requiring
\emph{long-range} attentiveness. In contrast, non-semantic acoustic features evolve continuously over time and are regularized by the \textit{Gradualness of Change} \cite{bregman1993auditory}, which favors \emph{short-span} conditioning.
 Likelihood patterns in Figure~\ref{fig:ppl-comp-vis} and \ref{fig:ppl-vis} echo these theoretical insights. The perplexity difference between positive and negative samples, when present, is short-spanned and concentrated near the onset of divergence.

In this work, we introduce novel evaluation protocols of SLMs that address fundamental asymmetries between speech and text modeling and place greater emphasis on local context sensitivity. 
%In addition, we propose generation-based evaluation methods that assess model behavior through controlled speech generation.
Specifically, we propose two families of evaluation methods and re-evaluate SLMs on SALMon.
The first family, \textbf{likelihood-based methods}, reformulates perplexity through localization and normalization, preserving the likelihood-based evaluation paradigm while emphasizing local context sensitivity (Sec.~\ref{sec:method-altppl}, Sec.~\ref{sec:exp-altppl}).
The second family is \textbf{generation-based methods}, where we evaluate on real generations produced by the SLM, since successful generation inherently demands sensitivity to local context (Sec.~\ref{sec:method-cont}, Sec.~\ref{sec:exp-cont}).  
%A model-based judge is used to determine generation success, using SALMon as reference.
New scores from these evaluation protocols yield substantially different conclusions from those based on naive global token perplexity.
To assess perception-faithfulness of the proposed methods, we conduct human-subject rating experiments on SLM-generated samples and collect mean opinion scores (MOS), which serve as the gold-standard reference for SLM performance.
Correlation analyses (Sec.~\ref{sec:analysis-corr}) show that our proposed methods align more closely with human ratings (Pearson correlation: $0.64 \rightarrow 0.8$, Spearman correlation: $0.67 \rightarrow 0.8$), thereby establishing a new paradigm for evaluating spoken language models.
%, with Pearson correlation increasing from $0.64$ to $0.80$, Spearman correlation from $0.67$ to $0.80$.
Our evaluation reshapes the performance landscape of SLMs: when evaluated correctly, the best-performing model closes 83\% of the gap to the human topline on SALMon, surprisingly achieving a new SOTA. %yielding conclusions that differ substantially from those based on naive global token perplexity.

\section{Background}
\begin{comment}
\subsection{Spoken Language Models}
Spoken Language Models have emerged as a holistic term encompassing a broad range of model architectures and functionalities. Prior work has charted their hierarchical organization, at the apex of which lies the idea of a \textit{universal speech processing system}—defined by its ability to (i) facilitate speech input and outputs (ii) tackle arbitrary spoken-language tasks  \cite{arora2025landscape}. At the foundation of this hierarchy are modality-specific model families, distinguished by their input-output configurations. \textbf{Pure speech LMs} operate exclusively on the speech modality \cite{gslm, pgslm, lin2024align}, typically trained using next-token prediction on unlabeled speech code sequences. \textbf{Speech-aware text LMs} take both speech and text as input, but restrict their output to text, thereby orienting the model use towards audio comprehension purposes \cite{chu2024qwen2audio, tangsalmonn, lu2025desta}. \textbf{Speech-text language models} are the most versatile, which support both speech and text for input and output, thereby enabling fully multimodal interactions and conversational abilities \cite{defossez2024moshi, spiritlm,  xie2024mini, ding2025kimi}. Architecturally, a central sequence model forms the core, with neural speech encoders and decoders attached as needed to map audio to and from compact, model-ready representations \cite{arora2025landscape}. 
\end{comment}

\subsection{Evaluation of Pretrained Spoken Language Models}
For pretrained SLMs, likelihood-based evaluation provides a principled way to probe speech consistency across multiple dimensions. 
Prior work has largely emphasized the content aspect, where sWUGGY and sBLIMP \cite{dunbar2021zerospeech} assess semantic coherence and lexical well-formedness. With automatic speech recognition (ASR), text-level perplexity can also be computed \cite{gslm, twist, wu2023speechgen}. More recently, paralinguistic modeling capabilities of pretrained SLMs are receiving growing attention, and SALMon \cite{maimon2025salmon} is established to measure the consistency of acoustic attributes. These include speaker identity, sentiment, background, and room conditions. In this paper, we focus on improving the evaluation of pre-trained SLMs specifically on acoustic attributes.

\subsection{Global Token Perplexity in Likelihood Modeling}
\label{sec:method-global-ppl}
Likelihood modeling offers a principled approach to evaluate sequence models.
In conventional practice, likelihood modeling is assessed via \emph{perplexity}, where higher perplexity indicates lower likelihood under the model. Given a sequence $\mathbf{s}$, perplexity is defined as the exponential of the negative log-likelihood loss (NLL): \(\mathrm{PPL}(\mathbf{s}) = \exp\!\left(\mathrm{NLL}(\mathbf{s})\right)\).
With this transformation, we will henceforth report perplexity in terms of NLL. In the context of SLMs, perplexity is computed over sequences of \emph{discrete speech tokens} \cite{maimon2025salmon, sugiura2025llama, chou2025flow}. We refer to this formulation as \emph{global token perplexity}. Formally, given a sequence $s = (x_1, \ldots, x_T)$, the $\mathrm{NLL}_{\mathrm{global}}$ formula yields

\begin{equation}
\label{eq:nll-global}
\scalebox{1}{$
%U(s) \propto 
\mathrm{NLL}_{\mathrm{global}}(s)
= \frac{1}{T}\sum_{t=1}^{T} - \log p(x_t \mid x_{<t})
$}
\end{equation}

Modern discrete
speech units
often consist of multiple
codebooks, as in hierarchical residual vector quantization (RVQ) architectures \cite{zeghidour2021soundstream, wang2023neural, defossezhigh, defossez2024moshi} or are multi-channeled, as in joint speech-text modeling \cite{tseng2025taste}. In such settings, we flatten the multi-channel likelihood outputs into a single serialized stream. This formulation treats all tokens across channels and time steps as equally informative, yielding a unified, holistic likelihood estimate for the entire speech signal.
\begin{comment}
global token perplexity further averages over channels in addition to time:
\begin{equation}
\scalebox{0.9}{$
\text{NLL}_{\text{global}}(\mathbf{s}) = \frac{1}{TC} \sum\nolimits_{t=1}^{T}\sum\nolimits_{c=1}^{C} -\log p(x^c_t \mid x^c_{<t})
$}
\end{equation}
where $C$ denotes the total number of channels. In what follows, we discuss additional perplexity variants under the single-channel formulation (Eq.~\ref{eq:nll-global}), where extention to multi-channel is trivial.
\end{comment}

With this definition, adopting perplexity to benchmarking is straightforward. In SALMon, each sample consists of a pair of sequences $(s_p, s_n)$, corresponding to the positive (preferred) sequence and the negative (dispreferred) sequence. A perception-faithful model is expected to assign lower perplexity to $s_p$ than to $s_n$. 
This comparison is equivalent in the NLL space as NLL is a strictly monotonic transformation of perplexity:
\begin{equation}
%P(s_p) > P(s_n) \quad \text{or} 
\quad \text{NLL}(s_p) < \text{NLL}(s_n)
\end{equation}

\subsection{Likelihood Modeling Calibration}

A fundamental principle of likelihood modeling is that a model’s predicted probabilities should align with the frequencies of occurrence of the global distribution \cite{kadavath2022language, ulmer2022exploring}. 
%when deviated, these models need to go. For auto-regressive modeling, a held-out corpus functions as a gold-standard reference, while the numeric score "perplexity"  provides a scalar measure of the model’s on-distribution likelihood assignment \cite{brown2020language}. In task-based generations with confined options (binary choice, multiple choice), a normalized probability distribution over the competing options is adopted so that the predicted probabilities lie on the same scale as empirical answer frequencies. 
When raw prediction scores deviate, calibration methods can be introduced to increase alignment.
Certain calibration approaches, such as Platt scaling \cite{platt1999probabilistic, guo2017calibration}, are monotonic and rank preserving. 
In contrast, other methods aim to alter the model’s selection, including option debiasing \cite{brown2020language}, option finetuning \cite{guo2017calibration}, and decoding-time interventions \cite{chuang2024dola}. 
In this work, we extend calibration efforts to SLMs by proposing localization and normalization methods that better align with the characteristics of speech.

\section{Proposed Method}
Using global token perplexity to assess acoustic consistency can lead to a measurement fallacy that does not correlate with human assessments.
Empirically, prior works have shown that perplexity in speech is often disproportionately influenced by semantic factors \cite{maimon-adi-2023-speaking, sicherman2023analysing, polyak2021speech}, limiting the expressiveness of acoustic features. These semantic contributions constitute a substantial part of global token perplexity, which introduces noise in modeling and cascades into fluctuations during sample-wise comparisons. SALMon reduced semantic volatility by ``enforcing the same speech context between comparisons,'' but strict attribute-wise independence is neither well defined nor practically attainable in speech.
We therefore derive alternative perplexity variants that focus on modeling the target attribute—the axis along which the positive and negative speech samples are contrasted.

\subsection{Proposed Likelihood-Based Evaluation}
\label{sec:method-altppl}
%The characteristic of speech continuity reveals a crucial flaw of modeling acoustic properties of speech using perplexity calculated over the global span. Unlike textual modeling where repeating the last words render uninformative, approximating $x_t$ with $x_{t-1}$ is already a good inductive bias in acoustic feature space. As a result, speech tokens which carry information of both has the majority of the uncertainty attributed to model the semantic aspect, and using it to estimate the acoustic aspect is thus subject to great volatility. We theorize that instantaneous perplexity should be much more significant than global perplexity. 
% Using global perplexity to assess non-semantic attributes risks a measurement fallacy. 
% Using global token perplexity to assess non-semantic attributes risks a evaluation fallacy. 
% Even when evaluation varies only a single attribute at a time (e.g., speaker), the perplexity signal attributable to that target factor can be overshadowed by contributions from non-target attributes. Empirically, prior work have shown that perplexity in speech is often disproportionately influenced by semantic factors \cite{maimon-adi-2023-speaking, sicherman2023analysing, polyak2021speech}, obscuring expressiveness of acoustic features. Volatility of perplexity from non-target attribute is inevitable as strict attribute-wise independence is neither well defined nor practically attainable, which cascades into fluctuations during sample-wise comparisons.

We start from Equation~\ref{eq:nll-global} and derive windowed, localized and normalized variants of token perplexity that focus on better capturing the response of SLM on acoustic discontinuity. 
The windowed perplexity variant replaces global aggregation over the full NLL array with a predefined short temporal window that slides over the speech sequence, producing a sequence of windowed perplexity values. Then, we take the maximum among these values to obtain the perplexity spike as a measure of local anomaly, allowing comparisons across samples to determine which one is more atypical.
\begin{equation}
\scalebox{0.93}{$
\begin{aligned}
\operatorname{NLL}_{\text{windowed}}(\mathbf{s})
&=
\max\limits_{i}
\frac{1}{\delta}
\sum_{t=i}^{i+\delta-1}
\bigl(-\log p(x_t \mid x_{<t})\bigr),
\end{aligned}
$}
\end{equation}
where $1 \le i \le T-\delta+1$ indexes all valid starting positions of a length-$\delta$ window in the token sequence. Since it is the temporal window that is fixed across all models, the corresponding token-based value of $\delta$ is derived accordingly and therefore differs across models; in some cases, it varies dynamically within a model.
%Table~\ref{tab:ppl-main} shows windowed perplexity adopted in both acoustic consistency and semantic-acoustic alignment subsets. 

In SALMon, each positive–negative pair shares a common prefix, which allows us to capitalize on the diverging timeframe as additional information. For each pair $(s_p, s_n)$, we extract the longest common prefix as the prompt ($S$), the remainder of the sequence forms the positive ($P$) and negative ($N$) responses, respectively, analogous to QA-style setups in NLP benchmarks \cite{brown2020language, hendrycksmeasuring}. Concretely, we write $s_p = S \frown P$ and $s_n = S \frown N$, where $\frown$ denotes sequence concatenation. The localized variant of token perplexity only accounts for information in a localized window of length $\delta$ starting from the timeframe where the speech prompt ends ($t_p$):

\begin{equation}
%U(\mathbf{s}) \propto
\scalebox{0.93}{$
\text{NLL}_{\text{localized}}(\mathbf{s})
= \frac{1}{\delta}\sum_{t=t_p}^{t_p+\delta-1}\bigl(-\log p(x_t\mid x_{<t})\bigr)
$}
\end{equation}

We also consider normalization,
which deducts the prompt-free probability of each response \cite{brown2020language}. Normalization applies to both global and localized perplexity by choosing the corresponding window $\Delta \in \{T, \delta\}$, where $T$ represents a value sufficiently large that it is ultimately bounded by the sequence length. The normalized perplexity aggregates normalized probabilities at each time step.
\begin{equation}
%U(\mathbf{s}) \propto 
\scalebox{0.93}{$
\text{NLL}_{\text{normalized}}(\mathbf{s})
= \frac{1}{\Delta}
\sum_{t=t_p}^{t_p+\Delta-1}
\bigl(
-\log\frac
{p(x_t\mid x_{<t})}
{p(x_t\mid x_{t_p:<t})}
\bigr)
$}
\end{equation}

% For models with non-uniform per-token durations (e.g., due to deduplication \cite{spiritlm,  pgslm, gslm, twist}), we cut-off tokens by the implied time index. For $\text{NLL}_{\text{windowed}}$, we reduce complexity by using the duplicated time-uniform token sequence. This results in a duration-weighted loss where token contributions is amortized by its span.

%\subsection{Measuring Consistency with Model Continuations}
%\subsection{Proposed Evaluations on Model Continuations}
\subsection{Proposed Generation-Based Evaluation}
\label{sec:method-cont}
The evaluation methods in 
Sections~\ref{sec:method-global-ppl} and~\ref{sec:method-altppl} compute the likelihood of the samples rather than letting the SLM continue the speech.
Here, we directly evaluate on continuations of SLMs given a speech prompt $S$, which provide multiple benefits. With real continuations, it is possible to conduct human evaluations to obtain mean opinion scores (MOS), which provide a perception-faithful estimate of model quality and can serve as the reference for the model’s \textit{true continuation performance}. Second, by approximating human judgements with scores from model-as-a-judge, we obtain another evaluator candidate to compete with global token perplexity. 

Given a speech prompt $S$, we define a continuation $G$ sampled from model $M$ by
\begin{equation}
G = M(\cdot \mid S).
\end{equation}

For human evaluations, annotators assign a quality score between 1 and 5 based on how good the generated continuation is relative to the positive continuation reference ($P$).

For scoring continuations with model-as-a-judge, success can be determined more straightforwardly using a contrastive criterion: a continuation is deemed correct if it is closer to the gold positive continuation than to the negative one.
\begin{equation}
\label{eq:distance}
  d(G, P) < d(G, N)
\end{equation}
%We then evaluate each generated continuation ( G ) using established benchmark criteria (e.g., factuality, coherence, preference scores), yielding ground-truth labels or scores that reflect real continuation performance. By correlating these with our proxy-based estimates computed on the same prompts ( S ), we quantify how faithfully the proxies predict downstream continuation behavior, independently of gold references.
%given how SALMon is structured such that P and N only differs in the attribute of consideration, 
where $d(\cdot)$ denotes a distance function. The following section describes the procedure for selecting a qualified model to serve as an automatic judge, as well as the corresponding evaluation strategy to assess SLM-generated continuations (Sec.~\ref{sec:judge-select}).

\subsection{Model-as-a-Judge for Scoring SLM Continuations}
\label{sec:judge-select}
To select an appropriate judge \(J\), we require (i) a labeled set with known correctness and (ii) a model that assigns a distance score to a prompt--response pair. Fortunately, the shared prompt \(S\) and paired responses \((P,N)\) in each contrastive example \((s_p,s_n)\) provides a natural labeled set with the following objective:
\begin{equation}
\label{eq:distance-gold}
d(S, P) < d(S, N).
\end{equation}

We explore using embedding models $E$ as judge candidates, leveraging their inherent distance metric, taking inspiration from retrieval systems \cite{feng2022language}:
\begin{equation}
 \operatorname{d}(A,B) = 1 - \cos\!\big(E(A), E(B)\big)
\label{eq:embed-obj}
\end{equation}

%While generative speech models, even proprietary ones, suffice to provide the uncertainty estimation (Sec~\ref{sec:method-ppl}), the mapping from probability to distance-based metrics is non-trivial, making it inherently difficult to calibrated the model for  scoring generations.

%Instead, we directly approximate probabilities using similarity scores. Inspired by semantic retrieval systems \cite{}, we define a scoring function based on embedding distances using a embedding model $E$:

\noindent where $\cos(\cdot, \cdot)$ denotes cosine similarity and $E(\cdot)$ denotes the forward pass through the embedding model. 
Plugging Equation~\ref{eq:embed-obj} into Equation~\ref{eq:distance} yields the explicit sample-wise objective for choosing $E$ as a qualified judge.

\begin{equation}
\label{eq:judge-dev-cmp}
\cos\!\big(E(S), E(P)\big) > \cos\!\big(E(S), E(N)\big) 
\end{equation}

%The interchangeability between the arguments in the distance functions assumes elimination of the covariance term of $(S, P)$ or $(S, N)$. 
The aggregation of correct predictions over the development set yields the accuracy. An ideal judge would achieve a perfect score; in practice, a more realistic qualification threshold is the human accuracy on the same benchmark. We begin with a comprehensive set of pretrained embedding models and narrow it to the best-performing model (over the qualification threshold), which we adopt as the final judge ($J$). The judge benchmarks continuations $G$ from SLM following Equation~\ref{eq:distance}:

%($J \coloneqq \operatorname*{argmax}_{E \in \mathcal{Q}} \; \mathrm{Score}(E)$).

% \begin{equation}
% \label{eq:judge-crit}
% \begin{aligned}
% Acc_{\mathrm{dev}}(E)
% &= \frac{1}{N}\sum_{i=1}^N \mathbf{1}\!\{ d_E(S_i,P_i) < d_M(S_i,N_i) \} \\
% &= \frac{1}{N}\sum_{i=1}^N \mathbf{1}\!\{ \cos(E(S_i),E(P_i)) > \cos(E(S_i),E(N_i)) \}
% \end{aligned}
% \end{equation}

% We define the automatic judge $J$ as the that achieves the highest development accuracy while surpassing the human-annotated accuracy:

% {\footnotesize
% \[
% J \coloneqq 
% \begin{cases}
% \displaystyle E^\star=\arg\max_{E \in \mathcal{E}} \mathrm{acc}_{\mathrm{dev}}(E), & \text{if } \mathrm{acc}_{\mathrm{dev}}(E^\star) > \mathrm{acc}_{\mathrm{human}}, \\[1em]
% \varnothing, & \text{otherwise.}
% \end{cases}
% \]
% }
%The existence of such a judge entails that modeling covariance between speech signals (e.g., $(S, P)$ or $(S, N)$) may not be necessary for the respective task, supporting speech modeling designs to pick up the isotropic speech feature from the judge \cite{}.

%\subsubsection{Evaluating Continuation Consistency}
\label{sec:judge-eval}

%Given the selected judge $J$, we evaluate whether each generated continuation $G$ is closer to its positive reference $P$ than to the negative reference $N$: %\textcolor{red}{delter}
\begin{equation}
\label{eq:judge-eval}
    \cos\!\big(J(G),J(P)\big) > \cos\!\big(J(G),J(N)\big),
\end{equation}
% \[
% \mathrm{Acc}_{\mathrm{testg}}
% =\frac{1}{N}\sum_{i=1}^N \mathbf{1}\!\{ d_J(G_i,P_i) < d_J(G_i,N_i) \}
% \]
% \[
% =\frac{1}{N}\sum_{i=1}^N \mathbf{1}\!\left\{ \cos\!\big(J(G_i),J(P_i)\big) > \cos\!\big(J(G_i),J(N_i)\big) \right\}.
% \]
where aggregation over the whole benchmark yields the accuracy score of the evaluated SLM. 

\begin{figure}[p]
%\left$(a)$
%$\left(a\right)$
\small{\raggedright(a1)\par}

  \centering
  % tighten spacing between cells

  \setlength{\tabcolsep}{4pt}
    \includegraphics[width=0.48\textwidth]{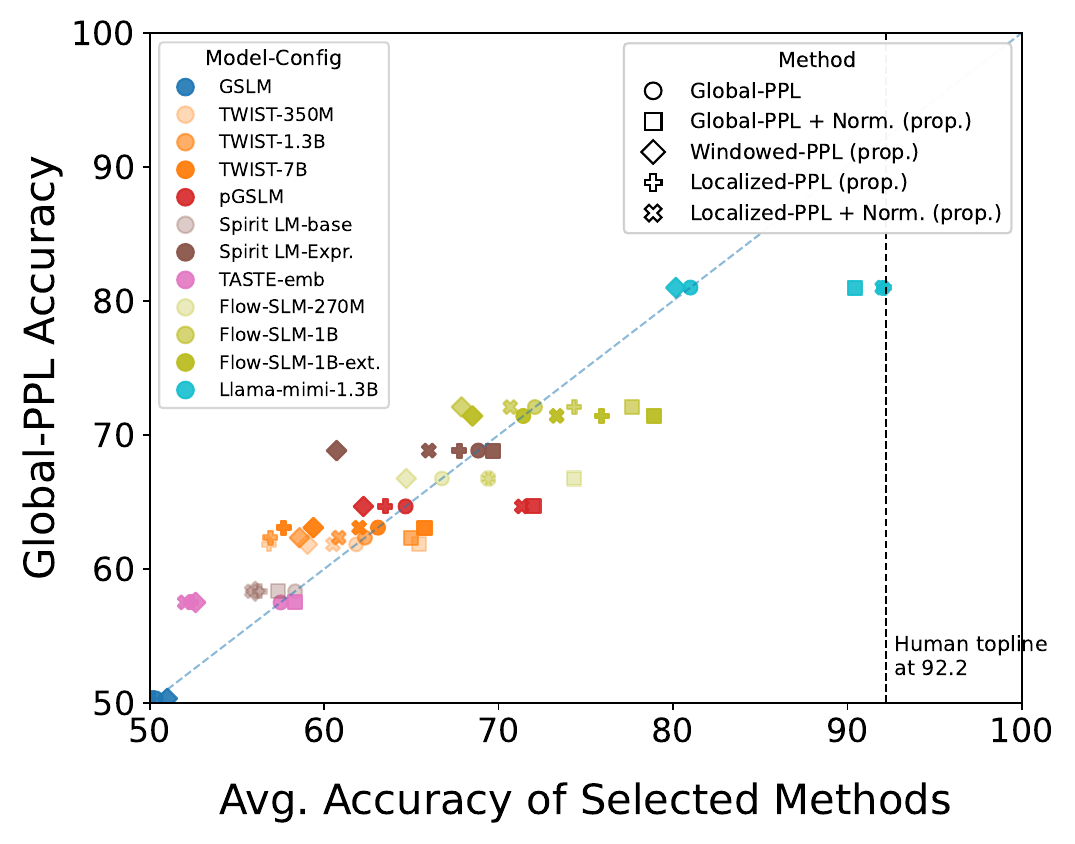}
    {\raggedright(a2)\par}
    \includegraphics[width=0.48\textwidth]{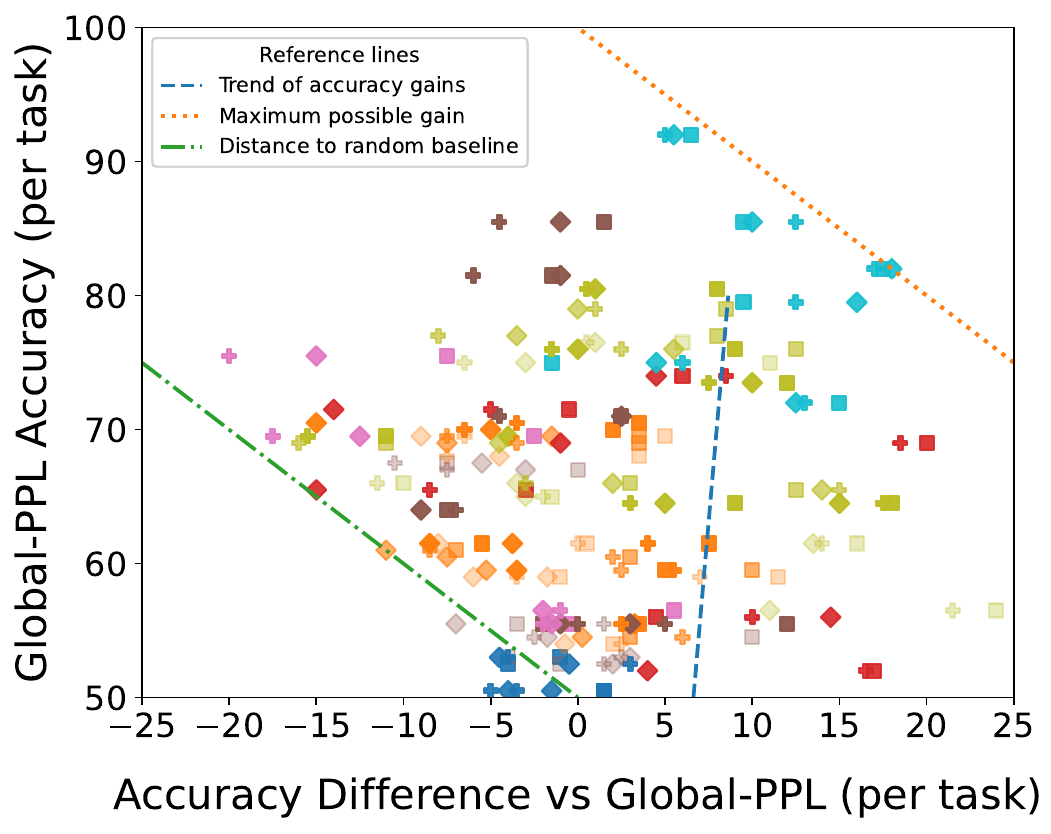}
    
    {\raggedright(b)\par}
    \includegraphics[width=0.48\textwidth]{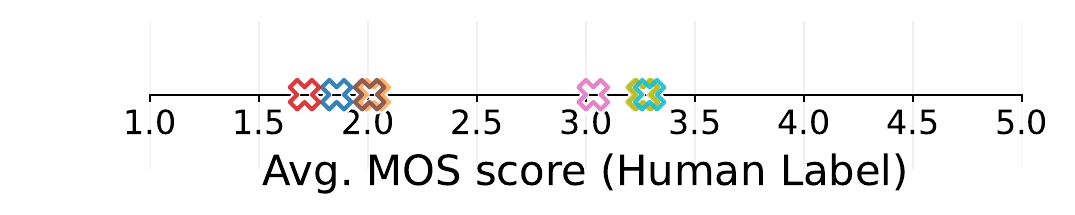}
    
    {\raggedright(c)\par}
    \includegraphics[width=0.48\textwidth]{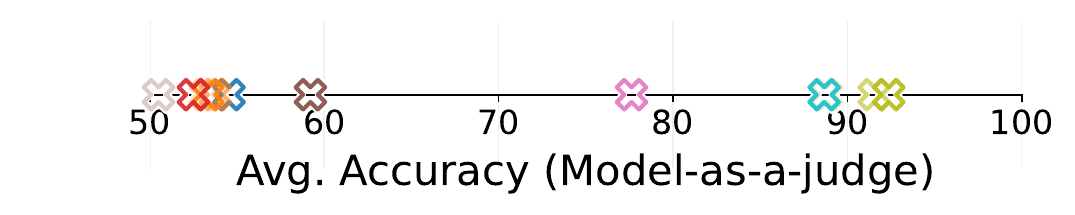}

 \caption{\textbf{Overall performance of spoken language models on consistency tasks.} The x-axis shows model accuracy (score) under different evaluators: (a1) alternative likelihood estimators, (b) MOS, and (c) embedding-as-a-judge, where model color codes are shown in (a1) and shared among all plots.  In (a1), we correlate scores from proposed methods against those from global token perplexity (\texttt{Global-PPL}); the horizontal spread highlights the discrepancy across evaluation methods. The alternative methods rate strong models more favorably than \texttt{Global-PPL}, substantially closing the gap to the human topline. In (a2), we correlate deviations from the proposed methods against \texttt{Global-PPL} scores. Deviations generally become larger at higher \texttt{Global-PPL} performance (blue), until it saturates due to the maximum performance ceiling (orange). Negative deviations exhibit a similar trend in absolute magnitude, though this is less surprising since they are soft-bounded by distance to random baseline (green).
 %Positive deviations become larger at higher \texttt{Global-PPL} performance (blue), until it saturates due to the maximum performance ceiling (orange). Negative deviations exhibit a similar trend in absolute magnitude, though this is less surprising since they are soft-bounded by distance to random baseline (green). 
 } 
  \label{fig:acc-acc}
\end{figure}
\TabExpHumanEval
\TabJudgeDevSmall

\section{Experimental Setup}
We adopt SALMon for all evaluations. SALMon includes 6 subsets that measure acoustic consistency in gender, speaker identity, sentiment, two background conditions, and room attributes. Each data point consists of a positive and a negative sample, where the negative sample contains an inconsistency in one of the attributes. Our evaluations of SLMs cover GSLM \cite{gslm}, TWIST \cite{twist}, pGSLM \cite{pgslm}, Spirit-LM \cite{spiritlm}, TASTE \cite{tseng2025taste}, Flow-SLM \cite{chou2025flow}, and Llama-Mimi \cite{sugiura2025llama}. We measure their performance under original and proposed methods. The localized window $\delta$ is set as 0.5s.

We use the Prolific service to obtain MOS scores. We evaluate 50 samples over 9 models, which yields 450 generations to be evaluated. Each generation is independently assessed by five annotators on a five-point Likert scale. 
%focusing on only the target attribute, with 1 indicating the least coherent continuation and 5 the most coherent.
The annotators are proficient English speakers, and they are fairly compensated for their time. See Appendix~\ref{sec:apx-MOS-prompt} for the annotation guidelines provided to annotators.

For model-as-a-judge, we consider a diverse pool of embedding models trained with different objectives and  datasets, including TITANET \cite{koluguri2022titanet}, CAM++ \cite{wang2023cam++}, CLAP \cite{elizalde2023clap}, and AudioSet-trained models \cite{ARCH}. We use SALMon prompts $S$ as the dev set to select the best judge for each subset, and use the judge to obtain SLM performance scores from continuations $G$.

\section{Experimental Results}

\subsection{Localized and Normalized Likelihood-Based Evaluation}
\label{sec:exp-altppl}

We first examine how the proposed likelihood-based estimators reshape the performance landscape of spoken language models. Figure~\ref{fig:acc-acc} (a1) shows the average accuracy over all consistency benchmarks for each SLM–method combination, plotted against the corresponding accuracy measured by conventional global token perplexity. From the plot, it can be seen that the horizontal spread is quite large, indicating systematic disagreement between perplexity methods. The degree of disagreement is quantified in Figure~\ref{fig:acc-acc} (a2), which reveals a positive association between disagreement and model competence (measured by \texttt{Global-PPL} performance). The linearly regressed positive deviation (blue line) starts at $+6.63\%$  at around $50\%$ accuracy, and increases to $+8.62\%$ at $80\%$ accuracy, which is substantial. 
%indicating systematic disagreement with the global perplexity method.
%Figure~\ref{fig:acc-acc} (d) quantifies the disagreement on a per-task basis and reveals a positive association between disagreement and model competence (measured by \texttt{Global-PPL} performance). The regression estimates a $+5.67\%$ deviation at around $50\%$ accuracy, increasing to $+8.67\%$ at $80\%$ accuracy. This large deviation in the high-competence regime further highlights systematic inconsistencies in SLM performance reporting.

A closer inspection shows that the shift is predominantly one-sided for each SLM configuration, reflecting a stable bias toward either over- or underestimation. These patterns are tightly linked to the underlying token type. Our proposed methods consistently assign lower scores to HuBERT-based SLMs (GSLM, TWIST, SpiritLM), while in contrast yielding higher scores for Mimi-based models (Flow-SLM, Llama-Mimi). pGSLM is the lone outlier among HuBERT-based models, likely due to its distinctive auxiliary training objective.
%This behavior aligns with the loss profiles in Figures~\ref{fig:ppl-comp-vis} and~\ref{fig:ppl-vis}, where the separation between positive and negative samples, when present, emerges primarily in localized regions. Consequently, uncertainty estimators that emphasize locality and downweight input-independent factors yield improved performance.
In contrast, model families trained with the same recipe but varying only in scale exhibit little behavioral difference. We report the full set of scores in Appendix~\ref{sec:apx-exp-ppl}. %suggesting that model complexity is not the primary bottleneck for the speech continuation task.

\begin{comment}
To resolve the inconsistencies between methods, we treat “true task performance” as a regression target (Section~\ref{sec:exp-cont}). %regress these scores to the estimators’ fidelity to ``true task performance'', which we undertake in Section~\ref{sec:exp-cont}.
Importantly, even without regression, any improvement induced by the alternative methods is highly meaningful, as it serves as evidence for the existence of encoded information. The implications can be large: Llama-Mimi’s score rises from 80.92 to 92.08, effectively closing the perceived gap to the human topline. We report the full set of scores in Appendix~\ref{sec:apx-exp-ppl}.
\end{comment}

\begin{comment}
Next, we analyze performance difference on the task and method dimensions. In Figure~\ref{fig:acc-acc} (d), method disagreement is largest for sentiment and speaker attributes, exceeding 7\%, followed by gender and room consistency at roughly 6--7\%, with background attributes trailing at about 5-6\%. Performance estimation differences are rather uniformly high across methods, with only Windowed-PPL showing a lower difference. As seen in the bar plot, the performance difference is larger in the regime where Global-PPL indicates task competence of the SLM. This suggests that the reported effective scores of SLM become even less reliable.
\end{comment}

\subsection{Generation-Based Evaluation}
\label{sec:exp-cont}

We conduct experiments on actual model continuations to gain a better understanding of the generative abilities of SLMs.

\subsubsection{MOS Evaluations by Human Labelers}
\label{sec:exp-cont-mos}

\begin{figure*}[t!]
  \centering
\includegraphics[width=\textwidth]{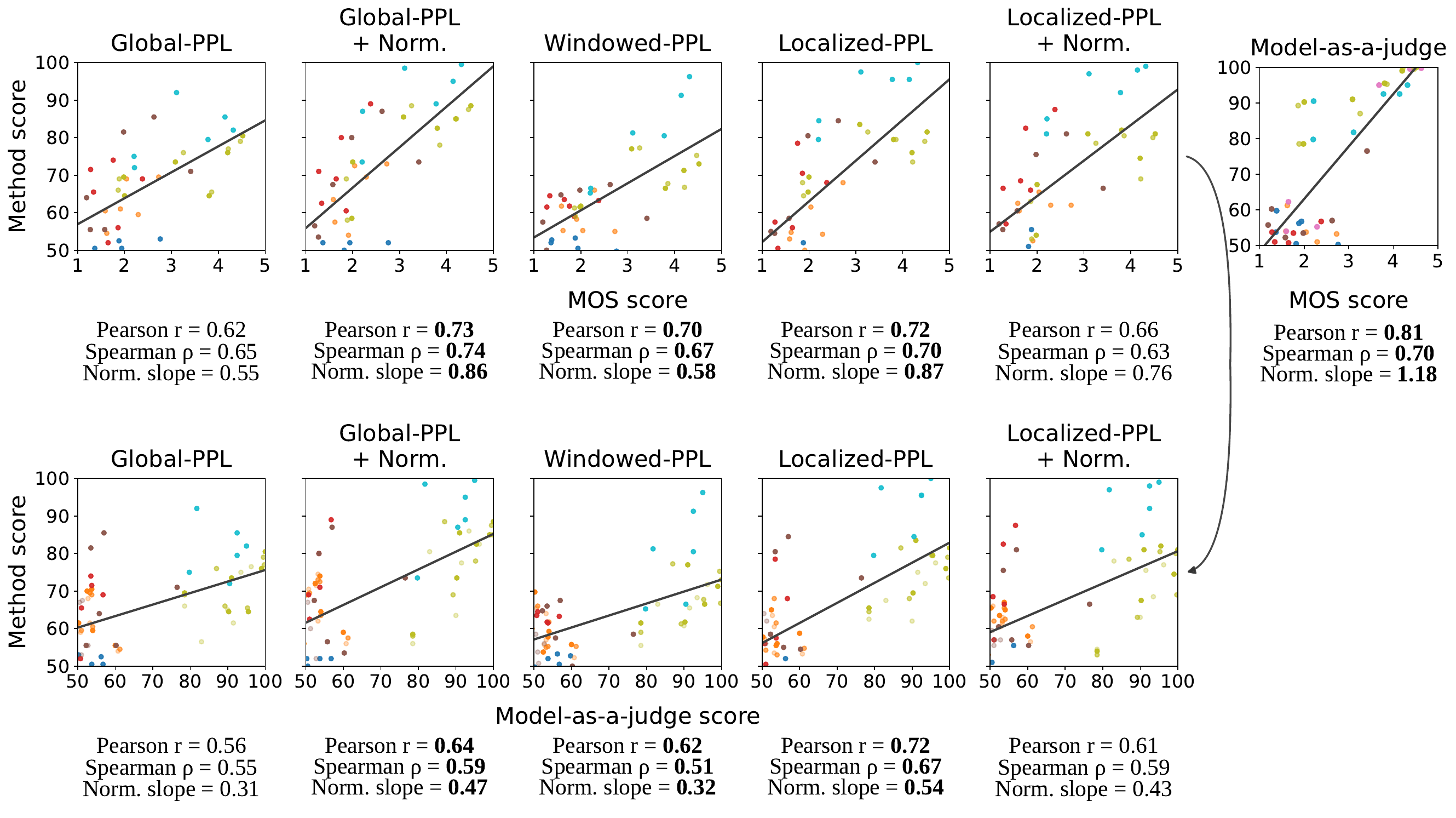}
\includegraphics[width=\textwidth]{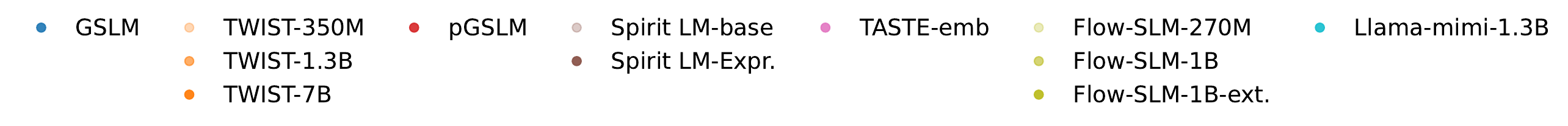}
  \caption{Correlation between perplexity evaluation methods vs golden labels provided by either MOS scores (top), or model-as-a-judge proxies (bottom) on the SALMon benchmark. Compared with using global perplexity as the aggregator (leftmost), windowing, normalization, and localization are more effective alternative operators and show stronger correlation with MOS scores. Using model-as-a-judge on generations (top right) also exhibits higher correlation with the MOS scores.}
  
  \label{fig:correlation}
\end{figure*}

MOS evaluation results are presented in Figure~\ref{fig:acc-acc} (b). Llama-Mimi obtains the top score of 3.29, followed by Flow-SLM, while models using HuBERT tokens (GSLM, TWIST, Spirit-LM, pGSLM) perform much worse. Full model-by-task results are reported in 
%Appendix~\ref{sec:apx-MOS} and suggest that stronger models improve primarily on speech-centric attributes (e.g., sentiment, speaker, and gender) as opposed to ambience-related information. 
Table~\ref{tab:exp-human-eval}, which presents the quantitative MOS scores across models and tasks. The results suggest that stronger models improve primarily on speech-centric attributes (e.g., sentiment, speaker identity, and gender), while substantial room for improvement remains in modeling ambience-related information.

%These observations resemble closely to ones found in perplexity methods. We show detailed correlation results between all perplexity methods in the next section.

\subsubsection{Model-as-a-Judge for Measuring Generation Consistency}
\label{sec:exp-cont-judge}

\textbf{Identification of suitable judge models.}  Table~\ref{tab:judge-dev-small} shows the top performing embedding model on the SALMon dev set along with its accuracy scores, following Equation~\ref{eq:judge-dev-cmp}. Results reveal that the selected embedding models consistently outperform human performance, and even reach ceiling-level performance ($>99\%$) in four out of six cases. Collectively, these results support the credibility of model-as-a-judge for the SALMon task using a combination of TITANET, HuBERT-large-audioset, and wav2vec2-large-audioset. Detailed model-by-task results are provided in Appendix~\ref{sec:apx-exp-cont-judge}.
%, which further reveal that no single embedding model aces all six attributes, motivating future work on more generalizable audio embedding approaches. 

\textbf{Generation consistency evaluated by qualified judge models.} 
Figure~\ref{fig:acc-acc} (c) shows the result of evaluating model generations on SALMon using the selected judge models. Consistent with findings in likelihood estimators and MOS scores, the speech tokenizer is the most dominant factor for the performance difference. 
Most HuBERT-based models struggle to retain speech properties during continuation, obtaining performance close to random choice. 
%A notable exception is SpiritLM-expressive, where additional pitch and style tokens assist in preserving sentiment information. 
On the other end of the spectrum, Flow-SLM and Llama-mimi exhibit strong performance, with scores in the vicinity of the human topline. Finally, TASTE generations perform relatively well, highlighting the importance of adopting a speaker vector during token-to-speech conversion. In Appendix~\ref{sec:apx-exp-cont-judge}, we present verbose task-wise results, which further suggest that continuation failures principally arise from inadequate information being preserved during the speech encoding phase.

\TabKendallTau

\subsection{Correlation between methods}
\label{sec:analysis-corr}
We now have four likelihood-based evaluators and a generation-based evaluator scored with an embedding judge model.
To determine the metrics that are more faithful to human perception for this task, we correlate them to the ``true continuation performance,'' provided by the MOS scores. The top row of Figure~\ref{fig:correlation} is a comprehensive display of these correlations. Global token perplexity sets the baseline with Pearson score of 0.62 and Spearman of 0.65. Localized perplexity achieves higher scores of Pearson of 0.72 and Spearman of 0.70, which are further surpassed by normalized perplexity, where both Pearson and Spearman correlations improve to 0.73 and 0.74 respectively. Continuations scored by embedding judges obtain the highest Pearson score overall at 0.81, with a slightly lower Spearman score of 0.70. The high Pearson correlation is best accounted for by the tighter dispersion of points toward high-performance regime. In addition, the regression slopes for normalized and localized perplexity are substantially closer to 1\footnote{The optimal slope is obtained by linearly anchoring the random-performance baseline, with accuracy = 50, to MOS = 1, and the perfect-performance ceiling, with accuracy = 100, to MOS = 5.}, indicating that they not only improve relative scoring, but also produce more accurate absolute scores by better matching the target scale. These results indicate that, on acoustic consistency benchmarks such as SALMon, evaluating SLMs with our proposed methods yields judgments that are better aligned with human perception than those obtained using global token perplexity.
% FORFINAL In addition, the regression slopes of these approaches are much closer to 1 \footnote{The optimal slope is obtained by linearly anchoring the random-performance baseline, accuracy (=50) to MOS (=1), and the perfect-performance ceiling, accuracy (=100) to MOS (=5).}, indicating that they not only improve relative scoring but also yield more accurate absolute scores. 
In the bottom row of Figure~\ref{fig:correlation}, we replicate the analysis using the embedding model-as-a-judge scores as a proxy of the MOS scores given its high correlation, to conduct correlations with likelihood-methods on \underline{all} models. The results reinforce the conclusion that windowed, normalized, and localized methods correlate better to true generation performance. 

Next, we examine how well evaluator rankings correlate with MOS scores for each individual split. In Table~\ref{tab:kendall-tau}, we report the Kendall $\tau$ scores on the ranking of the 7 models that are universally present in the top row of Figure~\ref{fig:correlation}. The results show that normalized and windowed likelihood-based metrics, as well as model-as-a-judge scoring methods, yield substantially stronger rank agreement with MOS across acoustic-consistency tasks, leading to higher average Kendall’s $\tau$ overall. In particular, the overall Kendall’s $\tau$ increases from 0.444 to 0.480 with normalized global perplexity, to 0.502 with windowed perplexity, and to 0.674 with model-as-a-judge evaluation.

Collectively, these results establish a principled and scalable evaluation framework for continuation quality. Among likelihood-based methods, normalized perplexity is preferred over global token perplexity because it shows stronger correlation with MOS and better preserves model rankings. With normalized perplexity as the evaluation metric, the performance landscape of SLMs is completely reshaped. Most notably, the best-performing model Llama-Mimi improves from 80.92 to 90.42, closing 84.1\% of the gap to the human topline on SALMon and achieving a new state of the art. While generation-based model-as-a-judge evaluation is not universally superior to normalized perplexity in terms of correlation with the MOS scores, it remains valuable for assessing SLMs in cases where token generation does not fully capture audio generation quality, as in TASTE. 
Despite these advancements, human inspection of the generated samples reveals that these SLMs still have substantial room for improvement in handling complex speech signals, underscoring the need for more rigorous benchmarks.

% \textcolor{red}{
% \begin{itemize}
%     \item For model generations that only take token as input, we conduct correlation on perplexity, continuation-judge, and continuation-human. These include GSLM series, spirit-lm, twist, llama-mimi. 
%     \item For model generations that also include an additional acoustic feature, intepretability deviates from pure token performance measurement. So, we contrast only continuation-judge with continuation-human.
%     \item Results show that the correlation is quite high, validating our selection of judge models. 
%     \item Results show that local perplexity correlates better with continuation performance. 
% \end{itemize}
% }
%\input{tables/composition/exp}

%\section{Analysis}
\subsection{A Tale of Two Models: How Loss Composition Shapes Performance}
\label{sec:analysis-comp}

Building on our earlier discussion that acoustic quality can be decomposed into interpretable axes (e.g., speaker- and background-related attributes), we expect the model’s NLL loss to be governed by an analogous set of disentangleable components.
Breaking the loss down by axis would quantify each attribute’s contribution and its effect on performance.
%and thus rationalize the contrasting behavior of our alternative methods versus global perplexity.
Nonetheless, an explicit axis-wise decomposition of the loss remains difficult in practice, because the relevant factors are entangled within a single embedding or even within individual tokens. Fortunately, we can consider Spirit-LM-Expressive and Llama-Mimi as analytical lenses for this study, since their token inventories are inherently functionally distinct. 
Spirit-LM-Expressive comprises three token types during sequence construction: HuBERT, pitch, and style; whereas Llama-Mimi employs tokens from different RVQ layers, which similarly exhibit functional heterogeneity.
Their original works \cite{sugiura2025llama, spiritlm} already categorized them as reflecting semantic versus acoustic utility. Using these models as representative cases, we contrast composition profiles to account for the opposite effects of our alternative methods in Llama-Mimi and Spirit-LM-Expressive.
We perform a comprehensive combinatorial ablation across token types, enabling Shapley value analysis \cite{value1953shapley} to quantify each token type’s marginal contribution. 
%Using these models as paradigmatic cases, we illustrate how loss composition along with its inherent volitity our proposed approaches affect uncertainty relative to global perplexity. 

\TabCompositionShapleyAvgOnly
% \begin{figure}[t!]
%   \centering
%   % tighten spacing between cells
%   \setlength{\tabcolsep}{4pt}
%     \includegraphics[width=0.48\textwidth]{figs/spiritlm-expressive-HPS-composition.pdf}\\
%     \includegraphics[width=0.48\textwidth]{figs/llama-mimi-composition.pdf}
%  \caption{Composition of the average per-sample advantage, which is defined as the difference between the negative loss and the positive loss. Advantage differs across evaluation methods in both token-type composition and loss magnitude.}
%   \label{fig:loss-composition}
% \end{figure}

Table~\ref{tab:composition-shapley-avg-only} exhibits the Shapley contributions of each token type. For Llama-Mimi, tokens from different residual layers all contribute positively, with the largest contribution coming from residual layer 1 ($\phi_1$). Contributions of individual layers are further amplified in normalized and localized settings. On average, normalized settings improved +2.4 points, and localized settings improved +2.8 points spread evenly across token types. This improvement clearly carries over to the final accuracy score, where the normalized method achieves +9.42 points and the localized method achieves +11.08 points.
% Figure~\ref{fig:loss-composition} shows the overall advantage across methods, which is defined as the average loss difference between negative and positive samples. We further decompose this gap into a token-level contribution profile, revealing which tokens drive the advantage. In Llama-Mimi, localization and normalization exhibit largely orthogonal effects: moving from a global loss to a transitioned-windowed loss roughly doubles the measured advantage, whereas normalization slightly shifts the relative contributions of individual token types. 

The case of Spirit-LM-Expressive is markedly different from Llama-Mimi. In Table~\ref{tab:composition-shapley-avg-only}, HuBERT tokens contribute to acoustic tasks more than the other tokens combined, despite being labeled as a "semantic" token type. Even with localization, the HuBERT Shapley value remains unchanged, suggesting that a non-trivial portion of the acoustic inconsistencies captured by the SLM is distributed throughout the full sequence. Furthermore, normalization leads to an additional decrease in the HuBERT tokens’ Shapley values, indicating a strong entanglement between semantic and acoustic information in Spirit-LM-Expressive’s HuBERT tokens. As a result, reducing semantic influence also diminishes acoustic information.
%acoustic information they encode is only elicitable through long-range dependencies.
%Surprisingly, normalization further reduces HuBERT’s Shapley values, pointing to that the modeling of non-acoustic information Spirit-LM-Expressive is highly volatile.
As reflected in the accuracy scores, these alternative methods do not improve performance on the SALMon benchmark. The increase in Shapley values for pitch tokens is offset by the decrease in Shapley values for HuBERT tokens, which remain equally crucial for capturing acoustic information.

As we uncover the role of HuBERT tokens in Spirit-LM-Expressive, it becomes clearer why models that adopt HuBERT tokens (GSLM, TWIST, pGSLM, Spirit-LM) fail to translate high classification scores into strong continuation performance as measured by MOS. The global-perplexity performance of HuBERT modeling depends substantially on long-range dependencies to judge correctness, whereas continuation quality must be established locally and cannot rely on future context. Our proposed evaluation penalizes this behavior, resulting in better agreement with MOS-based continuation outcomes.

\section{Conclusion}    

In this work, we revisit the use of global token perplexity for evaluating SLMs. We introduce a variety of likelihood- and generation-based evaluation methods, highlighting localization and normalization methods, to better reflect key characteristics of speech. Correlations with MOS indicate that our proposed methods better reflect human perception. Under this re-evaluation, the previously best-performing model closes 83\% of the gap to the human topline on SALMon. Together, these findings reshape the SLM performance landscape and establish a new evaluation paradigm for future studies.

\section*{Limitations}
We propose novel evaluation methods as alternatives to conventional global token perplexity. However, since these methods are still applied to existing benchmarks, their scope remains inherently constrained by the limitations of those benchmarks. For instance, SALMon does not systematically probe compounded variations (e.g., speaker changes under noisy background conditions), which restricts our ability to characterize SLM performance in such settings—even when using our improved evaluators. In addition, we focused our discussion on acoustic continuity, which constitutes a substantial and distinctive aspect of speech. For other dimensions, such as semantics, global perplexity may still be, and very likely remains, the most appropriate approach. Nonetheless, the broader notion of “speech perplexity” warrants careful scrutiny, as different aspects of speech are inherently entangled.

\bibliography{custom}

\appendix

%\section{Appendix}
\section{Further Evaluations of Spoken Language Models}
While \textit{universal task robustness} serves as the prevailing objective, evaluations follow a curriculum that mirrors the model’s training progression. In the main text, we have shown that for pretrained SLMs, likelihood-based evaluation is prominent. 
For finetuned models, task-based evaluation metrics pinpoint utility, such as speech recognition \cite{panayotov2015librispeech}, emotion recognition \cite{busso2008iemocap}, keyword spotting \cite{warden2018speechcommands}, and speaker identification \cite{nagrani2017voxceleb}. Aggregated suites such as the SUPERB series \cite{yang2021superb,tseng2024avsuperb,tsai2022superbsg, huang2024adynamic, huang2024bdynamic} broaden coverage, though their constituent tasks remain predefined. Prompt-following benchmarks built around arbitrary natural-language instructions highlight general intelligence of finetuned SLMs \cite{yang2024airbench,gao2025adubench,hou2025sovabench,chen2024voicebench}.

\section{Verbose Evaluation Settings}

\subsection{MOS Evaluation Prompt}
\label{sec:apx-MOS-prompt}

The annotators are given the following prompt:
\begin{mdframed}

You will hear a generated continuation that extends from an audio prompt. Compare the two audio clips only on the specific target **feature** that varies between samples, focus only on the similarity of that feature, and assign a similarity score from 1 to 5.

A score of 5 indicates perfect match on the feature (gender/speaker/sentiment/background, etc.). The two audios are indistinguishable on that feature. Naturally, two identical audios will score 5 on any given feature.

A score of 1 indicates complete mismatch on the feature (gender/speaker/sentiment/background, etc.). The two audios are easily distinguishable on that feature. Naturally, if the target feature is missing entirely, the score is unequivocally 1.

Target feature is considered missing when:

The audio is completely silent.
The attribute is gender / speaker / sentiment but there is no speech (sound made by humans).
The attribute is background but there is only human speech and no other acoustic source.
Use the guideline below for sample comparisons and guidance for scores 2–4.

Scoring guidelines:

5 – Indistinguishable from the prompt on the target attribute.

4 – Only distinguishable with close attention to small sections; casual listening still feels nearly identical.

3 – Distinguishable, but most attribute traits still feel similar.

2 – Clearly distinguishable with only minor attribute overlap.

1 – Totally distinguishable; no attribute similarity at all.

(Examples for each score)
\end{mdframed}

\section{Verbose Evaluation Results}

\subsection{Likelihood-Based Evaluations}
\label{sec:apx-exp-ppl}

Table~\ref{tab:ppl-main} shows full evaluation results on SALMon across spoken language models. Generally, alternative uncertainty methods evaluate HuBERT-based models (GSLM, TWIST, SpiritLM) lower, and Mimi-based models (Flow-SLM, Llama-Mimi) higher. For the top-performing models, our proposed evaluation methods occasionally deliver results that place model performance above the human topline. Notably, these cases are concentrated around speaker information. 

For semantic-acoustic alignment tasks, there is not a common speech prompt, hence only Global-PPL and Windowed-PPL evaluation are supported. Results show consistent low accuracies ($<60\%$), regardless of the model used or the methods selected. This agreement confirms that alignment is likely a trait that is not picked up by current SLMs, motivating future work on this direction.
\TabPPLMain

\subsection{Model-as-a-Judge for Measuring
Generation Consistency}
\label{sec:apx-exp-cont-judge}
\TabJudgeDev
%\TabJudgeTestEnt
\TabJudgeTestGen

Table~\ref{tab:judge-dev} exhibits task-wise performance on embedding judge candidates. From the results, we observe that no single embedding model aces all six attributes, motivating future work on more generalizable audio embedding approaches. In addition, these results indicate that acoustic features (speaker features, background features) are effectively time-invariant at the resolution probed by current evaluation protocols. Such stability endorses a prevalent architectural choice of conducting speech synthesis conditioned on a constant residual acoustic embedding. Notably, CAM++ serves as a highly versatile, general-purpose acoustic representation model and is widely adopted in generative speech systems \cite{tseng2025taste, du2024cosyvoice, hsu2025breezyvoiceadaptingttstaiwanese}. Related design choices also appear in other spoken-language models with speech generation \cite{spiritlm, gslm, twist} and conversational speech frameworks \cite{defossez2024moshi}.

Table~\ref{tab:judge-test-ent50} shows the results of SLM performance evaluated by selected judge models. Similar to the conclusions made in the proposed likelihood estimation methods, evaluating on true continuations shows scores are in the vicinity of the human topline for top performing models, especially on traits related to human speech (sentiment, speaker, gender). 

Whereas Llama-Mimi integrates a deeper hierarchy of Mimi token layers into its speech modeling pipeline, Flow-SLM deploys a more intricate flow-matching speech decoder, which may account for its elevated scores on certain subtasks measuring directly in the speech.

Overall, HuBERT-based models perform close to chance level. A notable exception is Spirit-LM-Expressive, which includes additional pitch and style tokens. Experiment shows that this additional information is best reflected in sentiment performance, reaching 72\%. 

A closer examination of reconstruction and continuation performance reveals that the failure in continuation arises from the reconstructed audio lacking the relevant content to begin with. In most cases, reconstructions achieve near-chance scores mirroring their continuation counterparts, which is strikingly poor given that evaluation on the original audio yields near-ceiling performance on the benchmark. Manual inspection of the audio reveals that the audio is indeed greatly distorted compared to the original sample, where semantic pronunciations are greatly preserved but speaker and background information collapses. This example illustrates a key limitation of global uncertainty measures. Even when they showed moderate performance, such performance failed to generalize to continuation performance that requires local acoustic fidelity at each generation step.

\subsection{Shapley Analysis}
\TabCompositionComprehensive

Table~\ref{tab:composition-comprehensive} shows token-type ablations and Shapley attributions for Spirit-LM-Expressive (HuBERT $H$, pitch $P$, style $S$) and Llama-Mimi (layers $0$--$3$), under four evaluation settings. While Spirit-LM-Expressive’s largest contribution comes from HuBERT tokens, Llama-Mimi’s largest contribution comes from layer~1. Across settings, the most pronounced differences concentrate on speaker-related attributes (sentiment, speaker, and gender).

\subsection{Loss Response Figures}
\begin{figure*}[t]
  \centering
  \includegraphics[width=\textwidth,keepaspectratio]{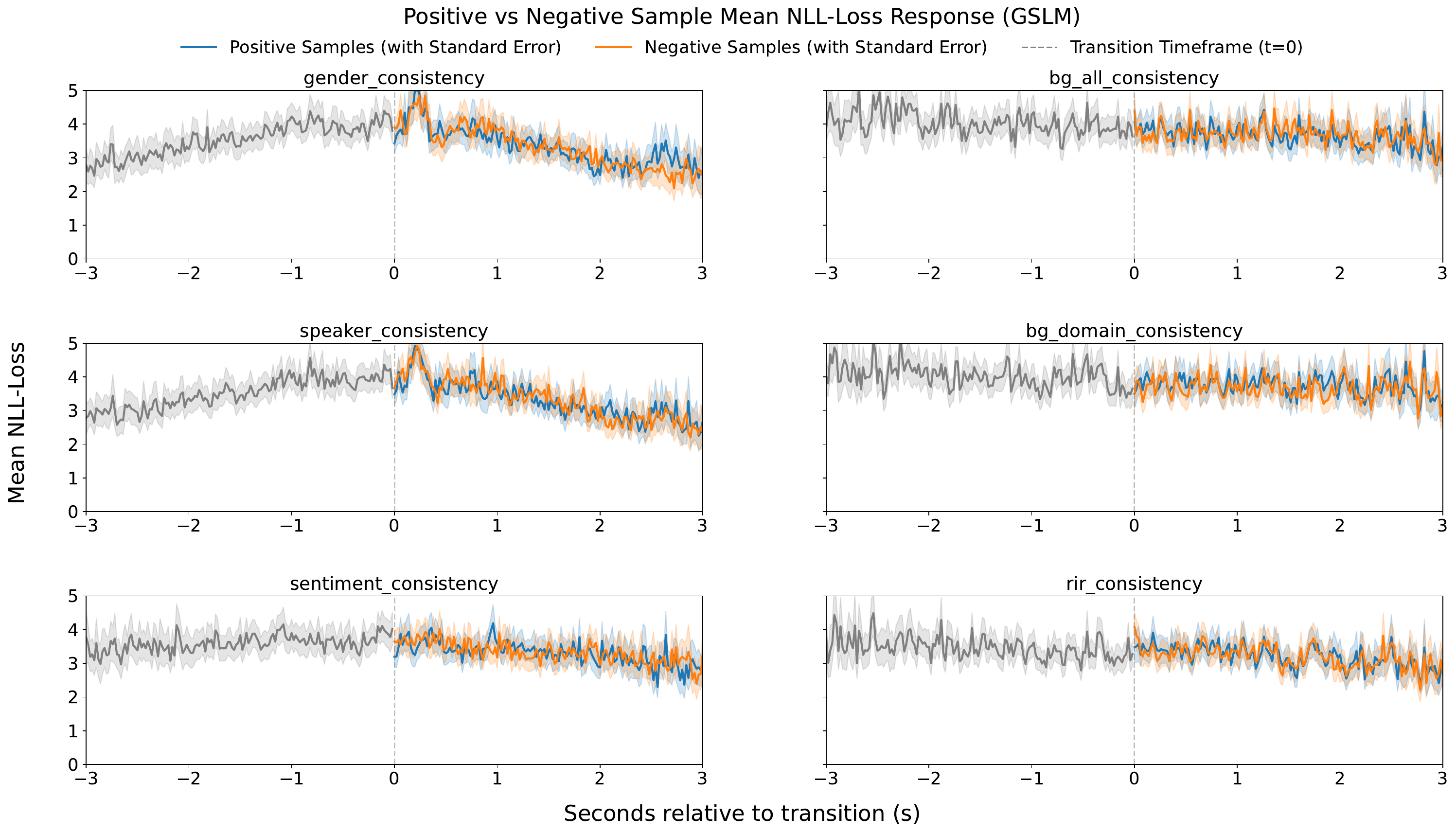}
  \caption{Positive vs. Negative Sample Mean NLL-Loss Response for \texttt{GSLM} across six consistency splits.}
  \label{fig:gslm-posneg-grid}
\end{figure*}

\begin{figure*}[t]
  \centering
  \includegraphics[width=\textwidth,keepaspectratio]{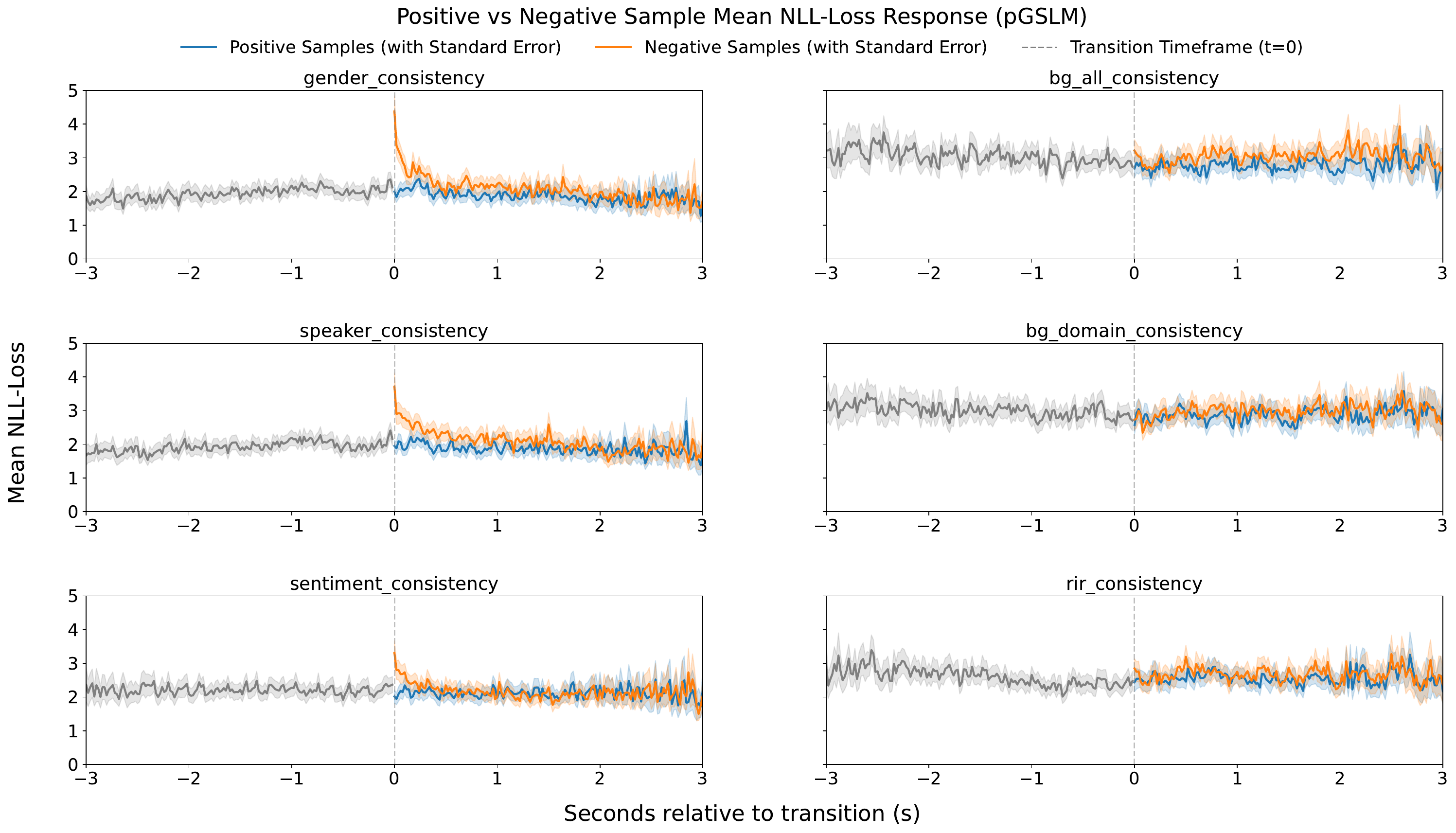}
  \caption{Positive vs. Negative Sample Mean NLL-Loss Response for \texttt{pGSLM} across six consistency splits.}
  \label{fig:pgslm-posneg-grid}
\end{figure*}

\begin{figure*}[t]
  \centering
  \includegraphics[width=\textwidth,keepaspectratio]{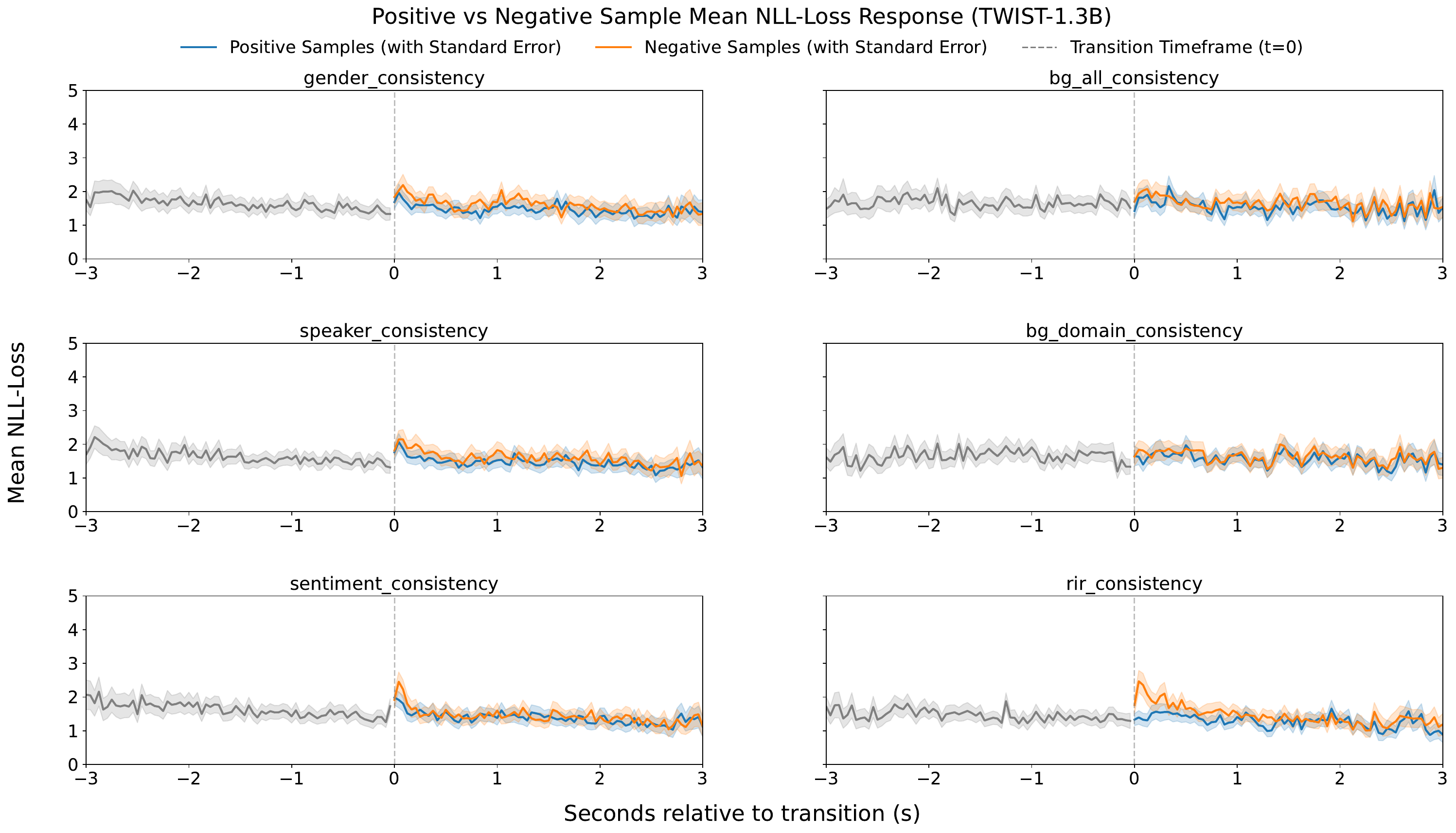}
  \caption{Positive vs. Negative Sample Mean NLL-Loss Response for \texttt{TWIST-1.3B} across six consistency splits.}
  \label{fig:TWIST-1.3B-posneg-grid}
\end{figure*}

\begin{figure*}[t]
  \centering
  \includegraphics[width=\textwidth,keepaspectratio]{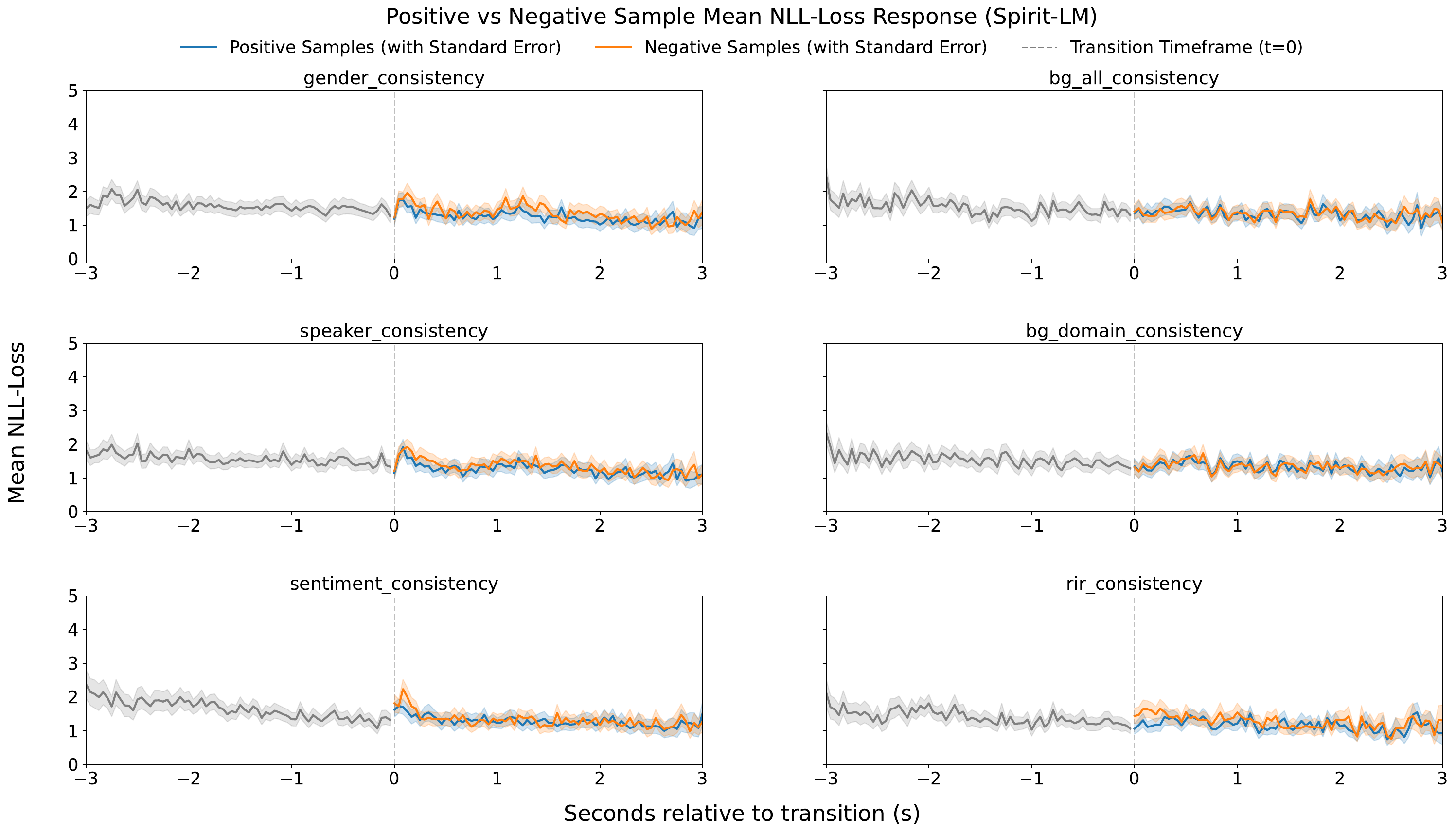}
  \caption{Positive vs. Negative Sample Mean NLL-Loss Response for \texttt{Spirit-LM} across six consistency splits.}
  \label{fig:spirit-posneg-grid}
\end{figure*}

\begin{figure*}[t]
  \centering
  \includegraphics[width=\textwidth,keepaspectratio]{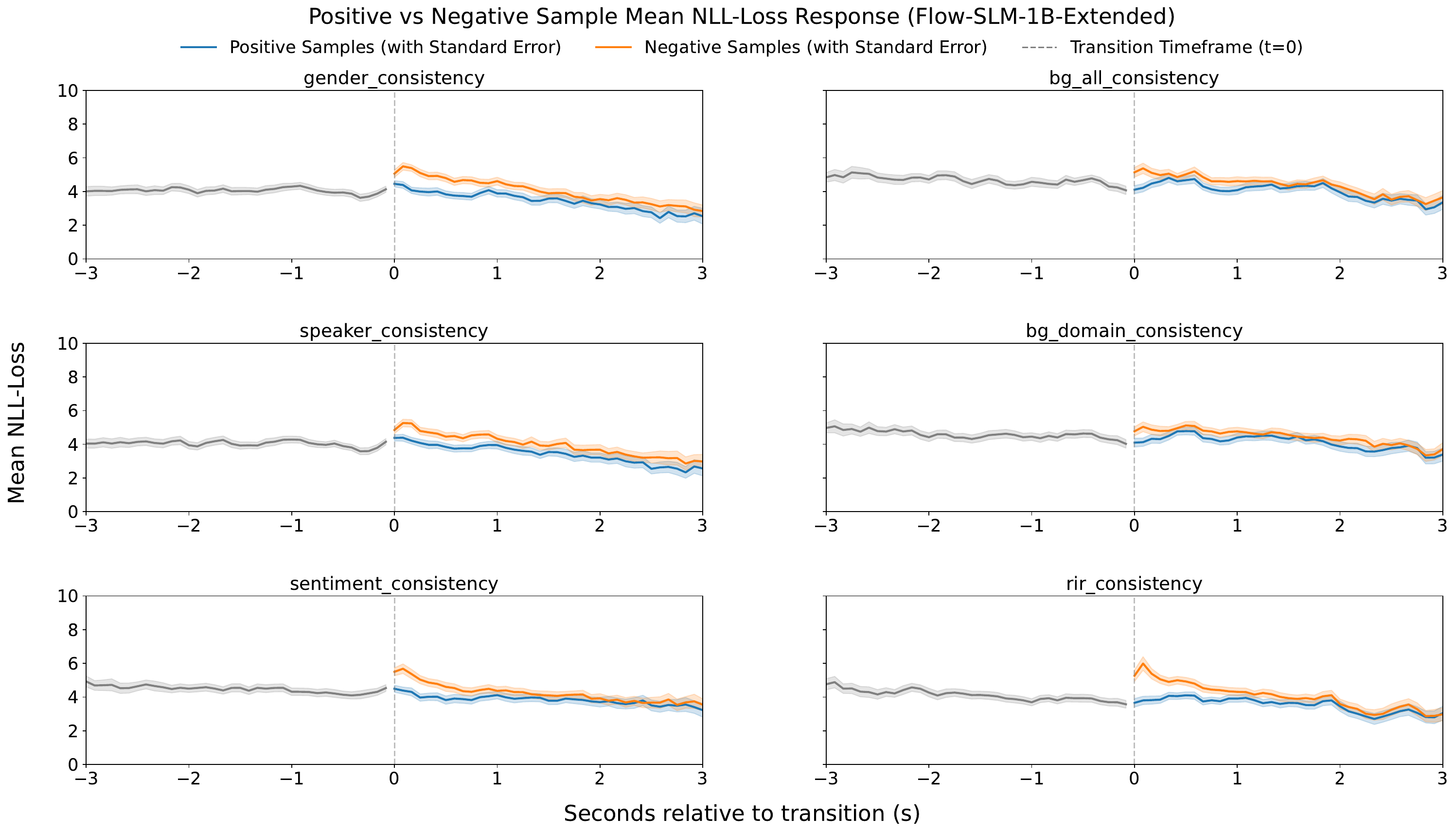}
  \caption{Positive vs. Negative Sample Mean NLL-Loss Response for \texttt{Flow-SLM-1B-Extended} across six consistency splits. At the transition timeframe (t=0), each of the category has distinct postive and negative response (clear separation by 95\% confidence interval). }
  \label{fig:Flow-SLM-posneg-grid}
\end{figure*}

\begin{figure*}[t]
  \centering
  \includegraphics[width=\textwidth,keepaspectratio]{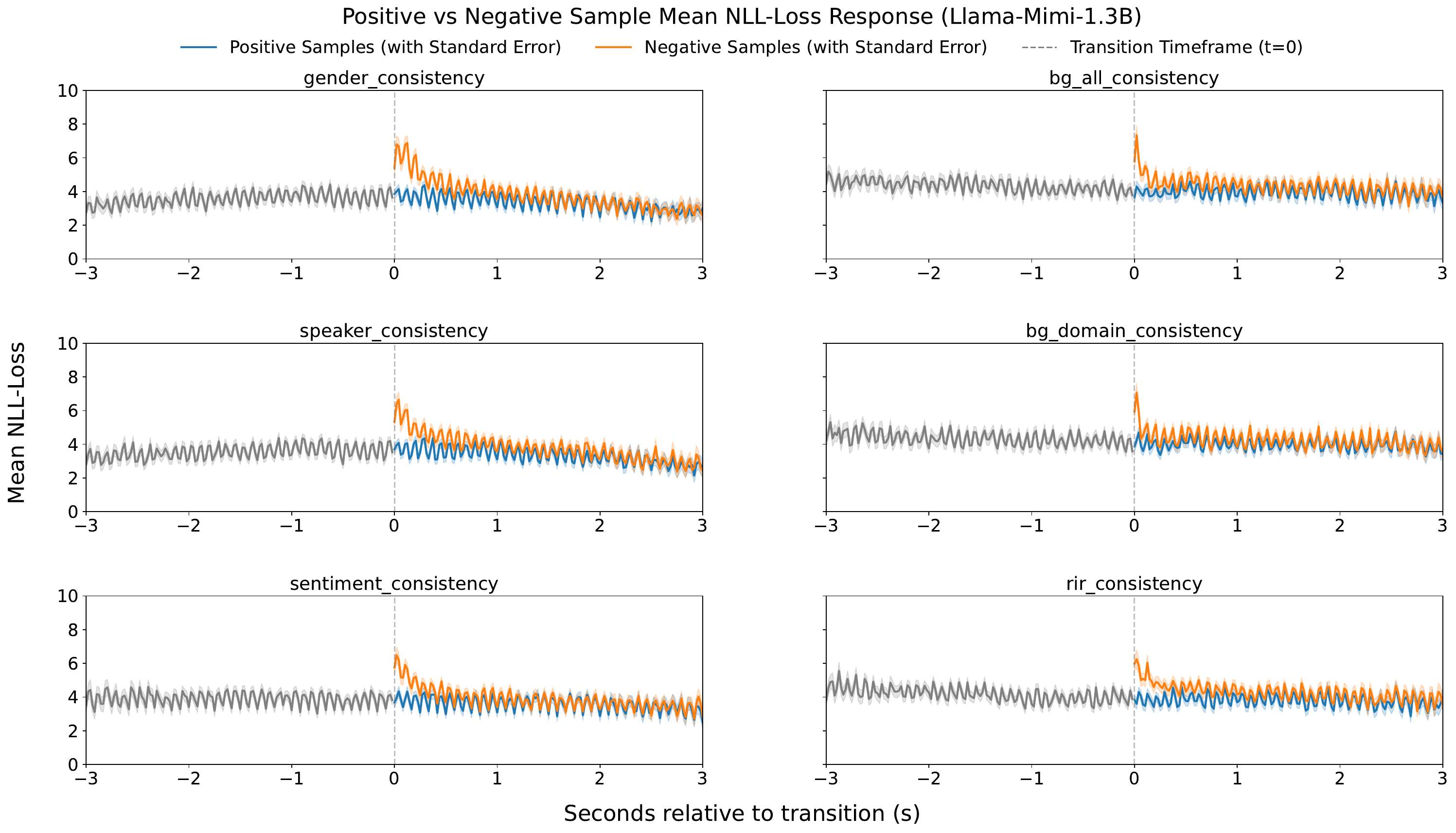}
  \caption{Positive vs. Negative Sample Mean NLL-Loss Response for \texttt{Llama-mimi-1.3B} across six consistency splits. At the transition timeframe (t=0), each of the category has distinct postive and negative response (clear separation by 95\% confidence interval). }
  \label{fig:llama-mimi-posneg-grid}
\end{figure*}

Figures~\ref{fig:gslm-posneg-grid} through~\ref{fig:llama-mimi-posneg-grid} show a per-task breakdown of the models’ NLL-loss responses. The degree of separability between the positive and negative NLL responses in these plots largely correlates with the resulting accuracy. Consistent with this, speaker-related attributes (sentiment, speaker identity, and gender) exhibit larger separations.

%\subsection{Ablation}

%\TabCompositionComprehensiveLlamaMimiOnly
%\TabCompositionSpirExpr
%\TabCompositionSpirExprNorm

\begin{figure*}[t]
  \centering
  \includegraphics[width=\textwidth,keepaspectratio]{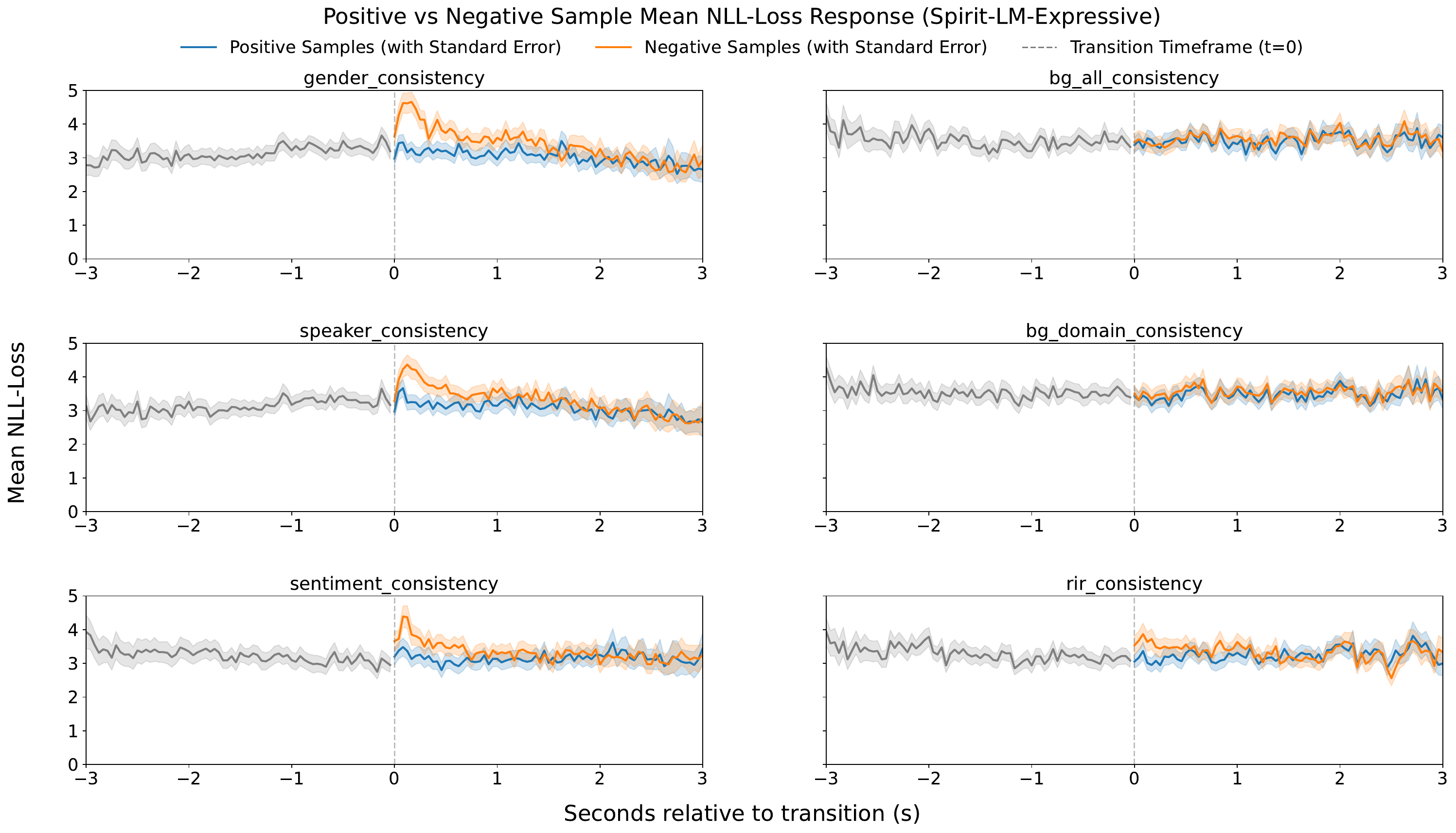}
  \caption{Positive vs. Negative Sample Mean NLL-Loss Response for \texttt{Spirit-LM-Expressive} across six consistency splits.}
  \label{fig:spirit-expr-posneg-grid}
\end{figure*}

\begin{figure*}[t]
  \centering
  \includegraphics[width=\textwidth,keepaspectratio]{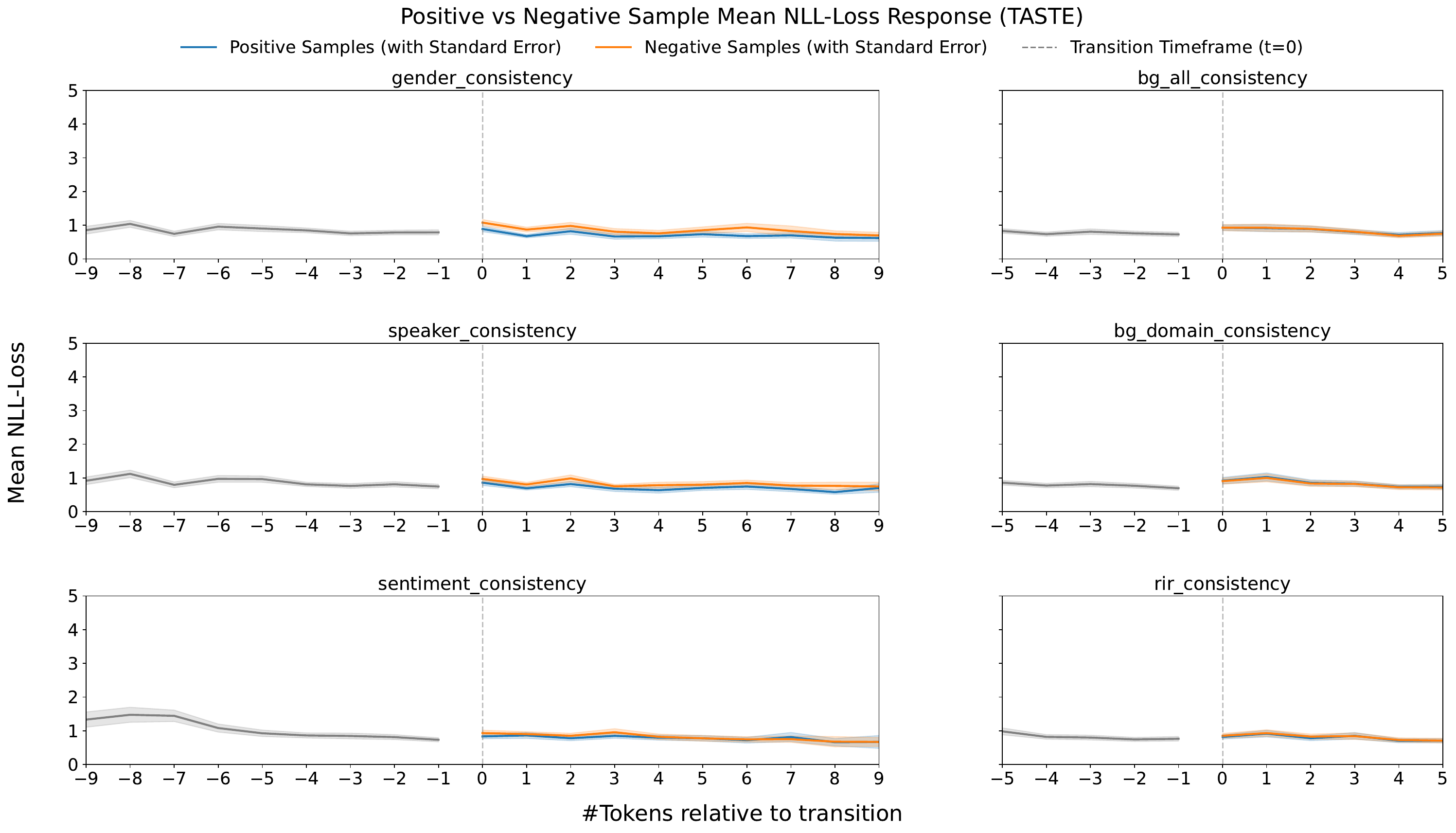}
  \caption{Positive vs. Negative Sample Mean NLL-Loss Response for \texttt{TASTE} across six consistency splits. We follow TASTE's audio-text alignment setting and report with textual tokens as the granularity of the x-axis. }
  \label{fig:spirit-expr-posneg-grid}
\end{figure*}

\end{document}